\documentclass[dvipsnames]{article} 
\usepackage{iclr2023_conference,times}

\usepackage{amsmath,amsfonts,bm}

\def\eqref#1{equation~\ref{#1}}

\def\1{\bm{1}}

\DeclareMathAlphabet{\mathsfit}{\encodingdefault}{\sfdefault}{m}{sl}
\SetMathAlphabet{\mathsfit}{bold}{\encodingdefault}{\sfdefault}{bx}{n}

\usepackage[colorlinks,
            linkcolor=black,
            anchorcolor=blue,
            citecolor=MidnightBlue,
            urlcolor=magenta
            ]{hyperref}

\usepackage{hyperref}
\usepackage{url}

\usepackage{booktabs}       
\usepackage{nicefrac}       
\usepackage{microtype}      
\usepackage{subfigure}
\usepackage{csquotes}
\usepackage{amsthm,amssymb,amsfonts,amsmath}

\usepackage{caption,subfigure,graphicx,epstopdf,multirow,multicol,booktabs,verbatim,wrapfig,bm}

\usepackage{makecell}
\theoremstyle{plain}

\newtheorem{theorem}{Theorem}
\newtheorem{proposition}[theorem]{Proposition}
\newtheorem{lemma}[theorem]{Lemma}

\newcommand{\red}[1]{{\textcolor{BrickRed}{#1}}}

\newcommand{\redbf}[1]{\bf{\textcolor{BrickRed}{#1}}}
\newcommand{\bluebf}[1]{\bf{\textcolor{MidnightBlue}{#1}}}

\usepackage[colorinlistoftodos,prependcaption]{todonotes}
\usepackage{xargs}
\newcommandx{\cmt}[2][1=Comment]{\vspace{2pt}\todo[inline,backgroundcolor=black!5]{\textbf{(#1)} \; #2}}

\title{Graph Neural Networks are Inherently Good Generalizers: Insights by Bridging GNNs and MLPs}

\author{Chenxiao Yang, Qitian Wu, Jiahua Wang \& Junchi Yan\thanks{Correspondence author is Junchi Yan who is also affiliated with Shanghai AI Laboratory. The work was in part supported by National Key Research and Development Program of China (2020AAA0107600), NSFC (62222607), and STCSM (22511105100).}\\
Department of CSE \& MoE Key Lab of Artificial Intelligence, Shanghai Jiao Tong University\\
\texttt{\{chr26195,echo740,wangjiahua2001,yanjunchi\}@sjtu.edu.cn} \\}

\newcommand{\bs}{\boldsymbol}

\iclrfinalcopy
\begin{document}

\maketitle

\begin{abstract}
Graph neural networks (GNNs), as the de-facto model class for representation learning on graphs, are built upon the multi-layer perceptrons (MLP) architecture with additional message passing layers to allow features to flow across nodes. While conventional wisdom commonly attributes the success of GNNs to their advanced expressivity, we conjecture that this is \emph{not} the main cause of GNNs' superiority in node-level prediction tasks. This paper pinpoints the major source of GNNs' performance gain to their intrinsic generalization capability, by introducing an intermediate model class dubbed as P(ropagational)MLP, which is identical to standard MLP in training, but then adopts GNN's architecture in testing. Intriguingly, we observe that PMLPs consistently perform on par with (or even exceed) their GNN counterparts, while being much more efficient in training. Codes are available at \url{https://github.com/chr26195/PMLP}. 

This finding sheds new insights into understanding the learning behavior of GNNs, and can be used as an analytic tool for dissecting various GNN-related research problems. As an initial step to analyze the inherent generalizability of GNNs, we show the essential difference between MLP and PMLP at infinite-width limit lies in the NTK feature map in the post-training stage. Moreover, by examining their extrapolation behavior, we find that though many GNNs and their PMLP counterparts cannot extrapolate non-linear functions for extremely out-of-distribution samples, they have greater potential to generalize to testing samples near the training data range as natural advantages of GNN architectures.
\end{abstract}

\section{Introduction}
In the past decades, \emph{Neural Networks} (NNs) have achieved great success in many areas. As a classic NN architecture, \emph{Multi-Layer Perceptrons} (MLPs)~\citep{rumelhart1986learning} stack multiple \emph{Feed-Forward} (FF) layers with nonlinearity to universally approximate functions. Later, \emph{Graph Neural Networks} (GNNs)~\citep{scarselli2008graph,bruna2014spectral,gilmer2017neural,kipf2017semi,velivckovic2017graph,hamilton2017inductive,klicpera2018predict,wu2019simplifying} build themselves upon the MLP architecture, e.g., by inserting additional \emph{Message Passing} (MP) operations amid FF layers~\citep{kipf2017semi} to accommodate the interdependence between instance pairs.

Two cornerstone concepts lying in the basis of deep learning research are model's \emph{representation} and \emph{generalization} power. While the former is concerned with what function class NNs can approximate and to what extent they can minimize the \emph{empirical risk} \scalebox{0.9}{${\widehat{\mathcal R}(\cdot)}$}, the latter instead focuses on the inductive bias in the learning procedure, asking how well the learned function can generalize to unseen in- and out-of-distribution samples, reflected by the \emph{generalization gap} \scalebox{0.9}{${\mathcal R}(\cdot) - {\widehat{\mathcal R}(\cdot)}$}. There exist a number of works trying to dissect GNNs' representational power (e.g., ~\cite{scarselli2008computational,xu2018powerful,maron2019provably,oono2019graph}), while their generalizability and connections with traditional neural networks are far less well-understood.

\begin{figure}[t!]
\hspace{-22pt}
	\includegraphics[width=1.06\textwidth]{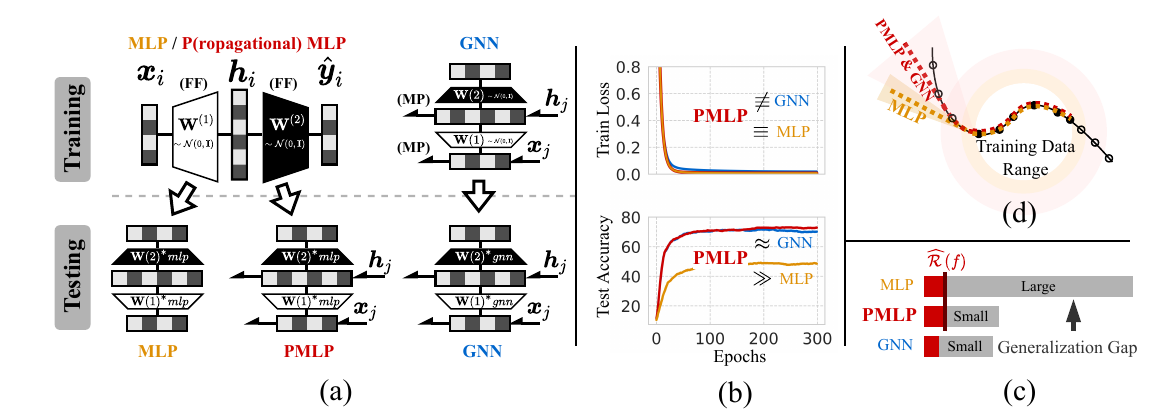}
	\caption{\textbf{\textit{(a) Model illustration}} for MLP, GNN and PMLP. The proposed PMLP is an intermediate model class that bridges MLP and GNN by using MLP architecture in training and GNN architecture in testing. \; \textbf{\textit{(b) Learning curves}} for node classification on Cora that depicts a typical empirical phenomenon: Both MLP and GNN are expressive enough to achieve almost zero loss in training, while PMLP and GNN that use message passing operations in testing have significantly better testing accuracy. \;\textbf{\textit{(c) Inherent generalizability of GNNs}}: The GNN architecture used in inference for PMLP and original GNN inherently encourages lower generalization gap. \;\textbf{\textit{(d) Extrapolation illustration}}: Both MLP and PMLP tend to linearize outside the training data support ($\bullet$: train sample, $\circ$: test sample), while PMLP and GNN transit more smoothly and thereby have larger potential to generalize to testing samples near the training data support under distribution shifts.} \label{fig_frame}
\end{figure}

In this work, we bridge GNNs and MLPs by introducing an intermediate model class called \emph{Propagational MLPs} (PMLPs). During training, PMLPs are exactly the same as a standard MLP (e.g., same architecture, data, loss function, optimization algorithm). In the testing phase, PMLPs additionally insert non-parametric MP layers amid FF layers, as shown in Fig.~\ref{fig_frame}(a), to align with various GNN architectures.

\subsection{Empirical Results and Implications}
According to experiments across various node classification benchmarks and additional discussions on different architectural choices (i.e., layer number, hidden size), model instantiations (i.e., FF/MP layer implementation) and data characteristics (i.e., data split, amount of structural information), we identify two-fold intriguing empirical phenomenons:

\textbf{$\circ$\; Phenomenon 1: PMLP significantly outperforms MLP.\;} Despite that PMLP shares the same weights (i.e., trained model parameters) with a vanilla MLP, it tends to yield lower generalization gap and thereby outperforms MLP by a large margin in testing, as illustrated in Fig.~\ref{fig_frame}(c) and (b) respectively. \emph{This observation suggests that the message passing / graph convolution modules in GNNs can inherently improve model’s generalization capability for handling unseen samples.}

The word ``inherently" underlines that such particular generalization effects are implicit in the GNN architectures (with message passing mechanism) used in inference, but isolated from factors in the training process, such as larger hypothesis space for representing a rich set of ``graph-aware" functions~\citep{scarselli2008computational}, more suitable inductive biases in model selection that prioritize those functions capable of relational reasoning~\citep{battaglia2018relational}, etc.

\textbf{$\circ$\; Phenomenon 2: PMLP performs on par with or even exceeds GNNs.\;} PMLP achieves close testing performance to its GNN counterpart in inductive node classification tasks, and can even outperform GNN by a large margin in some cases (i.e., removing self-loops and adding noisy edges). Given that the only difference between GNN and PMLP is the model architecture used in training and the representation power of PMLP is exactly the same with MLP before testing, \emph{this observation suggests that the major (but not only) source of performance improvement of GNNs over MLP in node classification stems from the aforementioned inherent generalization capability of GNNs.}

\subsection{Practical Significance}
We also highlight that PMLP, as a novel class of models (using MLP architecture in training and GNN architecture in inference), can be applied as a simple and efficient model for fast and scalable training, or used for broader analysis purposes.

\textbf{$\circ$\; PMLP as a new class of model architectures.\;} While being as effective as GNNs in many cases, PMLPs are \emph{significantly more efficient} in training ($5\sim17\times$ faster on large datasets, and $65\times$ faster for very deep GNNs).
Moreover, PMLPs are more robust against noisy edges, can be straight-forwardly combined with mini-batch training (and many other training tricks for general NNs), empower pre-trained models with test-time message passing operations, and help to quickly evaluate GNN architectures to facilitate architectural search. 


\textbf{$\circ$\; PMLP as an analytic tool.\;} PMLPs can be used for dissecting various GNN-related problems such as over-smoothing and heterophily (see Sec.~\ref{sec-exp-prob} for preliminary explorations), and in a broader sense can potentially bridge theoretical research in two areas by enabling us to conveniently leverage well-established theoretical frameworks for MLPs to enrich those for GNNs.

\subsection{Theoretical Results}
As mentioned above, our empirical finding narrows down many different factors between MLPs and GNNs to a key one that attributes to their performance gap, i.e., improvement in generalizability due to the change in network architecture. Then, a natural question arises: ``Why this is the case and how does the GNN architecture (in testing) helps the model to generalize?". We take an initial step towards answering this question:

\textbf{$\circ$\;Comparison of three classes of models in NTK regime.\;} We compare MLP, PMLP, and GNN in the \emph{Neural Tangent Kernel} (NTK) regime~\citep{jacot2018neural}, where models are over-parameterized and gradient descent finds the global optima. From this perspective, the distinction of PMLP and MLP is rooted in the change of NTK \emph{feature map} determined by model architecture while fixing their minimum RKHS-norm solutions (Proposition~\ref{prop1}). For deeper investigation, we extend the definition of \emph{Graph Neural Tangent Kernel} (GNTK)~\citep{du2019graph} to node-level tasks (Lemma~\ref{lemma3}), and derive the explicit formula for computing the feature map of a two-layer GNN.

\textbf{$\circ$\;Generalization analysis for GNNs.\;} We consider an important aspect of generalization analysis, i.e., \emph{extrapolation} for Out-of-Distribution (OoD) testing samples~\citep{xu2021neural} where testing node features become increasingly outside the training data support. Particularly, we reveal that alike MLP, both PMLP and GNN eventually converge to linear functions when testing samples are infinitely far away from the training data support (Theorem~\ref{thm4}). Nevertheless, their convergence rates are smaller than that of MLP by a factor related to node degrees and features' cosine similarity (Theorem~\ref{thm5}), which indicates both PMLP and GNN are more tolerant to OoD samples and thus have larger potential to generalize near the training data support (which is often the real-world case). We provide an illustration in Fig.~\ref{fig_frame}(d).

\section{Background and Model Formulation}~\label{sec_formulate}
Consider a graph dataset $\mathcal{G}=(\mathcal{V}, \mathcal{E})$ where the node set $\mathcal{V}$ contains $n$ nodes instances $\{(\bs x_u, y_u)\}_{u\in \mathcal V}$, where $\bs x_u\in \mathbb R^{d}$ denotes node features and $y_u$ is the label. Without loss of generality, $y_u$ can be a categorical variable or a continuous one depending on specific prediction tasks (classification or regression). Instance relations are described by the edge set $\mathcal{E}$ and an associated adjacency matrix $\mathbf{A} \in\{0,1\}^{n \times n}$.  In general, the problem is to learn a predictor model with $\hat y = f(\bs x;\mathbf \theta, \mathcal G^{k}_{\bs x})$ for node-level prediction, where $\mathcal G^{k}_{\bs x}$ denotes the $k$-hop ego-graph around $\bs x$ over $\mathcal G$.

\subsection{Graph Neural Networks and Multi-Layer Perceptrons.}
GNNs come to the spotlight for learning on graph with simple yet effective implementations and drive a surge of research works on building expressive model architectures for various graph-related tasks. The core design of GNNs lies in the message passing paradigm that recursively aggregates layer-wise neighbored information to encode structural features into node-level representations for downstream prediction. 
To probe into the connection between mainstream GNNs and MLP from the architectural view, we re-write the GNN formulation in a general form that explicitly disentangles each layer into two operations, namely a \emph{Message-Passing} (MP) operation and then a \emph{Feed-Forwarding} (FF) operation:
\begin{equation}\label{eqn-gnn-mpff}
\begin{split}
    \mbox{(MP):\quad} \tilde{\mathbf{h}}^{(l-1)}_{u}= \sum\mathop{}_{v\in\mathcal N_{u}\cup\{u\}} a_{u,v}  \cdot {\mathbf{h}}_{v}^{(l-1)},\quad 
    \mbox{(FF):\quad} {\mathbf{h}}_{u}^{(l)} = \psi^{(l)}\left(\tilde{\mathbf{h}}^{(l-1)}_{u} \right),
\end{split}
\end{equation}
where $\mathcal N_u$ is the set of neighbored nodes centered at $u$, $a_{u,v}$ is the affinity function dependent on graph structure $\mathcal G$, $\psi^{(l)}$ denotes the feed-forward process (consisting of a linear feature transformation and a non-linear activation) at the $l$-th layer, and $\mathbf h^{(0)}_u = \bs x_u$ is the initial node feature. For example, in Graph Convolution Network (GCN)~\citep{kipf2017semi}, $a_{u,v} = {\mathbf A_{uv}}/{\sqrt{\tilde d_u\tilde d_v}}$, where $\tilde d_u$ denotes the degree of node $u$. For an $L$-layer GNN, the prediction is given by $\hat y_u = \psi^{(L)}(\mathbf h_u^{(L-1)})$,
where $\psi^{(L)}$ is often set as linear transformation for regression tasks or with Softmax for classification tasks. Note that GNN models in forms of Eq.~\ref{eqn-gnn-mpff} degrade to an MLP with a series of FF layers after removing all the MP operations:
\begin{equation}
    \hat y_u = \psi^{(L)}(\cdots (\psi^{(1)}(\mathbf x_u)) = \psi(\mathbf x_u).
\end{equation}

Besides GCN, many other mainstream GNN models can be written as the architectural form defined by Eq.~\ref{eqn-gnn-mpff} whose layer-wise updating rule involves MP and FF operations, e.g., GAT~\citep{velivckovic2017graph} and GraphSAINT~\citep{zeng2019graphsaint}. 
Some recently proposed graph Transformer models such as NodeFormer~\citep{wu2022nodeformer} and DIFFormer~\citep{wu2023difformer} also fall into this category. Furthermore, there are also other types of GNN architecture represented by SGC~\citep{wu2019simplifying} and APPNP~\citep{klicpera2018predict} where the former adopts multiple MP operations on the initial node features, and the later stacks a series of MP operations at the end of FF layers. These two classes of GNNs are also widely explored and studied. For example, SIGN~\citep{rossi2020sign}, $\mbox{S}^2$GC~\citep{zhu2020simple} and GBP~\citep{chen2020scalable} follow the SGC-style, and DAGNN~\citep{liu2020towards}, AP-GCN~\citep{spinelli2020adaptive} and GPR-GNN~\citep{chien2020adaptive} follow the APPNP-style.

\begin{table}[t!]
\centering
\caption{Head-to-head comparison of three proposed PMLP models (PMLP$_{GCN}$, PMLP$_{SGC}$ and PMLP$_{APP}$) and the standard MLP. \label{tab-model}}
\resizebox{1.0\textwidth}{!}{
\setlength{\tabcolsep}{5mm}{
\begin{tabular}{@{}l|c|c|c@{}}
\toprule
        Model & Train and Valid & Inference & GNN Counterpart \\ \midrule
 MLP &  \multirow{6}{*}{$\hat y_u = \psi(\mathbf x_u)$}  & $\hat y_u = \psi(\mathbf x_u)$ & N/A   \\\cmidrule{1-1} \cmidrule{3-4}
PMLP$_{GCN}$ & &  ${\mathbf{h}}_{u}^{(l)} = \psi^{(l)}( \operatorname{MP}(\{\mathbf{h}^{(l-1)}_{v}\}_{v\in \mathcal N_u\cup\{u\}}))$ & GCN~\citep{kipf2017semi} \\ \cmidrule{1-1} \cmidrule{3-4}
PMLP$_{SGC}$ & &  $\hat y_u = \psi(\operatorname{Multi-MP}(\{\mathbf x_v\}_{v\in \mathcal V}))$ & SGC~\citep{wu2019simplifying}  \\\cmidrule{1-1} \cmidrule{3-4}
PMLP$_{APP}$ & &  $\hat y_u = \operatorname{Multi-MP}(\psi(\{\mathbf x_v\}_{v\in \mathcal V}))$  & APPNP~\citep{klicpera2018predict} \\\bottomrule
\end{tabular}}}
\end{table}

\subsection{Bridging GNNs and MLPs: Propagational MLP.}
After decoupling the MP and FF operations from GNNs' layer-wise updating, we notice that the unique and critical difference of GNNs and MLP lies in whether to adopt MP (somewhere between the input node features and output prediction). To connect two families, we introduce a new model class, dubbed as \emph{Propagational MLP} (PMLP), which has the same feed-forward architecture as a conventional MLP. During inference, PMLP incorporates message passing layers into the MLP architecture to align with different GNNs. For clear head-to-head comparison, Table~\ref{tab-model} summarizes the architecture of these models in training and testing stages.

\paragraph{Extensions of PMLP.} The proposed PMLP is generic and compatible with many other GNN architectures with some slight modifications. For example, we can extend the definition of PMLPs to GNNs with residual connections such as JKNet~\citep{xu2018representation} and GCNII~\citep{chen2020simple} by removing their message passing modules in training, and correspondingly, PMLP$_{JKNet}$ and PMLP$_{GCNII}$ become MLPs with different residual connections in training. For GNNs whose MP layers are parameterized such as GAT~\citep{velivckovic2017graph}, one can additionally fine-tune the model using the corresponding GNN architecture on top of pre-trained FF layers.

\section{Empirical Evaluation}\label{sec_exp}
We conduct experiments on a variety of node-level prediction benchmarks. \textbf{Section~\ref{sec-exp-result}} shows that the proposed PMLPs can significantly outperform the original MLP though they share the same weights, and approach or even exceed their GNN counterparts. \textbf{Section~\ref{sec-exp-dis}} shows this phenomenon holds across different experimental settings. \textbf{Section~\ref{sec-exp-prob}} sheds new insights on some research problems around GNNs including over-smoothing, model depth and heterophily. We present the key experimental results in the main text and defer extra results and discussions to Appendix~\ref{app_detail} and ~\ref{app_add}. 

\paragraph{Datasets.}
We use sixteen widely adopted node classification benchmarks involving different types of networks: three citations networks (Cora, Citeseer and Pubmed), two product co-occurrency networks (Amazon-Computer and Amazon-Photo), two coauthor-ship networks (Coauthor-CS, Coauthor-Computer), three large-scale networks (OGBN-Arxiv, OGBN-Products and Flickr), and six graph datasets with high heterophily levels (Chameleon, Squirrel, Film, Cornell, Texas, Wisconsin).

\paragraph{Baselines.}
For fair comparison, we set the layer number and hidden size to the same values for GNN, PMLP and MLP in the same dataset, namely all models can be viewed as sharing the same `backbone'. Furthermore, for variants PMLP$_{SGC}$ and PMLP$_{APP}$, we consider comparison with SGC and APPNP, respectively, and slightly modify their original architectures. For SGC, we use a multi-layer neural network instead of one-layer after linear message passing. For APPNP, we remove the residual connection (i.e., $\alpha=0$) such that the message passing scheme is aligned with other models. We basically use GCN convolution for MP layer and ReLU activation for FF layer for all models unless otherwise stated. PMLPs use the MLP architecture for validation.

\paragraph{Evaluation Protocol.} We adopt inductive learning setting as the evaluation protocol, which is a commonly used benchmark setting and close to the real scenarios where testing nodes are from new observations and unknown during training. This setting also guarantees a fair comparison between PMLP and its GNN counterpart by ensuring that the information of testing nodes can not be used in training for all models. Specifically, for node set $\mathcal V = \mathcal V_{tr} \cup \mathcal V_{te}$ where $\mathcal V_{tr}$ (resp. $\mathcal V_{te}$) denotes training (resp. testing) nodes, the training process is only exposed to $\mathcal G_{tr} = \{\mathcal V_{tr}, \mathcal E_{tr}\}$, where $\mathcal E_{tr}\subset \mathcal V_{tr} \times \mathcal V_{tr}$ only contains edges for nodes in $\mathcal V_{tr}$ and the trained model is tested with the whole graph $\mathcal G$ for prediction on $\mathcal V_{te}$.

\begin{table}[t!]
\centering
\caption{Mean and STD of testing accuracy on node-level prediction benchmark datasets.}
\resizebox{1.0\textwidth}{!}{
\setlength{\tabcolsep}{1.5mm}{
\begin{tabular}{@{}ll|ccccccc@{}}
\toprule
                      & Dataset   & Cora & Citeseer & Pubmed & A-Photo & A-Computer & Coauthor-CS & Coauthor-Physics  \\ 
                      & \#Nodes   &  2,708 & 3,327 & 19,717  & 7,650   & 13,752 & 18,333 & 34,493 \\ \hline
\hline\specialrule{0em}{2pt}{2pt}
\multirow{3}{*}{\rotatebox{90}{GNNs}} & GCN  &  $74.82 \pm 1.09$    & $67.60 \pm 0.96$ & $76.56 \pm 0.85$ & $89.69 \pm 0.87$  & $78.79 \pm 1.62$ & $91.79 \pm 0.35$ & $91.22 \pm 0.18$\\ \cmidrule(l){2-9} 
                      & SGC       &  $73.96 \pm 0.59$    & $67.34 \pm 0.54$ & $76.00 \pm 0.59$ & $83.42 \pm 2.47$  & $77.10 \pm 2.54$ & $91.24 \pm 0.59$ & $89.18 \pm 0.46$\\ \cmidrule(l){2-9} 
                      & APPNP     & $75.02 \pm 2.17$    & $66.58 \pm 0.77$ & $76.48 \pm 0.49$ & $89.51 \pm 0.86$  & $78.29 \pm 0.55$ & $91.64 \pm 0.34$ & $91.80 \pm 0.77$ \\ \hline
\hline\specialrule{0em}{2pt}{2pt}
\multirow{4}{*}{\rotatebox{90}{MLPs$\qquad\qquad$}} & MLP  &  $55.30 \pm 0.58$    & $56.20 \pm 1.27$ & $70.76 \pm 0.78$ & $75.61 \pm 0.63$  & $63.07 \pm 1.67$ & $87.51 \pm 0.51$ & $85.09 \pm 4.11$\\ \cmidrule(l){2-9} 
& \textbf{PMLP}$_{GCN}$   &  $\redbf{75.86 \pm 0.93}$    & $\redbf{68.00 \pm 0.70}$ & $\bluebf{76.06 \pm 0.55}$ & $\bluebf{89.10 \pm 0.88}$  & $\bluebf{78.05 \pm 1.21}$ & $\bluebf{91.76 \pm 0.27}$ & $\redbf{91.35 \pm 0.82}$\\
& $\Delta_{GNN}$   &   $\redbf{+1.39\%}$   & $\redbf{+0.59\%}$ & $\bluebf{-0.65\%}$ & $\bluebf{-0.66\%}$ & $\bluebf{-0.94\%}$ &      $\bluebf{-0.03\%}$  &  $\redbf{+0.14\%}$\\
& $\Delta_{MLP}$   &   $\redbf{+37.18\%}$   & $\redbf{+21.00\%}$ & $\bluebf{+7.49\%}$ & $\bluebf{+17.84\%}$ & $\bluebf{+23.75\%}$ &      $\bluebf{+4.86\%}$  & $\redbf{+7.36\%}$ \\ \cmidrule(l){2-9} 
& \textbf{PMLP}$_{SGC}$   &  $\redbf{75.04 \pm 0.95}$    & $\redbf{67.66 \pm 0.64}$ & $\redbf{76.02 \pm 0.57}$ & $\redbf{86.50 \pm 1.40}$  & $\bluebf{74.72 \pm 3.86}$ & $\bluebf{91.09 \pm 0.50}$ & $\redbf{89.34 \pm 1.40}$\\
& $\Delta_{GNN}$   &   $\redbf{+1.46\%}$   & $\redbf{+0.48\%}$ & $\redbf{+0.03\%}$ & $\redbf{+3.69\%}$ & $\bluebf{-3.09\%}$ &      $\bluebf{-0.16\%}$ & $\redbf{+0.18\%}$ \\
& $\Delta_{MLP}$   &   $\redbf{+35.70\%}$   & $\redbf{+20.39\%}$ & $\redbf{+7.43\%}$ & $\redbf{+14.40\%}$ & $\bluebf{+18.47\%}$ &      $\bluebf{+4.09\%}$ & $\redbf{+4.99\%}$ \\ \cmidrule(l){2-9} 
& \textbf{PMLP}$_{APP}$   &  $\redbf{75.84 \pm 1.36}$    & $\redbf{67.52 \pm 0.82}$ & $\bluebf{76.30 \pm 1.44}$ & $\bluebf{88.47 \pm 1.64}$  & $\bluebf{78.07 \pm 2.10}$ & $\bf 91.64 \pm 0.46$ & $\redbf{91.96 \pm 0.51}$\\
& $\Delta_{GNN}$   &   $\redbf{+1.09\%}$   & $\redbf{+1.41\%}$ & $\bluebf{-0.24\%}$ & $\bluebf{-1.16\%}$ & $\bluebf{-0.28\%}$ &      $\bf +0.00\%$ & $\redbf{+0.17\%}$ \\
& $\Delta_{MLP}$   &   $\redbf{+37.14\%}$   & $\redbf{+20.14\%}$ & $\bluebf{+7.83\%}$ & $\bluebf{+17.01\%}$ & $\bluebf{+23.78\%}$ &      $\bf +4.72\%$ & $\redbf{+8.07\%}$ \\  \bottomrule
\end{tabular}}} \label{main_result}
\end{table}

\begin{table}[t!]
\centering
 \caption{Mean and STD of testing accuracy on three large-scale datasets.}
  \label{large_result}
\label{tab_large}
\resizebox{1.0\textwidth}{!}{
\setlength{\tabcolsep}{2mm}{
\begin{tabular}{@{}c|ccc|c|ccc@{}}
\toprule
              & GCN & SGC & APPNP & MLP & \textbf{PMLP}$_{GCN}$ & \textbf{PMLP}$_{SGC}$ & \textbf{PMLP}$_{APP}$ \\ \hline
\hline\specialrule{0em}{2pt}{2pt}
$\mathop{\text{OGBN-Arxiv}}\limits_{(\Delta_{GNN}/\Delta_{MLP})}$         &  $69.04 \pm 0.18$    & $68.56 \pm 0.14$ & $69.19 \pm 0.12$ & $53.86 \pm 0.28$  & $\bluebf{\mathop{63.74 \pm 2.28}\limits_{(-7.68\%/+18.34\%)}}$ & $\bluebf{\mathop{62.65 \pm 0.35}\limits_{(-8.62\%/+16.32\%)}}$ & $\bluebf{\mathop{63.30 \pm 0.17}\limits_{(-8.51\%/+17.53\%)}}$ \\ \midrule
Train loss & $1.0480$  &  $1.1124$   & $1.0396$    &  $1.5917$   & $\bluebf{1.5917}$   &  $\bluebf{1.5917}$    &   $\bluebf{1.5917}$      \\ 
Train time &  $63.48$ ms  &  $90.50$ ms   &  $87.24$ ms   &  ${7.78}$ ms & $\redbf{7.78}$  \red{\textbf{ms}}  &  $\redbf{7.78}$  \red{\textbf{ms}}  &   $\redbf{7.78}$  \red{\textbf{ms}}    \\ \hline
\hline\specialrule{0em}{2pt}{2pt}
$\mathop{\text{OGBN-Products}}\limits_{(\Delta_{GNN}/\Delta_{MLP})}$         &  $71.35 \pm 0.19$    & $71.17 \pm 0.29$ & $70.41 \pm 0.07$ & $56.24 \pm 0.10$  & $\bluebf{\mathop{69.71 \pm 0.13}\limits_{(-2.30\%/+23.95\%)}}$ & $\bluebf{\mathop{70.09 \pm 0.13}\limits_{(-1.52\%/+24.63\%)}}$ & $\bluebf{\mathop{65.72 \pm 0.09}\limits_{(-6.66\%/+16.86\%)}}$ \\ \midrule
Train loss & $0.4013$   &  $0.4018$    &   $0.4311$   &  $1.0841$    &  $\bluebf{1.0841}$   &  $\bluebf{1.0841}$     &  $\bluebf{1.0841}$        \\ 
Train time &  $288.61$ ms  &   $665.06$ ms   &   $527.26$ ms   &    ${39.73}$  ms  &  $\redbf{39.73}$  \red{\textbf{ms}} &  $\redbf{39.73}$  \red{\textbf{ms}}     &    $\redbf{39.73}$  \red{\textbf{ms}}      \\ \hline
\hline\specialrule{0em}{2pt}{2pt}
$\mathop{\text{Flickr}}\limits_{(\Delta_{GNN}/\Delta_{MLP})}$         &  $49.66 \pm 0.57$    & $50.93 \pm 0.16$ & $45.31 \pm 0.28$ & $46.44 \pm 0.14$  & $\bluebf{\mathop{49.55 \pm 1.30}\limits_{(-0.22\%/+6.70\%)}}$ & $\mathop{\redbf{50.99 \pm 0.39}}\limits_{\redbf{(+0.12\%/+9.80\%)}}$ & $\bluebf{\mathop{44.31 \pm 0.24}\limits_{(-2.21 \%/-4.59 \%)}}$ \\ \midrule
Train loss & $1.1462$  &  $1.4030$  & $0.7449$   & $1.3344$   & $\bluebf{1.3344}$  & $\redbf{1.3344}$    &  $\bluebf{1.3344}$      \\ 
Train time &  $36.82$ ms  &  $50.42$ ms   &  $163.82$ ms   &   $7.44$ ms & $\redbf{7.44}$  \red{\textbf{ms}} & $\redbf{7.44}$  \red{\textbf{ms}}      &   $\redbf{7.44}$  \red{\textbf{ms}}    \\ \bottomrule
\end{tabular}}}
\end{table}

\subsection{Main Results}\label{sec-exp-result}

\paragraph{How do PMLPs perform compared with GNNs and MLP on common benchmarks?} 
The main results for comparing the testing accuracy of MLP, PMLPs and GNNs on seven benchmark datasets are shown in Table.~\ref{main_result}. We found that, intriguingly, different variants of PMLPs consistently outperform MLP by a large margin on all the datasets despite using the same model with the same set of trainable parameters.  Moreover, PMLPs are as effective as their GNN counterparts and can even exceed GNNs in some cases. These results suggest two implications. First, the performance improvement brought by GNNs (or more specifically, the MP operation) over MLP may not purely stem from the more advanced representational power, but the generalization ability. Second, the message passing can indeed contribute to better generalization ability of MLP, though it currently remains unclear how it helps MLP to generalize on unseen testing data. We will later try to shed some light on this question via theoretical analysis in Section~\ref{sec_theory}.

\paragraph{How do PMLPs perform on larger datasets?}
We next apply PMLPs to larger graphs which can be harder for extracting informative features from observed data. As shown in Table~\ref{large_result}, PMLP still considerably outperforms MLP. Yet differently, there is a certain gap between PMLP and GNN. We conjecture that this is because in such large graphs the relations between inputs and target labels can be more complex, which requires more expressive architectures for learning desired node-level representations. This hypothesis is further validated by the results of training losses in Table~\ref{large_result}, which indeed shows that GNNs can yield lower fitting error on the training data. Additionally, while we treat message passing operations in all models as part of their feed-forward computation for evaluating training time, in practice, one can pre-process node features for those GNNs~\citep{wu2019simplifying,chen2019powerful,rossi2020sign} where message passing operations are only used at the first layer, in which case they have the same level of efficiency as their PMLP counterparts.

\subsection{Further Discussions}\label{sec-exp-dis}

We next conduct more experiments and comparison for verifying the consistency of the observed phenomenon across different settings regarding model implementation and graph property. We also try to reveal how PMLPs work for representation learning through visualizations of the produced embeddings, and the results are deferred to Appendix~\ref{app_add}.

\begin{figure*}[t!]
\centering
\subfigure[Number of layers]{
\label{Fig.hp.1}
\includegraphics[width=0.49\textwidth]{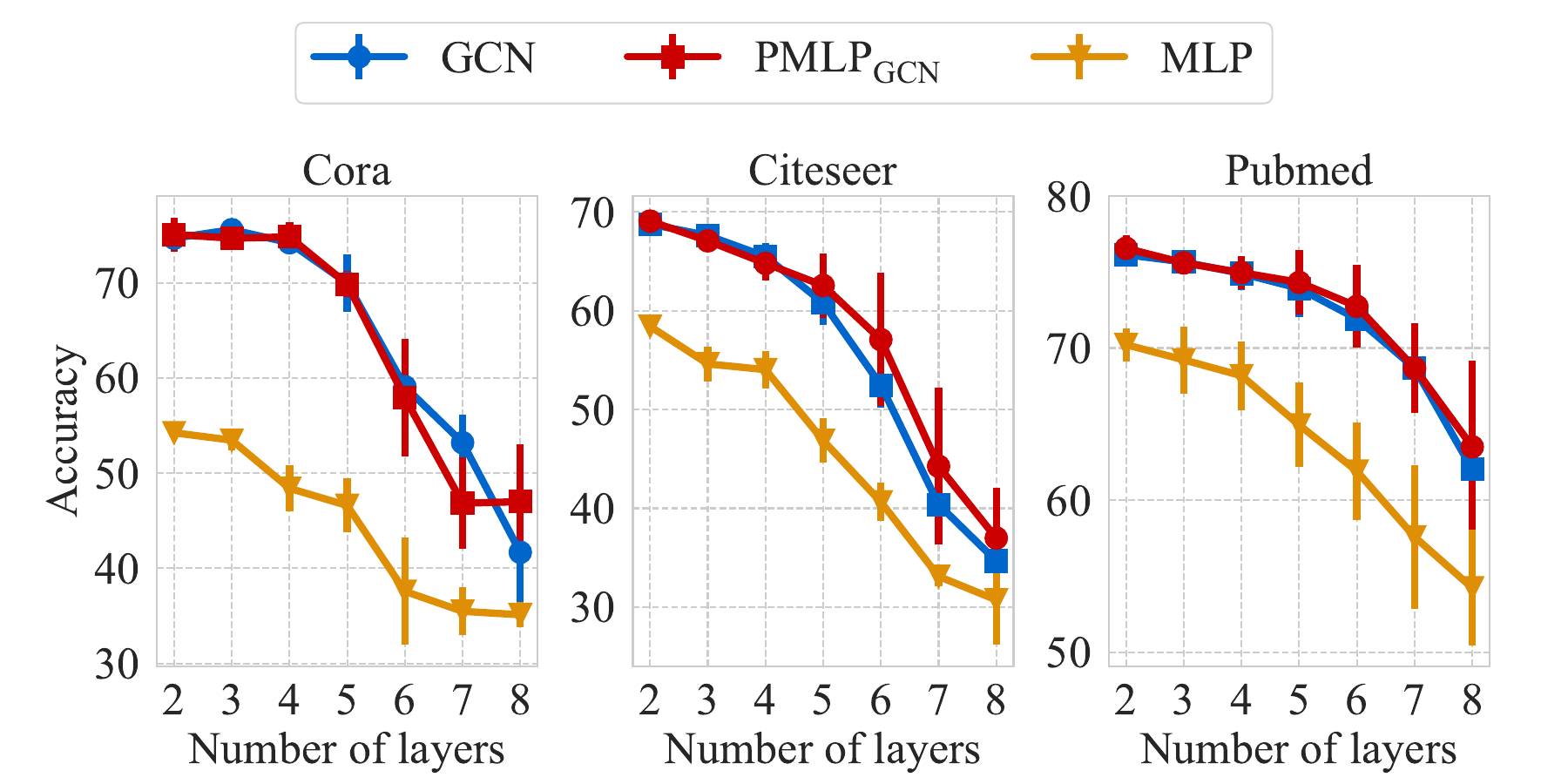}}
\subfigure[Size of hidden states]{
\label{Fig.hp.2}
\includegraphics[width=0.49\textwidth]{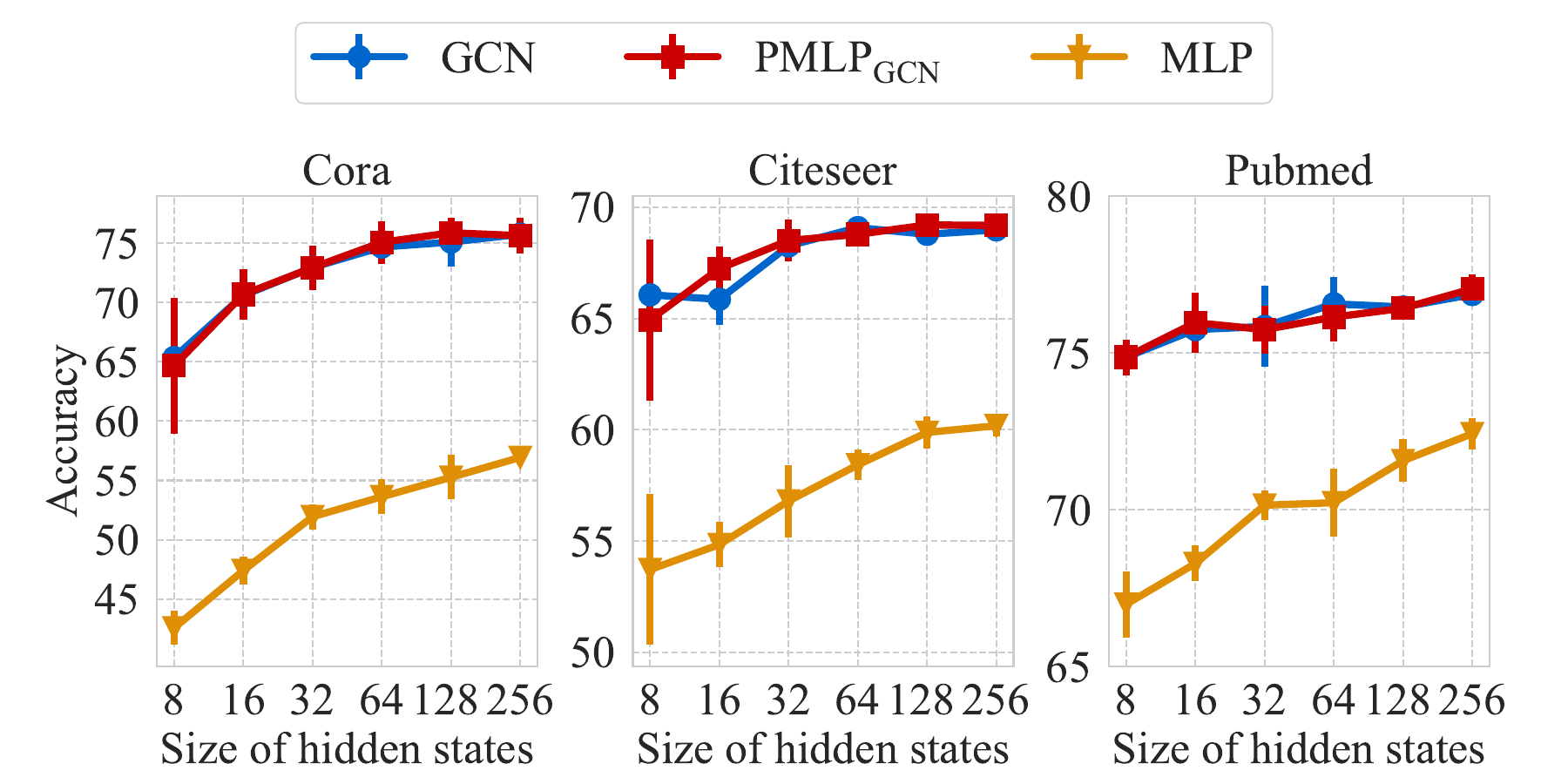}}
\caption{Performance variation with increasing layer number and size of hidden states. (See complete results in Appendix~\ref{app_add}).}
\label{fig_hyper}
\end{figure*}

\paragraph{Q1: What is the impact of model layers and hidden sizes?}
In Fig.~\ref{fig_hyper}(a) and (b), we plot the testing accuracy of GCN, PMLP$_{GCN}$ and MLP w.r.t. different layer numbers and hidden sizes. The results show that the observed phenomenon in Section~\ref{sec-exp-result} consistently holds with different settings of model depth and width, which suggests that the generalization effect of the MP operation is insensitive to model architectural hyperparameters. The increase of layer numbers cause performance degradation for all three models, presumably because of over-fitting. We will further discuss the impact of model depth (where layer number exceeds $100$) and residual connections in the next subsection.

\begin{figure*}[t!]
\centering
\subfigure[Activation functions]{
\label{Fig.hp.1}
\includegraphics[width=0.49\textwidth]{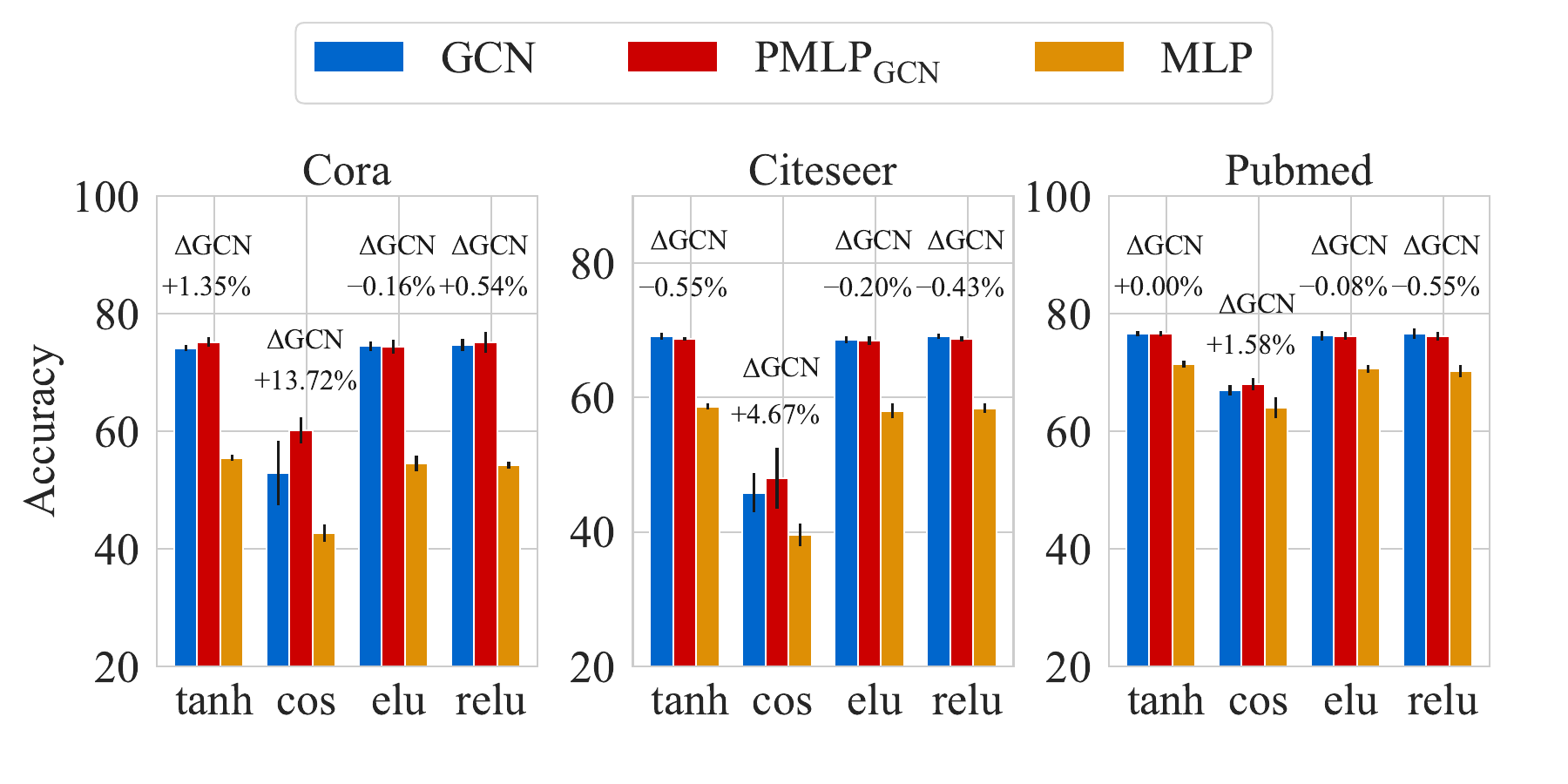}}
\subfigure[Transition matrix]{
\label{Fig.hp.2}
\includegraphics[width=0.49\textwidth]{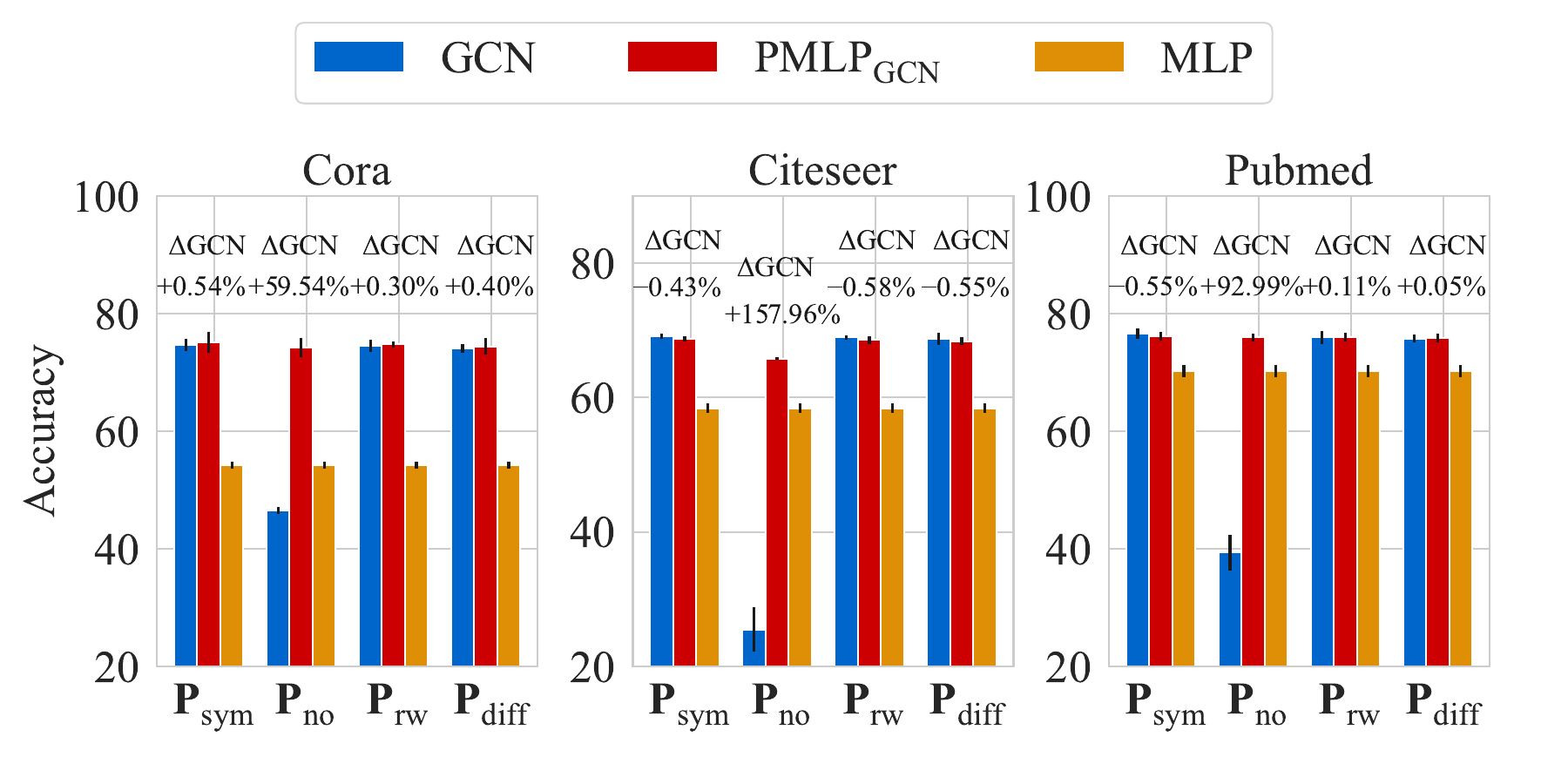}}
\caption{Performance variation with different activation functions in FF layer and different transition matrices in MP layer. (See complete results in Appendix~\ref{app_add})}
\label{fig_model}
\end{figure*}

\begin{figure}[t!]
	\centering
	\includegraphics[width=1.0\textwidth]{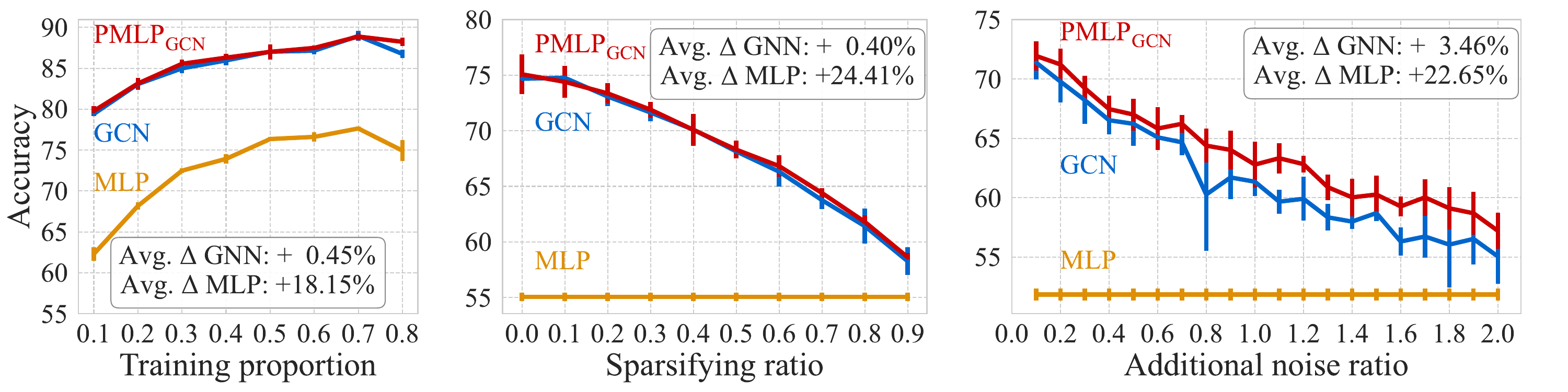}
	\caption{Impact of graph structural information by changing data split, sparsifying the graph, adding random structural noise on Cora. (See complete results in Appendix~\ref{app_add})} \label{fig_info}
\end{figure}

\paragraph{Q2: What is the impact of different activation functions in FF layers?}
As shown in Fig.~\ref{fig_model}(a), the relative performance rankings among GCN, PMLP and MLP stay consistent across four different activation functions (tanh, cos, ELU, ReLU) and in particular, in some cases the performance gain of PMLP over GCN is further amplified (e.g., with cos activation).

\paragraph{Q3: What is the impact of different propagation schemes in MP layers?}
We replace the original transition matrix ${\mathbf P}_{\text{sym}} = \tilde{\mathbf D}^{-\frac{1}{2}}\tilde{\mathbf A}\tilde{\mathbf D}^{-\frac{1}{2}}$ used in the MP layer by other commonly used transition matrices: 1) ${\mathbf P}_{\text{no-loop}} = {\mathbf D}^{-\frac{1}{2}}{\mathbf A}{\mathbf D}^{-\frac{1}{2}}$, i.e., removing self-loop; 2) ${\mathbf P}_{\text{rw}} = \tilde{\mathbf D}^{-1}\tilde{\mathbf A}$, i.e., random walk matrix; 3) ${\mathbf P}_{\text{diff}} = \sum_{k=0}^{\infty}\frac{1}{e\cdot k!}(\tilde{\mathbf D}^{-1}\tilde{\mathbf A})^k$, i.e., heat kernel diffusion matrix. The results are presented in Fig.~\ref{fig_model}(b) where we found that the relative performance rankings of three models keep nearly unchanged after replacing the original MP layer. And, intriguingly, the performance of GNNs degrade dramatically after removing the self-loop, while the accuracy of PMLPs stays at almost the same level. The possible reason is that the self-loop connection plays an important role in GCN's training stage for preserving enough centered nodes' information, but does not affect PMLP.

\paragraph{Q4: What is the impact of training proportion?} 
We use random split to control the amount of graph structure information used in training. As shown in Fig.~\ref{fig_info}(left), the labeled portion of nodes and the amount of training edges have negligible impacts on the relative performance of three models.

\paragraph{Q5: What is the impact of graph sparsity?} 
To further investigate the impact of graph structure, we randomly remove different ratios of edges to reduce graph sparsity when fixing the training proportion. As suggested by Fig.~\ref{fig_info}(middle), when the graph goes sparser, the absolute performance of GCN and PMLP degrades yet their performance gap remains unchanged. This shows that the quality of input graphs indeed impacts the testing performance and tends to control the performance upper bound of the models. Critically though, the generalization effect brought by PMLP is insensitive to the graph completeness.

\paragraph{Q6: What is the impact of noisy structure?} 
As shown in Fig.~\ref{fig_info}(right), the performances of both PMLP and GCN tend to decrease as we gradually add random connections to the graph, whose amount is controlled by the noise ratio (defined as $\mbox{\# noisy edges} / |\mathcal E|$), while PMLP shows better robustness to such noise. This indicates noisy structures have negative effects on both training and generalization of GNNs and one may use PMLP to mitigate this issue.

\subsection{Over-Smoothing, Model Depth, and Heterophily}\label{sec-exp-prob}

We conduct additional experiments as preliminary explorations on broader aspects of GNNs including over-smoothing, model depth and graph heterophily. Detailed results and respective discussions are deferred to Appendix~\ref{app_oversmooth}. In a nutshell, we find the phenomenon still holds for GNNs with residual connections (i.e., JKNet and GCNII), very deep GNNs (with more than $100$ layers), and for heterophilic graphs, PMLPs are outperformed by MLP when their GNN counterparts are inferior than MLP. This indicates that both over-smoothing and heterophily are problems closely related to failure cases of GNN generalization, and can be mitigated by using MP layers that are more suitable for the data or using other backbone models (ResNet) that support deeper layers, aligning with some previous studies~\citep{battaglia2018relational, cong2021provable}.

\subsection{Extension to Transductive Settings}\label{sec-exp-trans}
Notably, the current training procedure of PMLP does not involve unlabeled nodes but can be extended to such scenario by combining with existing semi-supervised learning approaches~\citep{van2020survey} such as using label propagation~\citep{zhu2003semi} to generate pseudo labels for unlabeled nodes. Alternatively, one can further use the GNN architecture in training for fine-tuning the model~\citep{han2022mlpinit}, incorporation of which can improve PMLP while preserving its advantage in training efficiency.

\section{Theoretical Insights on GNN Generalization}\label{sec_theory}
Towards theoretically answering why ``GNNs are inherently good generalizers" and explaining the superior generalization performance of PMLP and GNN, we next compare MLP, PMLP and GNN from the \emph{Neural Tangent Kernel} (NTK) perspective, analyze the effects of model architectures in testing time by examining their generalization behaviors under distribution shifts. Note that our analysis focuses on test-time model architectures, and thus the results presented in Sec.~\ref{subsec2} are applicable for both GNN and PMLP in both inductive and transductive settings.

\subsection{NTK Perspective on MLP, PMLP and GNN} \label{subsec1}

\paragraph{Neural Tangent Kernel.}  
For a neural network $f(\bs x; \bs{\theta}): \mathcal X \rightarrow \mathbb R$ with initial parameters $\bs{\theta}_0$ and a fixed input sample $\bs x$, performing first-order Taylor expansion around $\bs{\theta}_0$ yields the linearized form of NNs~\citep{lee2019wide} as follows:
\begin{equation} \label{eq_linearization}
    {f}^{lin}(\bs x; \bs{\theta})=f(\bs x; \bs{\theta}_0)+\nabla_{\bs{\theta}} f(\bs x; \bs{\theta})^\top\left(\bs{\theta}-\bs{\theta}_0\right),
\end{equation}
where the gradient $\nabla_{\bs{\theta}} f(\bs x; \bs{\theta})$ could be thought of as a feature map $\phi(\bs x): \mathcal X \rightarrow \mathbb R^{|\bs{\theta}|}$ from input space to a kernel feature space. As such, when $\bs{\theta}_0$ is initialized by Gaussian distribution with certain scaling and the network width tends to infinity (i.e., $m \rightarrow \infty$, where $m$ denotes the layer width), the feature map becomes constant and is determined by the model architecture (e.g., MLP, GNN and CNN), inducing a kernel called NTK~\citep{jacot2018neural}:
\begin{equation}
    \operatorname{NTK}(\bs x_i, \bs x_j) = \phi_{ntk}(\bs x_i)^\top \phi_{ntk}(\bs x_j) = \left\langle \nabla_{\bs{\theta}} f(\bs x_i; \bs{\theta}), \nabla_{\bs{\theta}} f(\bs x_j; \bs{\theta}) \right\rangle
\end{equation}

Recent works~\citep{liu2020linearity,liu2020toward} further show that the spectral norm of Hessian matrix in the Taylor series tends to zero with increasing width by $\Theta(1/\sqrt{m})$, and hence the linearization becomes almost exact. Therefore, training an over-parameterized NN using gradient descent with infinitesimal step size is equivalent to kernel regression with NTK~\citep{arora2019exact}:
\begin{equation}
    f(\bs x; \bs w) = \bs w^\top \phi_{ntk}(\bs x), \quad \mathcal{L}(\bs w)=\frac{1}{2} \sum_{i=1}^n\left(y_i-\bs w^\top \phi_{ntk}(\bs x_i )\right)^2.
\end{equation}

We next show the equivalence of MLP and PMLP in training at infinite width limit (NTK regime).

\begin{proposition}\label{prop1} 
MLP and its corresponding PMLP have the same minimum RKHS-norm NTK kernel regression solution, but differ from that of GNNs, i.e., $\bs w^*_{mlp} = \bs w^*_{pmlp} \neq \bs w^*_{gnn}$.
\end{proposition}

\textbf{Implications}. From the NTK perspective, stacking additional message passing layers in the testing phase implies transforming the fixed feature map from that of MLP $\phi_{mlp}(\bs x)$ to that of GNN $\phi_{gnn}(\bs x)$, while fixing $\bs w$. Given $\bs w^*_{mlp} = \bs w^*_{pmlp}$, the superior generalization performance of PMLP (i.e., the key factor of performance surge from MLP to GNN) can be explained by such transformation of feature map from $\phi_{mlp}(\bs x)$ to $\phi_{gnn}(\bs x)$ in testing:
\begin{equation}
\begin{split}
    f_{mlp}(\bs x) = \bs w^{*^\top}_{mlp} \;\phi_{mlp}(\bs x), \quad f_{pmlp}(\bs x) = \bs w^{*^\top}_{mlp} \;\phi_{gnn}(\bs x), \quad f_{gnn}(\bs x) = \bs w^{*^\top}_{gnn} \;\phi_{gnn}(\bs x).
\end{split}
\end{equation}
This perspective simplifies the subsequent theoretical analysis by setting the focus on the difference of feature map, which is determined by the network architecture used in inference, and empirically suggested to be the key factor attributing to superior generalization performance of GNN and PMLP. To step further, we next derive the formula for computing NTK feature map of PMLP and GNN (i.e., $\phi_{gnn}(\bs x)$) in node regression tasks.

\paragraph{GNTK in Node Regression.}
Following previous works that analyse shallow and wide NNs~\citep{arora2019fine,chizat2020implicit,xu2021neural}, we focus on a two-layer GNN using average aggregation with self-connection. By extending the original definition of GNTK~\citep{du2019graph} from graph-level regression to node-level regression as specified in Appendix.~\ref{app_lemma3}, we have the following explicit form of $\phi_{gnn}(\bs x)$ for a two-layer GNN.

\begin{lemma}\label{lemma3} The explicit form of GNTK feature map for a two-layer GNN $\phi_{gnn}(x)$ with average aggregation and ReLU activation in node regression is
\begin{equation}
\phi_{gnn}(\bs x_i) = c \sum_{j \in \mathcal N_i \cup\{i\}}\left[\mathbf{X}^{\top} \bs{a}_j \cdot \mathbb{I}_{+}\left(\bs{w}^{(k)^{\top}} \mathbf{X}^{\top} \bs{a}_j \right), \bs{w}^{(k)^{\top}} \mathbf{X}^{\top} \bs{a}_j \cdot \mathbb{I}_{+}\left(\bs{w}^{(k)^{\top}} \mathbf{X}^{\top} \bs{a}_j \right), \ldots\right],
\end{equation}
where $c=O(\tilde{d}^{-1})$ is a constant proportional to the inverse of node degree, $\mathbf X\in \mathbb R^{n\times d}$ is node features, $\bs a_i\in \mathbb R^{n}$ denotes adjacency vector of node $i$, $\bs{w}^{(k)} \sim \mathcal{N}(\mathbf{0}, \bs{I}_d)$ is a random Gaussian vector in $\mathbb R^d$, two components in the brackets repeat infinitely many times with $k$ ranging from $1$ to $\infty$, and $\mathbb{I}_{+}$ is an indicator function that outputs $1$ if the input is positive otherwise $0$.
\end{lemma}

\subsection{MLP v.s. PMLP in Extrapolation} \label{subsec2}
As indicated by Proposition~\ref{prop1}, the fundamental difference between MLP and PMLP at infinite width limit stems from the difference of feature map in the testing phase. This reduces the problem of explaining the success of GNN to the question that why this change is significant for generalizability.

\paragraph{Extrapolation Behavior of MLP.} One important aspect of generalization analysis is regarding model's behavior when confronted with OoD testing samples (i.e., testing nodes that are considerably outside the training support), a.k.a. \emph{extrapolation analysis}. A previous study on this direction~\citep{xu2021neural} reveal that a standard MLP with ReLU activation quickly converges to a linear function as the testing sample escapes the training support, which is formalized by the following theorem.

\begin{theorem}\label{thm_hownn}\textbf{\citep{xu2021neural}}. Suppose $f_{mlp}(\bs x)$ is an infinitely-wide two-layer MLP with ReLU trained by square loss. For any direction $\bs{v} \in \mathbb{R}^d$ and step size $\Delta t>0$, let $\bs{x}_0 = t \bs{v}$, we have
\begin{equation} \label{eq_thm3}
    \left|\frac{\left(f_{mlp}(\bs{x}_0+\Delta t \bs{v})-f_{mlp}(\bs{x}_0)\right)/\Delta t}{c_{\bs{v}}} - 1\right|= O(\frac{1}{t}).
\end{equation}
where $c_{\bs{v}}$ is a constant linear coefficient. That is, as $t\rightarrow \infty$, $f_{mlp}(\bs x_0)$ converges to a linear function.
\end{theorem}

The intuition behind this phenomenon is the fact that ReLU MLPs learn piece-wise linear functions with finitely many linear regions and thus eventually becomes linear outside training data support. Now a naturally arising question is how does PMLP compare with MLP regarding extrapolation? 

\paragraph{Extrapolation Behavior of PMLP.} 
Based on the explicit formula for $\phi_{gnn}(\bs x)$ in Lemma~\ref{lemma3}, we extend the theoretical result of extrapolation analysis from MLP to PMLP. Our first finding is that, as the testing node feature becomes increasingly outside the range of training data, alike MLP, PMLP (as well as GNN) with average aggregation also converges to a linear function, yet the corresponding linear coefficient reflects its ego-graph property rather than being a fixed constant.

\begin{theorem}\label{thm4} Suppose $f_{pmlp}(\bs x)$ is an infinitely-wide two-layer MLP with ReLU activation trained using squared loss, and adds average message passing layer before each feed-forward layer in the testing phase. For any direction $\bs{v} \in \mathbb{R}^d$ and step size $\Delta t>0$, let $\bs{x}_0 = t \bs{v}$, and as $t\rightarrow \infty$, we have
\begin{equation}
    \left(f_{pmlp}(\bs{x}_0+\Delta t \bs{v})-f_{pmlp}(\bs{x}_0)\right)/\Delta t \quad \rightarrow \quad c_{\bs v}\sum_{i\in \mathcal N_{0} \cup \{0\}} ({\tilde d}\cdot\tilde{d}_i)^{-1}.
\end{equation}
where $c_{\bs{v}}$ is the same constant as in Theorem~\ref{thm_hownn}, $\tilde d_0 = \tilde d$ is the node degree (with self-connection) of $\bs x_0$, and $\tilde d_i$ is the node degree of its neighbors.
\end{theorem}

\textbf{Remark}. This result also applies to infinitely-wide two-layer GNN with ReLU activation and average aggregation in node regression settings, except the constant $c_{\bs v}$ is different from that of MLP. 

\paragraph{Convergence Comparison.} Though both MLP and PMLP tend to linearize for outlier testing samples, indicating that they have common difficulty to extrapolate non-linear functions, we find that PMLP in general has more freedom to deviate from the convergent linear coefficient, implying smoother transition from in-distribution (non-linear) to out-of-distribution (linear) regime and thus could potentially generalize to out-of-distribution samples near the range of training data.

\begin{theorem}\label{thm5} Suppose all node features are normalized, and the cosine similarity of node $\bs x_i$ and the average of its neighbors is deonoted as $\alpha_i \in [0,1]$. Then, the convergence rate for $f_{pmlp}(\bs x)$ is
\begin{equation} \label{eq_thm5}
    \left|\frac{\left(f_{pmlp}(\bs{x}_0+\Delta t \bs{v})-f_{pmlp}(\bs{x}_0)\right)/\Delta t}{c_{\bs v}\sum_{i\in \mathcal N_{0} \cup \{0\}} ({\tilde d}\cdot\tilde{d}_i)^{-1}} - 1\right| = O\left(\frac{1+(\tilde{d}_{max} -1) \sqrt{1-\alpha_{min}^2}}{ t}\right).
\end{equation}
where $\alpha_{min} = \operatorname{min}\{\alpha_i\}_{i\in \mathcal N_0 \cup \{0\}} \in [0,1]$, and $\tilde{d}_{max}\geq 1$ denotes the maximum node degree in the testing node $\bs x_0$'s neighbors (including itself).
\end{theorem}

This result indicates larger node degree and feature dissimilarity imply smoother transition and better compatibility with OoD samples. Specifically, when the testing node's degree is $1$ (connection to itself), PMLP becomes equivalent to MLP. As reflection in Eq.~\ref{eq_thm5}, $\tilde{d}_{max} = 1$, and the bound degrades to that of MLP in Eq.~\ref{eq_thm3}. Moreover, when all node features are equal, message passing will become meaningless. Correspondingly, $\tilde{\alpha}_{min} = 1$, and the bound also degrades to that of MLP.

\section{Related Works}

\paragraph{Generalization Analysis for GNNs.} Generalization, especially for feed-forward NNs (i.e., MLPs), has been extensively studied in the general ML field~\citep{arora2019fine,allen2019learning,cao2019generalization}. However for GNNs, the large body of existing theoretical works focus on their representational power(e.g., ~\cite{scarselli2008computational,xu2018powerful,maron2019provably,oono2019graph}), while their generalization capability is less well-understood. For node-level prediction setting, those works in generalization analysis~\citep{scarselli2018vapnik,verma2019stability,baranwal2021graph,ma2021subgroup} mainly aim to derive generalization bounds, but did not establish connections with MLPs since they assume the same GNN architecture in training and testing. For theoretical analysis, the most relevant work is~\citep{xu2021neural} that studies the extrapolation behavior of MLP. Part of their results are used in this work. The authors also shed lights on the extrapolation power of GNNs, but for graph-level prediction with max/min propagation from the perspective of algorithmic alignment, which cannot apply to GNNs with average/sum propagation in node-level prediction that are more commonly used. 

\paragraph{Connection between MLP and GNN.} Regarding the relation between MLPs and GNNs, there are some recent attempts to boost the performance of MLP to approach that of GNN, by using label propagation~\citep{huang2020combining}, contrastive learning~\citep{hu2021graph}, knowledge distillation~\citep{zhang2021graph} or additional regularization in training~\citep{zhang2023orthoreg}. However, it is unclear whether these graph-enhanced MLPs can explain the success of GNNs since it is still an open research question to understand these training techniques themselves. There are also few works probing into similar model architectures as PMLP, e.g.~\cite{klicpera2018predict}, which can generally be seen as special cases of PMLP. A concurrent work~\citep{han2022mlpinit} further finds that PMLP with an additional fine-tuning procedure can be used to significantly accelerate GNN training. Moreover, a recent work~\citep{baranwal2023effects} also theoretically studies how message passing operations benefit multi-layer networks in node classification tasks, which complements our results.

\paragraph{Efficient GNN Training and Inference.} GNNs usually face the challenge of scalability due to their reliance on graph structures. Many approaches~\citep{spielman2011spectral,calandriello2018improved,chen2018fastgcn,rong2019dropedge,cai2021graph,yang2022geometric} aim to tackle such a problem by sparsifying the graph structure. For example, \citet{yang2022geometric} resorts to geometric knowledge distillation for efficient inference on smaller graphs; \citet{rong2019dropedge} proposes to drop edges in training which can often improve training efficiency. In the context of this line of works, PMLPs can be treated as GNNs with all edges except self-loops dropped in training. Another line of works develop GNN models where message passing operations are used to augment node features~\cite{wu2019simplifying,chen2019powerful,rossi2020sign,chen2020graph}. Since such a process can be performed before training, they could have the same level of efficiency as their PMLP counterparts. Our empirical findings further suggest that the inner workings of these GNNs can also be attributed to the generalization effects of message passing operations.

\section{More Discussions and Conclusion}\label{sec_conclu}

\paragraph{Other sources of performance gap between MLP and GNN / When PMLP fails?} Besides the intrinsic generalizability of GNNs that is revealed by the performance gain from MLP to PMLP in this work, we note that there are some other less significant but non-negligible sources that attributes to the performance gap between GNN and MLP in node prediction tasks:

    $\bullet$\; \textbf{Expressiveness:} While our experiments find that GNNs and PMLPs can perform similarly in most cases, showing great advantage over MLPs in generalization, in practice, there still exists a certain gap between their expressiveness, which can be amplified in large datasets and causes certain degrees of performance difference. This is reflected by our experiments on three large-scale datasets. Despite that, we can see from Table~\ref{tab_large} that the intrinsic generalizability of GNN (corresponding to $\Delta_{mlp}$) is still the major source of performance gain from MLP.
    
    $\bullet$\; \textbf{Semi-Supervised / Transductive Learning:} As our default experimental setting, inductive learning ensures that testing samples are unobserved during training and keeps the comparison among models fair. However in practice, the ability to leverage the information of unlabeled nodes in training is a well-known advantage of GNN (but not the advantage of PMLP). Still, PMLP can be used in transductive setting with training techniques as described in Sec.~\ref{sec-exp-trans}.

\paragraph{Current Limitations and Outlooks.} The result in Theorem~\ref{thm5} provides a bound to show PMLP's better potential in OoD generalization, rather than guaranteeing its superior generalization capability. Explaining when and why PMLPs perform closely to their GNN counterparts also need further investigations. Moreover, following most theoretical works on NTK, we consider the regression task with squared loss for analysis instead of classification. However, as evidences~\citep{janocha2017loss,hui2020evaluation} show squared loss can be as competitive as softmax cross-entropy loss, the insights obtained from regression tasks could also adapt to classification tasks.

\paragraph{Conclusion.}
In this work, we bridge MLP and GNN by introducing an intermediate model class called PMLP, which is equivalent to MLP in training, but shows significantly better generalization performance after adding unlearned message passing layers in testing and can rival with its GNN counterpart in most cases. This phenomenon is consistent across different datasets and experimental settings. To shed some lights on this phenomenon, we show despite that both MLP and PMLP cannot extrapolate non-linear functions, PMLP converges slower, indicating smoother transition and better tolerance for out-of-distribution samples.

\clearpage
\bibliography{ref}

\begin{thebibliography}{69}
\providecommand{\natexlab}[1]{#1}
\providecommand{\url}[1]{\texttt{#1}}
\expandafter\ifx\csname urlstyle\endcsname\relax
  \providecommand{\doi}[1]{doi: #1}\else
  \providecommand{\doi}{doi: \begingroup \urlstyle{rm}\Url}\fi

\bibitem[Allen-Zhu et~al.(2019)Allen-Zhu, Li, and Liang]{allen2019learning}
Zeyuan Allen-Zhu, Yuanzhi Li, and Yingyu Liang.
\newblock Learning and generalization in overparameterized neural networks,
  going beyond two layers.
\newblock \emph{Advances in neural information processing systems}, 32, 2019.

\bibitem[Arora et~al.(2019{\natexlab{a}})Arora, Du, Hu, Li, and
  Wang]{arora2019fine}
Sanjeev Arora, Simon Du, Wei Hu, Zhiyuan Li, and Ruosong Wang.
\newblock Fine-grained analysis of optimization and generalization for
  overparameterized two-layer neural networks.
\newblock In \emph{International Conference on Machine Learning}, pp.\
  322--332. PMLR, 2019{\natexlab{a}}.

\bibitem[Arora et~al.(2019{\natexlab{b}})Arora, Du, Hu, Li, Salakhutdinov, and
  Wang]{arora2019exact}
Sanjeev Arora, Simon~S Du, Wei Hu, Zhiyuan Li, Russ~R Salakhutdinov, and
  Ruosong Wang.
\newblock On exact computation with an infinitely wide neural net.
\newblock \emph{Advances in Neural Information Processing Systems}, 32,
  2019{\natexlab{b}}.

\bibitem[Baranwal et~al.(2021)Baranwal, Fountoulakis, and
  Jagannath]{baranwal2021graph}
Aseem Baranwal, Kimon Fountoulakis, and Aukosh Jagannath.
\newblock Graph convolution for semi-supervised classification: Improved linear
  separability and out-of-distribution generalization.
\newblock \emph{arXiv preprint arXiv:2102.06966}, 2021.

\bibitem[Baranwal et~al.(2023)Baranwal, Fountoulakis, and
  Jagannath]{baranwal2023effects}
Aseem Baranwal, Kimon Fountoulakis, and Aukosh Jagannath.
\newblock Effects of graph convolutions in multi-layer networks.
\newblock In \emph{International Conference on Learning Representations}, 2023.

\bibitem[Battaglia et~al.(2018)Battaglia, Hamrick, Bapst, Sanchez-Gonzalez,
  Zambaldi, Malinowski, Tacchetti, Raposo, Santoro, Faulkner,
  et~al.]{battaglia2018relational}
Peter~W Battaglia, Jessica~B Hamrick, Victor Bapst, Alvaro Sanchez-Gonzalez,
  Vinicius Zambaldi, Mateusz Malinowski, Andrea Tacchetti, David Raposo, Adam
  Santoro, Ryan Faulkner, et~al.
\newblock Relational inductive biases, deep learning, and graph networks.
\newblock \emph{arXiv preprint arXiv:1806.01261}, 2018.

\bibitem[Bietti \& Mairal(2019)Bietti and Mairal]{bietti2019inductive}
Alberto Bietti and Julien Mairal.
\newblock On the inductive bias of neural tangent kernels.
\newblock \emph{Advances in Neural Information Processing Systems}, 32, 2019.

\bibitem[Bruna et~al.(2014)Bruna, Zaremba, Szlam, and LeCun]{bruna2014spectral}
Joan Bruna, Wojciech Zaremba, Arthur Szlam, and Yann LeCun.
\newblock Spectral networks and locally connected networks on graphs.
\newblock \emph{ICLR}, 2014.

\bibitem[Cai et~al.(2021)Cai, Wang, and Wang]{cai2021graph}
Chen Cai, Dingkang Wang, and Yusu Wang.
\newblock Graph coarsening with neural networks.
\newblock \emph{International Conference on Learning Representations}, 2021.

\bibitem[Calandriello et~al.(2018)Calandriello, Lazaric, Koutis, and
  Valko]{calandriello2018improved}
Daniele Calandriello, Alessandro Lazaric, Ioannis Koutis, and Michal Valko.
\newblock Improved large-scale graph learning through ridge spectral
  sparsification.
\newblock In \emph{International Conference on Machine Learning}, pp.\
  688--697. PMLR, 2018.

\bibitem[Cao \& Gu(2019)Cao and Gu]{cao2019generalization}
Yuan Cao and Quanquan Gu.
\newblock Generalization bounds of stochastic gradient descent for wide and
  deep neural networks.
\newblock \emph{Advances in neural information processing systems}, 32, 2019.

\bibitem[Chen et~al.(2018)Chen, Ma, and Xiao]{chen2018fastgcn}
Jie Chen, Tengfei Ma, and Cao Xiao.
\newblock Fastgcn: fast learning with graph convolutional networks via
  importance sampling.
\newblock \emph{International Conference on Learning Representations}, 2018.

\bibitem[Chen et~al.(2021)Chen, Chen, and Bruna]{chen2020graph}
Lei Chen, Zhengdao Chen, and Joan Bruna.
\newblock On graph neural networks versus graph-augmented mlps.
\newblock \emph{International Conference on Learning Representations}, 2021.

\bibitem[Chen et~al.(2020{\natexlab{a}})Chen, Wei, Ding, Li, Yuan, Du, and
  Wen]{chen2020scalable}
Ming Chen, Zhewei Wei, Bolin Ding, Yaliang Li, Ye~Yuan, Xiaoyong Du, and
  Ji-Rong Wen.
\newblock Scalable graph neural networks via bidirectional propagation.
\newblock \emph{Advances in neural information processing systems},
  33:\penalty0 14556--14566, 2020{\natexlab{a}}.

\bibitem[Chen et~al.(2020{\natexlab{b}})Chen, Wei, Huang, Ding, and
  Li]{chen2020simple}
Ming Chen, Zhewei Wei, Zengfeng Huang, Bolin Ding, and Yaliang Li.
\newblock Simple and deep graph convolutional networks.
\newblock In \emph{International Conference on Machine Learning}, pp.\
  1725--1735. PMLR, 2020{\natexlab{b}}.

\bibitem[Chen et~al.(2019)Chen, Bian, and Sun]{chen2019powerful}
Ting Chen, Song Bian, and Yizhou Sun.
\newblock Are powerful graph neural nets necessary? a dissection on graph
  classification.
\newblock \emph{arXiv preprint arXiv:1905.04579}, 2019.

\bibitem[Chien et~al.(2020)Chien, Peng, Li, and Milenkovic]{chien2020adaptive}
Eli Chien, Jianhao Peng, Pan Li, and Olgica Milenkovic.
\newblock Adaptive universal generalized pagerank graph neural network.
\newblock \emph{arXiv preprint arXiv:2006.07988}, 2020.

\bibitem[Chizat \& Bach(2020)Chizat and Bach]{chizat2020implicit}
Lenaic Chizat and Francis Bach.
\newblock Implicit bias of gradient descent for wide two-layer neural networks
  trained with the logistic loss.
\newblock In \emph{Conference on Learning Theory}, pp.\  1305--1338. PMLR,
  2020.

\bibitem[Cong et~al.(2021)Cong, Ramezani, and Mahdavi]{cong2021provable}
Weilin Cong, Morteza Ramezani, and Mehrdad Mahdavi.
\newblock On provable benefits of depth in training graph convolutional
  networks.
\newblock \emph{Advances in Neural Information Processing Systems},
  34:\penalty0 9936--9949, 2021.

\bibitem[Du et~al.(2019)Du, Hou, Salakhutdinov, Poczos, Wang, and
  Xu]{du2019graph}
Simon~S Du, Kangcheng Hou, Russ~R Salakhutdinov, Barnabas Poczos, Ruosong Wang,
  and Keyulu Xu.
\newblock Graph neural tangent kernel: Fusing graph neural networks with graph
  kernels.
\newblock \emph{Advances in neural information processing systems}, 32, 2019.

\bibitem[Frasca et~al.(2020)Frasca, Rossi, Chamberlain, Eynard, Bronstein, and
  Monti]{rossi2020sign}
Fabrizio Frasca, Emanuele Rossi, Ben Chamberlain, Davide Eynard, Michael
  Bronstein, and Federico Monti.
\newblock Sign: Scalable inception graph neural networks.
\newblock \emph{arXiv preprint arXiv:2004.11198}, 7:\penalty0 15, 2020.

\bibitem[Gilmer et~al.(2017)Gilmer, Schoenholz, Riley, Vinyals, and
  Dahl]{gilmer2017neural}
Justin Gilmer, Samuel~S Schoenholz, Patrick~F Riley, Oriol Vinyals, and
  George~E Dahl.
\newblock Neural message passing for quantum chemistry.
\newblock In \emph{International conference on machine learning}, pp.\
  1263--1272. PMLR, 2017.

\bibitem[Hamilton et~al.(2017)Hamilton, Ying, and
  Leskovec]{hamilton2017inductive}
Will Hamilton, Zhitao Ying, and Jure Leskovec.
\newblock Inductive representation learning on large graphs.
\newblock \emph{Advances in neural information processing systems}, 30, 2017.

\bibitem[Han et~al.(2023)Han, Zhao, Liu, Hu, and Shah]{han2022mlpinit}
Xiaotian Han, Tong Zhao, Yozen Liu, Xia Hu, and Neil Shah.
\newblock Mlpinit: Embarrassingly simple gnn training acceleration with mlp
  initialization.
\newblock In \emph{International Conference on Learning Representations}, 2023.

\bibitem[Hu et~al.(2020)Hu, Fey, Zitnik, Dong, Ren, Liu, Catasta, and
  Leskovec]{hu2020open}
Weihua Hu, Matthias Fey, Marinka Zitnik, Yuxiao Dong, Hongyu Ren, Bowen Liu,
  Michele Catasta, and Jure Leskovec.
\newblock Open graph benchmark: Datasets for machine learning on graphs.
\newblock \emph{NeurIPS}, 33:\penalty0 22118--22133, 2020.

\bibitem[Hu et~al.(2021)Hu, You, Wang, Wang, Zhou, and Gao]{hu2021graph}
Yang Hu, Haoxuan You, Zhecan Wang, Zhicheng Wang, Erjin Zhou, and Yue Gao.
\newblock Graph-mlp: node classification without message passing in graph.
\newblock \emph{arXiv preprint arXiv:2106.04051}, 2021.

\bibitem[Huang et~al.(2021)Huang, He, Singh, Lim, and
  Benson]{huang2020combining}
Qian Huang, Horace He, Abhay Singh, Ser-Nam Lim, and Austin~R Benson.
\newblock Combining label propagation and simple models out-performs graph
  neural networks.
\newblock \emph{International Conference on Learning Representations}, 2021.

\bibitem[Hui \& Belkin(2020)Hui and Belkin]{hui2020evaluation}
Like Hui and Mikhail Belkin.
\newblock Evaluation of neural architectures trained with square loss vs
  cross-entropy in classification tasks.
\newblock \emph{arXiv preprint arXiv:2006.07322}, 2020.

\bibitem[Jacot et~al.(2018)Jacot, Gabriel, and Hongler]{jacot2018neural}
Arthur Jacot, Franck Gabriel, and Cl{\'e}ment Hongler.
\newblock Neural tangent kernel: Convergence and generalization in neural
  networks.
\newblock \emph{Advances in neural information processing systems}, 31, 2018.

\bibitem[Janocha \& Czarnecki(2017)Janocha and Czarnecki]{janocha2017loss}
Katarzyna Janocha and Wojciech~Marian Czarnecki.
\newblock On loss functions for deep neural networks in classification.
\newblock \emph{arXiv preprint arXiv:1702.05659}, 2017.

\bibitem[Kipf \& Welling(2017)Kipf and Welling]{kipf2017semi}
Thomas~N Kipf and Max Welling.
\newblock Semi-supervised classification with graph convolutional networks.
\newblock \emph{ICLR}, 2017.

\bibitem[Klicpera et~al.(2019)Klicpera, Bojchevski, and
  G{\"u}nnemann]{klicpera2018predict}
Johannes Klicpera, Aleksandar Bojchevski, and Stephan G{\"u}nnemann.
\newblock Predict then propagate: Graph neural networks meet personalized
  pagerank.
\newblock \emph{ICLR}, 2019.

\bibitem[Lee et~al.(2019)Lee, Xiao, Schoenholz, Bahri, Novak, Sohl-Dickstein,
  and Pennington]{lee2019wide}
Jaehoon Lee, Lechao Xiao, Samuel Schoenholz, Yasaman Bahri, Roman Novak, Jascha
  Sohl-Dickstein, and Jeffrey Pennington.
\newblock Wide neural networks of any depth evolve as linear models under
  gradient descent.
\newblock \emph{Advances in neural information processing systems}, 32, 2019.

\bibitem[Liu et~al.(2020{\natexlab{a}})Liu, Zhu, and Belkin]{liu2020toward}
Chaoyue Liu, Libin Zhu, and Mikhail Belkin.
\newblock Toward a theory of optimization for over-parameterized systems of
  non-linear equations: the lessons of deep learning.
\newblock \emph{arXiv preprint arXiv:2003.00307}, 2020{\natexlab{a}}.

\bibitem[Liu et~al.(2020{\natexlab{b}})Liu, Zhu, and Belkin]{liu2020linearity}
Chaoyue Liu, Libin Zhu, and Misha Belkin.
\newblock On the linearity of large non-linear models: when and why the tangent
  kernel is constant.
\newblock \emph{Advances in Neural Information Processing Systems},
  33:\penalty0 15954--15964, 2020{\natexlab{b}}.

\bibitem[Liu et~al.(2020{\natexlab{c}})Liu, Gao, and Ji]{liu2020towards}
Meng Liu, Hongyang Gao, and Shuiwang Ji.
\newblock Towards deeper graph neural networks.
\newblock In \emph{Proceedings of the 26th ACM SIGKDD international conference
  on knowledge discovery \& data mining}, pp.\  338--348, 2020{\natexlab{c}}.

\bibitem[Ma et~al.(2021)Ma, Deng, and Mei]{ma2021subgroup}
Jiaqi Ma, Junwei Deng, and Qiaozhu Mei.
\newblock Subgroup generalization and fairness of graph neural networks.
\newblock \emph{Advances in Neural Information Processing Systems},
  34:\penalty0 1048--1061, 2021.

\bibitem[Maron et~al.(2019)Maron, Ben-Hamu, Serviansky, and
  Lipman]{maron2019provably}
Haggai Maron, Heli Ben-Hamu, Hadar Serviansky, and Yaron Lipman.
\newblock Provably powerful graph networks.
\newblock \emph{Advances in neural information processing systems}, 32, 2019.

\bibitem[McAuley et~al.(2015)McAuley, Targett, Shi, and Van
  Den~Hengel]{amazon-dataset}
Julian McAuley, Christopher Targett, Qinfeng Shi, and Anton Van Den~Hengel.
\newblock Image-based recommendations on styles and substitutes.
\newblock In \emph{Proceedings of the 38th international ACM SIGIR conference
  on research and development in information retrieval}, pp.\  43--52, 2015.

\bibitem[McCallum et~al.(2000)McCallum, Nigam, Rennie, and
  Seymore]{mccallum2000automating}
Andrew~Kachites McCallum, Kamal Nigam, Jason Rennie, and Kristie Seymore.
\newblock Automating the construction of internet portals with machine
  learning.
\newblock \emph{Information Retrieval}, 2000.

\bibitem[Namata et~al.(2012)Namata, London, Getoor, Huang, and
  EDU]{namata2012query}
Galileo Namata, Ben London, Lise Getoor, Bert Huang, and UMD EDU.
\newblock Query-driven active surveying for collective classification.
\newblock In \emph{International Workshop on Mining and Learning with Graphs},
  2012.

\bibitem[Oono \& Suzuki(2019)Oono and Suzuki]{oono2019graph}
Kenta Oono and Taiji Suzuki.
\newblock Graph neural networks exponentially lose expressive power for node
  classification.
\newblock \emph{arXiv preprint arXiv:1905.10947}, 2019.

\bibitem[Pei et~al.(2020)Pei, Wei, Chang, Lei, and Yang]{pei2020geom}
Hongbin Pei, Bingzhe Wei, Kevin Chen-Chuan Chang, Yu~Lei, and Bo~Yang.
\newblock Geom-gcn: Geometric graph convolutional networks.
\newblock \emph{International Conference on Learning Representations}, 2020.

\bibitem[Rong et~al.(2020)Rong, Huang, Xu, and Huang]{rong2019dropedge}
Yu~Rong, Wenbing Huang, Tingyang Xu, and Junzhou Huang.
\newblock Dropedge: Towards deep graph convolutional networks on node
  classification.
\newblock \emph{International Conference on Learning Representations}, 2020.

\bibitem[Rumelhart et~al.(1986)Rumelhart, Hinton, and
  Williams]{rumelhart1986learning}
David~E Rumelhart, Geoffrey~E Hinton, and Ronald~J Williams.
\newblock Learning representations by back-propagating errors.
\newblock \emph{nature}, 323\penalty0 (6088):\penalty0 533--536, 1986.

\bibitem[Scarselli et~al.(2008{\natexlab{a}})Scarselli, Gori, Tsoi,
  Hagenbuchner, and Monfardini]{scarselli2008computational}
Franco Scarselli, Marco Gori, Ah~Chung Tsoi, Markus Hagenbuchner, and Gabriele
  Monfardini.
\newblock Computational capabilities of graph neural networks.
\newblock \emph{IEEE Transactions on Neural Networks}, 20\penalty0
  (1):\penalty0 81--102, 2008{\natexlab{a}}.

\bibitem[Scarselli et~al.(2008{\natexlab{b}})Scarselli, Gori, Tsoi,
  Hagenbuchner, and Monfardini]{scarselli2008graph}
Franco Scarselli, Marco Gori, Ah~Chung Tsoi, Markus Hagenbuchner, and Gabriele
  Monfardini.
\newblock The graph neural network model.
\newblock \emph{IEEE transactions on neural networks}, 20\penalty0
  (1):\penalty0 61--80, 2008{\natexlab{b}}.

\bibitem[Scarselli et~al.(2018)Scarselli, Tsoi, and
  Hagenbuchner]{scarselli2018vapnik}
Franco Scarselli, Ah~Chung Tsoi, and Markus Hagenbuchner.
\newblock The vapnik--chervonenkis dimension of graph and recursive neural
  networks.
\newblock \emph{Neural Networks}, 108:\penalty0 248--259, 2018.

\bibitem[Sen et~al.(2008)Sen, Namata, Bilgic, Getoor, Galligher, and
  Eliassi-Rad]{sen2008collective}
Prithviraj Sen, Galileo Namata, Mustafa Bilgic, Lise Getoor, Brian Galligher,
  and Tina Eliassi-Rad.
\newblock Collective classification in network data.
\newblock \emph{AI magazine}, 2008.

\bibitem[Shchur et~al.(2018)Shchur, Mumme, Bojchevski, and
  G{\"u}nnemann]{shchur2018pitfalls}
Oleksandr Shchur, Maximilian Mumme, Aleksandar Bojchevski, and Stephan
  G{\"u}nnemann.
\newblock Pitfalls of graph neural network evaluation.
\newblock \emph{arXiv preprint arXiv:1811.05868}, 2018.

\bibitem[Sinha et~al.(2015)Sinha, Shen, Song, Ma, Eide, Hsu, and
  Wang]{coauthor-dataset}
Arnab Sinha, Zhihong Shen, Yang Song, Hao Ma, Darrin Eide, Bo-June Hsu, and
  Kuansan Wang.
\newblock An overview of microsoft academic service (mas) and applications.
\newblock In \emph{Proceedings of the 24th international conference on world
  wide web}, pp.\  243--246, 2015.

\bibitem[Spielman \& Teng(2011)Spielman and Teng]{spielman2011spectral}
Daniel~A Spielman and Shang-Hua Teng.
\newblock Spectral sparsification of graphs.
\newblock \emph{SIAM Journal on Computing}, 40\penalty0 (4):\penalty0
  981--1025, 2011.

\bibitem[Spinelli et~al.(2020)Spinelli, Scardapane, and
  Uncini]{spinelli2020adaptive}
Indro Spinelli, Simone Scardapane, and Aurelio Uncini.
\newblock Adaptive propagation graph convolutional network.
\newblock \emph{IEEE Transactions on Neural Networks and Learning Systems},
  32\penalty0 (10):\penalty0 4755--4760, 2020.

\bibitem[Van~Engelen \& Hoos(2020)Van~Engelen and Hoos]{van2020survey}
Jesper~E Van~Engelen and Holger~H Hoos.
\newblock A survey on semi-supervised learning.
\newblock \emph{Machine learning}, 109\penalty0 (2):\penalty0 373--440, 2020.

\bibitem[Veli{\v{c}}kovi{\'c} et~al.(2017)Veli{\v{c}}kovi{\'c}, Cucurull,
  Casanova, Romero, Lio, and Bengio]{velivckovic2017graph}
Petar Veli{\v{c}}kovi{\'c}, Guillem Cucurull, Arantxa Casanova, Adriana Romero,
  Pietro Lio, and Yoshua Bengio.
\newblock Graph attention networks.
\newblock \emph{arXiv preprint arXiv:1710.10903}, 2017.

\bibitem[Verma \& Zhang(2019)Verma and Zhang]{verma2019stability}
Saurabh Verma and Zhi-Li Zhang.
\newblock Stability and generalization of graph convolutional neural networks.
\newblock In \emph{Proceedings of the 25th ACM SIGKDD International Conference
  on Knowledge Discovery \& Data Mining}, pp.\  1539--1548, 2019.

\bibitem[Wu et~al.(2019)Wu, Souza, Zhang, Fifty, Yu, and
  Weinberger]{wu2019simplifying}
Felix Wu, Amauri Souza, Tianyi Zhang, Christopher Fifty, Tao Yu, and Kilian
  Weinberger.
\newblock Simplifying graph convolutional networks.
\newblock In \emph{International conference on machine learning}, pp.\
  6861--6871. PMLR, 2019.

\bibitem[Wu et~al.(2022)Wu, Zhao, Li, Wipf, and Yan]{wu2022nodeformer}
Qitian Wu, Wentao Zhao, Zenan Li, David Wipf, and Junchi Yan.
\newblock Nodeformer: A scalable graph structure learning transformer for node
  classification.
\newblock In \emph{Advances in Neural Information Processing Systems
  (NeurIPS)}, 2022.

\bibitem[Wu et~al.(2023)Wu, Yang, Zhao, He, Wipf, and Yan]{wu2023difformer}
Qitian Wu, Chenxiao Yang, Wentao Zhao, Yixuan He, David Wipf, and Junchi Yan.
\newblock Difformer: Scalable (graph) transformers induced by energy
  constrained diffusion.
\newblock \emph{ICLR}, 2023.

\bibitem[Xu et~al.(2018{\natexlab{a}})Xu, Hu, Leskovec, and
  Jegelka]{xu2018powerful}
Keyulu Xu, Weihua Hu, Jure Leskovec, and Stefanie Jegelka.
\newblock How powerful are graph neural networks?
\newblock \emph{arXiv preprint arXiv:1810.00826}, 2018{\natexlab{a}}.

\bibitem[Xu et~al.(2018{\natexlab{b}})Xu, Li, Tian, Sonobe, Kawarabayashi, and
  Jegelka]{xu2018representation}
Keyulu Xu, Chengtao Li, Yonglong Tian, Tomohiro Sonobe, Ken-ichi Kawarabayashi,
  and Stefanie Jegelka.
\newblock Representation learning on graphs with jumping knowledge networks.
\newblock In \emph{International conference on machine learning}, pp.\
  5453--5462. PMLR, 2018{\natexlab{b}}.

\bibitem[Xu et~al.(2021)Xu, Zhang, Li, Du, Kawarabayashi, and
  Jegelka]{xu2021neural}
Keyulu Xu, Mozhi Zhang, Jingling Li, Simon~S Du, Ken-ichi Kawarabayashi, and
  Stefanie Jegelka.
\newblock How neural networks extrapolate: From feedforward to graph neural
  networks.
\newblock \emph{ICLR}, 2021.

\bibitem[Yang et~al.(2022)Yang, Wu, and Yan]{yang2022geometric}
Chenxiao Yang, Qitian Wu, and Junchi Yan.
\newblock Geometric knowledge distillation: Topology compression for graph
  neural networks.
\newblock 2022.

\bibitem[Zeng et~al.(2019)Zeng, Zhou, Srivastava, Kannan, and
  Prasanna]{zeng2019graphsaint}
Hanqing Zeng, Hongkuan Zhou, Ajitesh Srivastava, Rajgopal Kannan, and Viktor
  Prasanna.
\newblock Graphsaint: Graph sampling based inductive learning method.
\newblock \emph{arXiv preprint arXiv:1907.04931}, 2019.

\bibitem[Zhang et~al.(2023)Zhang, Wang, Ioannidis, Adeshina, Zhang, Qin,
  Faloutsos, Zheng, Karypis, and Yu]{zhang2023orthoreg}
Hengrui Zhang, Shen Wang, Vassilis~N Ioannidis, Soji Adeshina, Jiani Zhang,
  Xiao Qin, Christos Faloutsos, Da~Zheng, George Karypis, and Philip~S Yu.
\newblock Orthoreg: Improving graph-regularized mlps via orthogonality
  regularization.
\newblock \emph{arXiv preprint arXiv:2302.00109}, 2023.

\bibitem[Zhang et~al.(2022)Zhang, Liu, Sun, and Shah]{zhang2021graph}
Shichang Zhang, Yozen Liu, Yizhou Sun, and Neil Shah.
\newblock Graph-less neural networks: Teaching old mlps new tricks via
  distillation.
\newblock In \emph{International Conference on Learning Representations}, 2022.

\bibitem[Zhu \& Koniusz(2021)Zhu and Koniusz]{zhu2020simple}
Hao Zhu and Piotr Koniusz.
\newblock Simple spectral graph convolution.
\newblock In \emph{International Conference on Learning Representations}, 2021.

\bibitem[Zhu et~al.(2020)Zhu, Yan, Zhao, Heimann, Akoglu, and
  Koutra]{zhu2020beyond}
Jiong Zhu, Yujun Yan, Lingxiao Zhao, Mark Heimann, Leman Akoglu, and Danai
  Koutra.
\newblock Beyond homophily in graph neural networks: Current limitations and
  effective designs.
\newblock \emph{Advances in Neural Information Processing Systems},
  33:\penalty0 7793--7804, 2020.

\bibitem[Zhu et~al.(2003)Zhu, Ghahramani, and Lafferty]{zhu2003semi}
Xiaojin Zhu, Zoubin Ghahramani, and John~D Lafferty.
\newblock Semi-supervised learning using gaussian fields and harmonic
  functions.
\newblock In \emph{Proceedings of the 20th International conference on Machine
  learning (ICML-03)}, pp.\  912--919, 2003.

\end{thebibliography}
\bibliographystyle{iclr2023_conference}

\clearpage
\tableofcontents

\clearpage
\appendix

\section{Proof for Proposition 1} \label{app_prop1}
To analyse the extrapolation behavior of PMLP and compare it with MLP, 
As mentioned in the main text, training an infinitely wide neural network using gradient descent with infinitesimal step size is equivalent to solving kernel regression with the so-called NTK by minimizing the following squared loss function:
\begin{equation} \label{eq_kernelreg}
    f(\bs x; \bs w) = \bs w^\top \phi_{ntk}(\bs x), \quad \mathcal{L}(\bs w)=\frac{1}{2} \sum_{i=1}^n\left(y_i-\bs w^\top \phi_{ntk}(\bs x_i )\right)^2.
\end{equation}
Let us now consider an arbitrary minimizer $\bs w^* \in \mathcal H$ in NTK's reproducing kernel Hilbert space. The minimizer could be further decomposed as $\bs w^* = \hat{\bs w}^* + \bs w_{\bot}$, where $\hat{\bs w}^*$ lies in the linear span of feature mappings for training data and $\bs w_{\bot}$ is orthogonal to $\hat{\bs w}^*$, i.e., 
\begin{equation}
    \hat{\bs w}^* = \sum_{i=1}^{n} \lambda_i \cdot \phi_{ntk}(\bs x_i), \quad \langle \hat{\bs w}^*, \bs w_{\bot} \rangle_{\mathcal H} = 0.
\end{equation}
One observation is that the loss function is unchanged after removing the orthogonal component $\bs w_{\bot}$:
\begin{equation}
\begin{split}
        \mathcal{L}(\bs w^*) &= \frac{1}{2} \sum_{i=1}^n\left(y_i-\langle\sum_{i=1}^{n} \lambda_i \cdot \phi_{ntk}(\bs x_i) + \bs w_{\bot}, \phi_{ntk}(\bs x_i )\rangle_{\mathcal H}\right)^2\\
        &= \frac{1}{2} \sum_{i=1}^n\left(y_i-\langle\sum_{i=1}^{n} \lambda_i \cdot \phi_{ntk}(\bs x_i), \phi_{ntk}(\bs x_i )\rangle_{\mathcal H}\right)^2 = \mathcal{L}(\hat{\bs w}^*).
\end{split}
\end{equation}
This indicates that $\hat{\bs w}^*$ is also a minimizer whose $\mathcal H$-norm is smaller than that of $\bs w^*$, i.e., $\|\hat{\bs w}^*\|_{\mathcal H} \leq \|{\bs w}^*\|_{\mathcal H}$. It follows that the minimum $\mathcal H$-norm solution for Eq.~\ref{eq_kernelreg} can be expressed as a linear combination of feature mappings for training data. Therefore, solving Eq.~\ref{eq_kernelreg} boils down to solving a linear system with coefficients $\bs{\lambda} = [\lambda_i]_{i=1}^n$. Resultingly, the minimum $\mathcal H$-norm solution is
\begin{equation}
    \bs w^*=\sum_{i=1}^n\left[y_i {\mathbf K}^{-1}\right]_i \phi_{ntk}(\bs x_i),
\end{equation}
where $\mathbf K \in \mathbb R^{n\times n}$ is the kernel matrix for training data. We see that the final solution is only dependent on training data $\{\bs x_i, y_i\}_{i=1}^n$ and the model architecture used in training (since it determines the form of $\phi_{ntk}(\bs x_i)$). It follows immediately that the min-norm NTK kernel regression solution is equivalent for MLP and PMLP (i.e., $\bs w^*_{mlp} = \bs w^*_{pmlp}$) given that they are the same model trained on the same set of data. In contrast, the architecture of GNN is different from that of MLP, implying different form of feature map, and hence they have different solutions in their respective NTK kernel regression problems (i.e., $\bs w^*_{mlp} \neq \bs w^*_{gnn}$).

\section{Proof for Lemma 2} \label{app_lemma3}
\subsection{GNTK for Graph-Level Tasks}
Graph Neural Tangent Kernel (GNTK)~\citep{du2019graph} is a natural extension of NTK to graph neural networks. Originally, the squared loss function is defined for graph-level regression and the kernel function is defined over a pair of graphs:
\begin{equation}
    \operatorname{GNTK}(\mathcal G_i, \mathcal G_j) = \phi_{gnn}(\mathcal G_i)^\top \phi_{gnn}(\mathcal G_j) = \left\langle \nabla_{\bs{\theta}} f(\mathcal G_i; \bs{\theta}), \nabla_{\bs{\theta}} f(\mathcal G_j; \bs{\theta}) \right\rangle,
\end{equation}
\begin{equation}
\mathcal{L}(\bs{\theta})=\frac{1}{2} \sum_{i=1}^n\left(y_i-f(\mathcal G_i; \bs{\theta})\right)^2,
\end{equation}
where $f(\mathcal G_i; \bs{\theta})$ yields prediction for a graph such as the property of a molecule. The formula for calculating GNTK is given by the following (where we modify the original notation for clarity and alignment with our definition of GNTK in node-level regression setting):
\begin{equation}
\begin{split}
        \left[\operatorname{GNTK}^{(0)}(\mathcal G_i, \mathcal G_j)\right]_{uu'}& =\left[\bs{\Sigma}^{(0)}\left(\mathcal G_i, \mathcal G_j\right)\right]_{uu'} = \bs x_u^\top \bs x_{u'}, \\
        \text{where}\quad&\bs x_u\in \mathcal V_i, \bs x_{u'}\in \mathcal V_j.
\end{split}
\end{equation}
The message passing operation in each layer corresponds to:
\begin{equation}
\begin{split}
    \left[\bs{\Sigma}_{mp}^{(\ell)}\left(\mathcal G_i, \mathcal G_j\right)\right]_{u u^{\prime}}&=c_u c_{u^{\prime}} \sum_{v \in \mathcal{N}_u \cup\{u\}} \sum_{v^{\prime} \in \mathcal{N}_{u^{\prime}} \cup\left\{u^{\prime}\right\}}\left[\bs{\Sigma}^{(\ell)}\left(\mathcal G_i, \mathcal G_j\right)\right]_{v v^{\prime}} \\
    \left[\operatorname{GNTK}_{mp}^{(\ell)}\left(\mathcal G_i, \mathcal G_j\right)\right]_{u u^{\prime}}&=c_u c_{u^{\prime}} \sum_{v \in \mathcal{N}_u \cup\{u\}} \sum_{v^{\prime} \in \mathcal{N}_{u^{\prime}} \cup\left\{u^{\prime}\right\}}\left[\operatorname{GNTK}^{(\ell)}\left(\mathcal G_i, \mathcal G_j\right)\right]_{v v^{\prime}},
\end{split}
\end{equation}
where $c_u$ denotes a scaling factor. The calculation formula of feed-forward operation (from $\operatorname{GNTK}_{mp}^{(\ell-1)}\left(\mathcal G_i, \mathcal G_j\right)$ to $\operatorname{GNTK}^{(\ell)}\left(\mathcal G_i, \mathcal G_j\right)$) is similar to that for NTKs of MLP~\citep{jacot2018neural}. The final output of GNTK (without jumping knowledge) is calculated by
\begin{equation}
    \operatorname{GNTK}\left(\mathcal G_i, \mathcal G_j\right)= \sum_{u \in \mathcal V_i, u^{\prime} \in \mathcal V_j}\left[\operatorname{GNTK}^{(L-1)}\left(\mathcal G_i, \mathcal G_j\right)\right]_{u u^{\prime}}.
\end{equation}

\subsection{GNTK for Node-Level Tasks}
We next extend the above definition of GNTK to the node-level regression setting, where the model $f(\bs x; \bs{\theta}, \mathcal G)$ outputs prediction of a node, and the kernel function is defined over a pair of nodes in a single graph:
\begin{equation}
    \operatorname{GNTK}(\bs x_i, \bs x_j) = \phi_{gnn}(\bs x_i)^\top \phi_{gnn}(\bs x_j) = \left\langle \nabla_{\bs{\theta}} f(\bs x_i; \bs{\theta}, \mathcal G), \nabla_{\bs{\theta}} f(\bs x_j; \bs{\theta}, \mathcal G) \right\rangle,
\end{equation}
\begin{equation}
\mathcal{L}(\bs{\theta})=\frac{1}{2} \sum_{i=1}^n\left(y_i-f(\bs x_i; \bs{\theta}, \mathcal G)\right)^2.
\end{equation}
Then, the explicit formula for GNTK in node-level regression is as follows. 
\begin{equation}
\begin{split}
\operatorname{GNTK}^{(0)}(\bs x_i, \bs x_j) = \bs{\Sigma}^{(0)}\left(\bs x_i, \bs x_j\right) = \bs x_i^\top \bs x_j, 
\end{split}
\end{equation}
Without loss of generality, we consider using random walk matrix as implementation of message passing. Then, the \textbf{message passing operation} in each layer corresponds to:
\begin{equation}
\begin{split}
    \bs{\Sigma}_{mp}^{(\ell)}\left(\bs x_i, \bs x_j\right)&=\frac{1}{(|\mathcal N_i|+1)(|\mathcal N_j|+1)} \sum_{i' \in \mathcal N_i \cup\{i\}} \sum_{j' \in \mathcal N_j \cup\left\{j\right\}} \bs{\Sigma}^{(\ell)}\left(\bs x_{i'}, \bs x_{j'}\right) \\
    \operatorname{GNTK}_{mp}^{(\ell)}\left(\bs x_i, \bs x_j\right)&=\frac{1}{(|\mathcal N_i|+1)(|\mathcal N_j|+1)} \sum_{i' \in \mathcal N_i \cup\{i\}} \sum_{j' \in \mathcal N_j \cup\left\{j\right\}} \operatorname{GNTK}^{(\ell)}\left(\bs x_{i'}, \bs x_{j'}\right),
\end{split}
\end{equation}
Moreover, the \textbf{feed-forward operation} in each layer corresponds to:
\begin{equation}
\begin{split}
    \bs{\Sigma}^{(\ell)}\left(\bs x_i, \bs x_j\right)&=c \cdot \underset{u, v \sim \mathcal{N}\left(\mathbf{0}, \bs{\Lambda}^{(\ell)}\right)}{\mathbb{E}}[\sigma(u) \sigma(v)], \\
    \dot{\bs{\Sigma}}^{(\ell)}\left(\bs x_i, \bs x_j\right)&=c \cdot \underset{u, v \sim \mathcal{N}\left(\mathbf{0}, \bs{\Lambda}^{(\ell)}\right)}{\mathbb{E}}[\dot{\sigma}(u) \dot{\sigma}(v)], \\
    \operatorname{GNTK}^{(\ell)}(\bs x_i, \bs x_j) &= \operatorname{GNTK}_{mp}^{(\ell-1)}(\bs x_i, \bs x_j) \cdot \dot{\bs{\Sigma}}^{(\ell)}\left(\bs x_i, \bs x_j\right) + {\bs{\Sigma}}^{(\ell)}\left(\bs x_i, \bs x_j\right),\\
    \text{where}\qquad\bs{\Lambda}^{(\ell)} &= \left[\begin{array}{cc}
     \bs{\Sigma}_{mp}^{(\ell-1)}(\bs x_i, \bs x_i) & \bs{\Sigma}_{mp}^{(\ell-1)}(\bs x_i, \bs x_j) \\
    \bs{\Sigma}_{mp}^{(\ell-1)}(\bs x_j, \bs x_i) & \bs{\Sigma}_{mp}^{(\ell-1)}(\bs x_j, \bs x_j)
    \end{array}\right]
\end{split}
\end{equation}
Suppose the GNN has $L$ layers and the last layer uses linear transformation that is akin to MLP, the final GNTK in node-level regression is defined as
\begin{equation}
    \operatorname{GNTK}(\bs x_i, \bs x_j) = \operatorname{GNTK}_{mp}^{(L-1)}(\bs x_i, \bs x_j)
\end{equation}

\subsection{GNTK and Feature Map for Two-Layer GNNs}
We next derive the explicit NTK formula for a two-layer graph neural network in node-level regression setting. For notational convenience, we use $\bs a_i \in \mathbb R^n$ to denote adjacency, i.e.,
\begin{equation}
    (\bs a_i)_j =  \begin{cases}1/(|\mathcal N_i|+1) & \text{ if }(i,j)\in \mathcal E  \\
    0 & \text{ if }(i,j)\notin \mathcal E\end{cases},
\end{equation}
and $\mathbf G = \mathbf X \mathbf X^\top \in \mathbb R^{n\times n}$ to denote the Gram matrix of all nodes. Then we have

\textit{(First message passing layer)}
\begin{equation}
\begin{split}
&\operatorname{GNTK}^{(0)}(\bs x_i, \bs x_j) = \bs{\Sigma}^{(0)}\left(\bs x_i, \bs x_j\right) = \bs x_i^\top \bs x_j, \\
&\operatorname{GNTK}_{mp}^{(0)}(\bs x_i, \bs x_j) = \bs{\Sigma}_{mp}^{(0)}\left(\bs x_i, \bs x_j\right) = \bs a_i^\top \mathbf G \bs a_j, \\
\end{split}
\end{equation}

\textit{(First feed-forward layer)}
\begin{equation}
\begin{split}
\bs{\Sigma}^{(1)}\left(\bs x_i, \bs x_j\right)&=c \cdot \underset{u, v \sim \mathcal{N}\left(\mathbf{0}, \bs{\Lambda}^{(1)}\right)}{\mathbb{E}}[\sigma(u) \sigma(v)], \\
\dot{\bs{\Sigma}}^{(1)}\left(\bs x_i, \bs x_j\right)&=c \cdot \underset{u, v \sim \mathcal{N}\left(\mathbf{0},\bs{\Lambda}^{(1)}\right)}{\mathbb{E}}[\dot{\sigma}(u) \dot{\sigma}(v)], \\
\bs{\Lambda}^{(1)} = \left[\begin{array}{cc}
     \bs a_i^\top \mathbf G \bs a_i & \bs a_i^\top \mathbf G \bs a_j \\
    \bs a_j^\top \mathbf G \bs a_i & \bs a_j^\top \mathbf G \bs a_j
    \end{array}\right] &=  \left[\begin{array}{c}
      \mathbf X^\top \bs a_i \\
      \mathbf X^\top \bs a_j 
    \end{array}\right]\cdot\left[\begin{array}{cc}
      \mathbf X^\top \bs a_i & \mathbf X^\top \bs a_j 
    \end{array}\right]\qquad\qquad\quad
\end{split}
\end{equation}
By noting that $\bs a_i^\top \mathbf G \bs a_j =  (\mathbf X^\top \bs a_i)^\top \mathbf X^\top \bs a_j$ and substituting $\sigma(k) = k\cdot \mathbb{I}_{+}(k)$, $\dot{\sigma}(k) = \mathbb{I}_{+}(k)$, where $\mathbb{I}_{+}(k)$ is an indicator function that outputs $1$ if $k$ is positive otherwise $0$, we have the following equivalent form for the covariance
\begin{equation}
\begin{split}
\bs{\Sigma}^{(1)}\left(\bs x_i, \bs x_j\right)&=c \cdot \underset{\bs w \sim \mathcal{N}\left(\mathbf{0}, I_d\right)}{\mathbb{E}}\left[\bs w^\top \mathbf X^\top \bs a_i \cdot \mathbb{I}_{+}(\bs w^\top \mathbf X^\top \bs a_i)\cdot \bs w^\top \mathbf X^\top \bs a_j \cdot \mathbb{I}_{+}(\bs w^\top \mathbf X^\top \bs a_j)\right], \\
\dot{\bs{\Sigma}}^{(1)}\left(\bs x_i, \bs x_j\right)&=c \cdot \underset{\bs w \sim \mathcal{N}\left(\mathbf{0}, I_d\right)}{\mathbb{E}}\left[\mathbb{I}_{+}(\bs w^\top \mathbf X^\top \bs a_i)\cdot \mathbb{I}_{+}(\bs w^\top \mathbf X^\top \bs a_j)\right].
\end{split}
\end{equation}
Hence, we have
\begin{equation}\label{eq_gntk1}
\begin{split}
\operatorname{GNTK}^{(1)}(\bs x_i, \bs x_j) &= \operatorname{GNTK}_{mp}^{(0)}(\bs x_i, \bs x_j) \cdot \dot{\bs{\Sigma}}^{(1)}\left(\bs x_i, \bs x_j\right) + {\bs{\Sigma}}^{(1)}\left(\bs x_i, \bs x_j\right)\\
&= c \cdot \underset{\bs w \sim \mathcal{N}\left(\mathbf{0}, I_d\right)}{\mathbb{E}}\left[\bs a_i^\top \mathbf G \bs a_j\cdot \mathbb{I}_{+}(\bs w^\top \mathbf X^\top \bs a_i)\cdot \mathbb{I}_{+}(\bs w^\top \mathbf X^\top \bs a_j)\right]\\
&+c \cdot \underset{\bs w \sim \mathcal{N}\left(\mathbf{0}, I_d\right)}{\mathbb{E}}\left[\bs w^\top \mathbf X^\top \bs a_i \cdot \mathbb{I}_{+}(\bs w^\top \mathbf X^\top \bs a_i)\cdot \bs w^\top \mathbf X^\top \bs a_j \cdot \mathbb{I}_{+}(\bs w^\top \mathbf X^\top \bs a_j)\right]
\end{split}
\end{equation}

\textit{(Second message passing layer / The last layer)}

Since the GNN uses a linear transformation on top of the (second) message passing layer for output, the neural tangent kernel for a two-layer GNN is given by
\begin{equation}\label{eq_finalgntk}
\begin{split}
        \operatorname{GNTK}(\bs x_i, \bs x_j) &=\operatorname{GNTK}_{mp}^{(1)}(\bs x_i, \bs x_j)\\
        &= \frac{1}{(|\mathcal N_i|+1)(|\mathcal N_j|+1)} \sum_{i' \in \mathcal N_i \cup\{i\}} \sum_{j' \in \mathcal N_j \cup\left\{j\right\}} \operatorname{GNTK}^{(1)}\left(\bs x_{i'}, \bs x_{j'}\right)\\
        &= \left\langle \left[\phi^{(1)}(\bs x_1), \cdots, \phi^{(1)}(\bs x_n)\right]^\top \bs a_i, \left[\phi^{(1)}(\bs x_1), \cdots, \phi^{(1)}(\bs x_n)\right]^\top \bs a_j\right\rangle_{\mathcal H}
\end{split}
\end{equation}
where $\mathbf K^{(1)} \in \mathbb R^{n\times n}$ and $\phi^{(1)}:\mathbb R^{d}\rightarrow \mathcal H$ is the kernel matrix and feature map induced by $\operatorname{GNTK}^{(1)}$. By Eq.~\ref{eq_finalgntk}, the final feature map $\phi_{gnn}(\bs x)$ is
\begin{equation} \label{eqn_phignn}
    \phi_{gnn}(\bs x_i) = \left[\phi^{(1)}(\bs x_1), \cdots, \phi^{(1)}(\bs x_n)\right]^\top \bs a_i.
\end{equation}
Also, notice that in Eq.~\ref{eq_gntk1},
\begin{equation}
\begin{split}
    \bs a_i^\top \mathbf G \bs a_j\cdot \mathbb{I}_{+}(\bs w^\top \mathbf X^\top \bs a_i)\cdot \mathbb{I}_{+}(\bs w^\top \mathbf X^\top \bs a_j) &= \phi_i^\top \phi_j, \\
    \bs w^\top \mathbf X^\top \bs a_i \cdot \mathbb{I}_{+}(\bs w^\top \mathbf X^\top \bs a_i)\cdot \bs w^\top \mathbf X^\top \bs a_j \cdot \mathbb{I}_{+}(\bs w^\top \mathbf X^\top \bs a_j) &= (\bs w^\top \phi_i)^\top \bs w^\top \phi_j,\\
    \mbox{where\;\;} \phi_i = \mathbf X^\top \bs a_i \cdot \mathbb{I}_{+}(\bs w^\top \mathbf X^\top \bs a_i)&.
\end{split}
\end{equation}
Then, the feature map $\phi^{(1)}$ can be written as
\begin{equation}\label{eqn_phi1}
\begin{split}
    \phi^{(1)}(\bs x_i) = c'\cdot\Big[\mathbf{X}^{\top} \bs{a}_i \cdot \mathbb{I}_{+}\left(\bs{w}^{(1)^{\top}} \mathbf{X}^{\top} \bs{a}_i \right)&, \bs{w}^{(1)^{\top}} \mathbf{X}^{\top} \bs{a}_i \cdot \mathbb{I}_{+}\left(\bs{w}^{(1)^{\top}} \mathbf{X}^{\top} \bs{a}_i \right), \\
    \mathbf{X}^{\top} \bs{a}_i \cdot \mathbb{I}_{+}\left(\bs{w}^{(2){\top}} \mathbf{X}^{\top} \bs{a}_i \right)&, \bs{w}^{(2)^{\top}} \mathbf{X}^{\top} \bs{a}_i \cdot \mathbb{I}_{+}\left(\bs{w}^{(2)^{\top}} \mathbf{X}^{\top} \bs{a}_i \right), \\
    &\cdots\\
    \mathbf{X}^{\top} \bs{a}_i \cdot \mathbb{I}_{+}\left(\bs{w}^{(\infty){\top}} \mathbf{X}^{\top} \bs{a}_i \right)&, \bs{w}^{(\infty)^{\top}} \mathbf{X}^{\top} \bs{a}_i \cdot \mathbb{I}_{+}\left(\bs{w}^{(\infty)^{\top}} \mathbf{X}^{\top} \bs{a}_i \right)\Big]\\
\end{split}
\end{equation}
where $\bs{w}^{(k)}\sim \mathcal{N}\left(\mathbf{0}, I_d\right)$ is random Gaussian vector in $\mathbb R^d$, with the superscript $^{(k)}$ denoting that it is the $k$-th sample among infinitely many i.i.d. sampled ones, $c'$ is a constant. We write Eq.~\ref{eqn_phi1} in short as
\begin{equation}\label{eqn_phi1short}
\begin{split}
    \phi^{(1)}(\bs x_i) = c'\cdot\Big[\mathbf{X}^{\top} \bs{a}_i \cdot \mathbb{I}_{+}\left(\bs{w}^{(k)^{\top}} \mathbf{X}^{\top} \bs{a}_i \right)&, \bs{w}^{(k)^{\top}} \mathbf{X}^{\top} \bs{a}_i \cdot \mathbb{I}_{+}\left(\bs{w}^{(k)^{\top}} \mathbf{X}^{\top} \bs{a}_i \right), \cdots\Big].\\
\end{split}
\end{equation}
Finally, substituting Eq.~\ref{eqn_phi1short} into Eq.~\ref{eqn_phignn} completes the proof.

\section{Proof for Theorem 4 and Theorem 5}
To analyse the extrapolation behavior of PMLP along a certain direction $\bs v$ in the testing phase and compare it to MLP, we consider a newly arrived testing node $\bs x_0 = t\bs{v}$, whose degree (with self-connection) is $\tilde{d}$ and its corresponding adjacency vector is $\bs a\in \mathbb R^{n+1}$, where $(\bs a)_i = 1/\tilde{d}$ if $(i,0)\in \mathcal E$ otherwise $0$. Following~\citep{xu2021neural,bietti2019inductive}, we consider a constant bias term and denote the data $\bs x$ plus this term as $\hat{\bs x} = [\bs x | 1]$. Then, the asymptotic behavior of $f(\cdot)$ at large distances from the training data range can be characterized by the change of network output with a fixed-length step $\Delta t\cdot \bs v$ along the direction $\bs v$, which is given by the following in the NTK regime
\begin{align}
    \frac{1}{{\Delta t}} \left( f_{mlp}\left(\hat{\bs{x}}\right) - f_{mlp}\left(\hat{\bs{x}}_0\right) \right) &= \frac{1}{{\Delta t}} \bs w_{mlp}^{*^\top} (\phi_{mlp}(\hat{\bs{x}}) - \phi_{mlp}(\hat{\bs{x}}_0))\label{eq_mlp}\\
    \frac{1}{{\Delta t}} \left( f_{pmlp}\left(\hat{\bs{x}}\right) - f_{pmlp}\left(\hat{\bs{x}}_0\right) \right) &= \frac{1}{{\Delta t}} \bs w_{mlp}^{*^\top} (\phi_{gnn}(\hat{\bs{x}}) - \phi_{gnn}(\hat{\bs{x}}_0)) \label{eq_target}
\end{align}
where $\bs x = \bs x_0 + \Delta t\cdot \bs v= (t+\Delta t)\bs v$, $\hat{\bs x} = [\hat{\bs x} | 1]$ and $\hat{\bs x}_0 = [\hat{\bs x}_0 | 1]$. As our interest is in how PMLP extrapolation, we use $f(\cdot)$ to refer to $f_{pmlp}(\cdot)$ in the rest of the proof. By Lemma.~\ref{lemma3}, the explicit formula for computing this node's feature map is given by
\begin{equation}
    \phi_{gnn}(\hat{\bs x}_0) = \left[\phi^{(1)}(\hat{\bs x}_0), \phi^{(1)}(\bs x_1; \hat{\bs x}_0), \cdots, \phi^{(1)}(\bs x_n; \hat{\bs x}_0) \right]^\top \bs a,
\end{equation}
where
\begin{equation} \label{eqn_phi0}
\begin{split}
\phi^{(1)}(\hat{\bs x}_0) = c'\cdot\Big[[\hat{\bs x}_0,\mathbf{X}]^{\top} \bs{a} \cdot \mathbb{I}_{+}&\left(\bs{w}^{(k)^{\top}} [\hat{\bs x}_0,\mathbf{X}]^{\top} \bs{a} \right), \\
&\bs{w}^{(k)^{\top}} [\hat{\bs x}_0,\mathbf{X}]^{\top} \bs{a} \cdot \mathbb{I}_{+}\left(\bs{w}^{(k)^{\top}} [\hat{\bs x}_0,\mathbf{X}]^{\top} \bs{a} \right), \ldots\Big],
\end{split}
\end{equation}
and similarly
\begin{equation} \label{eqn_phii}
\begin{split}
\phi^{(1)}(\bs x_i; \hat{\bs x}_0) = c'\cdot\Big[[\hat{\bs x}_0,\mathbf{X}]^{\top} \bs{a}_i \cdot \mathbb{I}_{+}&\left(\bs{w}^{(k)^{\top}} [\hat{\bs x}_0,\mathbf{X}]^{\top} \bs{a}_i \right), \\
&\bs{w}^{(k)^{\top}} [\hat{\bs x}_0,\mathbf{X}]^{\top} \bs{a}_i \cdot \mathbb{I}_{+}\left(\bs{w}^{(k)^{\top}} [\hat{\bs x}_0,\mathbf{X}]^{\top} \bs{a}_i \right), \ldots\Big],
\end{split}
\end{equation}
$\bs{w}^{(k)} \sim \mathcal{N}(\mathbf{0}, \bs{I}_d)$, with $k$ going to infinity, $c^{\prime}$ is a constant, $\mathbb{I}_{+}(k)$ is an indicator function that outputs $1$ if $k$ is positive otherwise $0$, and $(\bs a_i)_1 = 1/(|\mathcal N_i| + 1)$ if $i$ is connected to the new testing node. It follows from Eq.~\ref{eq_target} that
\begin{equation} \label{eq_target_new}
\begin{split}
        &\frac{1}{\Delta t}\left(f\left(\hat{\bs{x}}\right) - f\left(\hat{\bs{x}}_0\right)\right)\\
        = &\frac{1}{\Delta t} \bs{w}^{*^\top}_{mlp} \left[\phi^{(1)}(\hat{\bs x})-\phi^{(1)}(\hat{\bs x}_0), \cdots, \phi^{(1)}(\bs x_n; \hat{\bs x}) - \phi^{(1)}(\bs x_n; \hat{\bs x}_0) \right]^\top \bs a\\
        = &\frac{1}{\tilde d \Delta t} \bs{w}^{*^\top}_{mlp} \left( \phi^{(1)}(\hat{\bs x})-\phi^{(1)}(\hat{\bs x}_0)\right) + \frac{1}{\tilde d \Delta t} \sum_{i\in \mathcal N_0} \bs{w}^{*^\top}_{mlp} \left(\phi^{(1)}(\bs x_i; \hat{\bs x}) - \phi^{(1)}(\bs x_i; \hat{\bs x}_0)\right)
\end{split}
\end{equation}
Now, let us consider $\bs{w}^{*^\top}_{mlp} \left(\phi^{(1)}(\hat{\bs x})-\phi^{(1)}(\hat{\bs x}_0)\right)$. Recall that $\bs{w}^{*^\top}_{mlp}$ is from infinite dimensional Hilbert space, and $\bs w^{(k)}$ is drawn from Gaussian with $k$ going to infinity in Eq.~\ref{eqn_phi0} and Eq.~\ref{eqn_phii}, where each $\bs w^{(k)}$ corresponds to some certain dimensions of $\bs{w}^{*}_{mlp}$. Let us denote the part that corresponds to the first line in Eq.~\ref{eqn_phi0} as $\bs \beta_{\bs w^{(k)}}$ and the second line in Eq.~\ref{eqn_phi0} as $\bs \gamma_{\bs w^{(k)}}$. Consider the following way of rearrangement for (the first element in) $\phi^{(1)}(\hat{\bs x})-\phi^{(1)}(\hat{\bs x}_0)$
\begin{equation}
\begin{split}
    &[\hat{\bs x},\mathbf{X}]^{\top} \bs{a} \cdot \mathbb{I}_{+}\left(\bs{w}^{(k)^{\top}} [\hat{\bs x},\mathbf{X}]^{\top} \bs{a} \right)- [\hat{\bs x}_0,\mathbf{X}]^{\top} \bs{a} \cdot \mathbb{I}_{+}\left(\bs{w}^{(k)^{\top}} [\hat{\bs x}_0,\mathbf{X}]^{\top} \bs{a} \right)\\
    = &[\hat{\bs x},\mathbf{X}]^{\top} \bs{a} \left(\mathbb{I}_{+}\left(\bs{w}^{(k)^{\top}} [\hat{\bs x},\mathbf{X}]^{\top} \bs{a} \right)-\mathbb{I}_{+}\left(\bs{w}^{(k)^{\top}} [\hat{\bs x}_0,\mathbf{X}]^{\top} \bs{a} \right)\right)\\
    &\qquad\qquad\qquad\qquad\qquad\qquad\qquad+ \frac{1}{\tilde d}\;[\Delta t \bs{v} \;|\; 0] \cdot \mathbb{I}_{+}\left(\bs{w}^{(k)^{\top}} [\hat{\bs x}_0,\mathbf{X}]^{\top} \bs{a} \right),
\end{split}
\end{equation}
where $\frac{1}{\tilde d}\;[\Delta t \bs{v} \;|\; 0]$ is obtained by subtracting $[\hat{\bs x}_0,\mathbf{X}]^{\top} \bs{a}$ from $[\hat{\bs x},\mathbf{X}]^{\top} \bs{a}$. Then, we can re-write $\bs{w}^{*^\top}_{mlp} \left(\phi^{(1)}(\hat{\bs x})-\phi^{(1)}(\hat{\bs x}_0)\right)$ into a more convenient form:
\begin{align}
&\frac{1}{\Delta t}\bs{w}^{*^\top}_{mlp}\left( \phi^{(1)}(\hat{\bs x})-\phi^{(1)}(\hat{\bs x}_0)\right) \label{eq_phiall} \\ 
=& \int \frac{1}{\tilde d}\; \bs{\beta}_{\bs{w}}^{{\top}} [\bs{v} \;|\; 0] \cdot \mathbb{I}_{+}\left(\bs{w}^{\top} [\hat{\bs x}_0,\mathbf{X}]^{\top} \bs{a} \right) \mathrm{d} \mathbb{P}(\bs{w}) \label{eq1term} \\
+& \int \bs{\beta}_{\bs{w}}^{{\top}} [\hat{\bs x}/\Delta t,\mathbf{X}/\Delta t]^{\top} \bs{a} \left(\mathbb{I}_{+}\left(\bs{w}^{{\top}} [\hat{\bs x},\mathbf{X}]^{\top} \bs{a} \right)-\mathbb{I}_{+}\left(\bs{w}^{{\top}} [\hat{\bs x}_0,\mathbf{X}]^{\top} \bs{a} \right)\right) \mathrm{d} \mathbb{P}(\bs{w}) \label{eq2term} \\ 
+& \int \frac{1}{\tilde d}\; \bs{\gamma}_{\bs{w}} \bs{w}^\top [\bs{v} \;|\; 0] \cdot \mathbb{I}_{+}\left(\bs{w}^{\top} [\hat{\bs x}_0,\mathbf{X}]^{\top} \bs{a} \right) \mathrm{d} \mathbb{P}(\bs{w}) \label{eq3term}\\
+& \int \bs{\gamma}_{\bs{w}} \bs{w}^\top [\hat{\bs x}/\Delta t,\mathbf{X}/\Delta t]^{\top} \bs{a} \left(\mathbb{I}_{+}\left(\bs{w}^{{\top}} [\hat{\bs x},\mathbf{X}]^{\top} \bs{a} \right)-\mathbb{I}_{+}\left(\bs{w}^{{\top}} [\hat{\bs x}_0,\mathbf{X}]^{\top} \bs{a} \right)\right) \mathrm{d} \mathbb{P}(\bs{w}) \label{eq4term}
\end{align}
\textbf{Remark.} For other components in Eq.~\ref{eq_target_new}, i.e., $\frac{1}{\Delta t}\bs{w}^{*^\top}_{mlp}\left(\phi^{(1)}(\bs x_i; \hat{\bs x}) - \phi^{(1)}(\bs x_i; \hat{\bs x}_0)\right)$ where $i \in \mathcal N_0$, the corresponding result of the expansion in Eq.~\ref{eq_phiall} is similar, which only differs by a scaling factor (since the first element in both $\bs a$ and $\bs a_i$ indicating whether the current node is connected to the new testing node is non-zero) and replacing $\bs a$ with $\bs a_i$. Therefore, in the following proof we focus on Eq.~\ref{eq_phiall} and then generalize the result to other components in Eq.~\ref{eq_target_new}.

\subsection{Proof for Theorem 4}
We first analyse the convergence of Eq.~\ref{eq_phiall}. Specifically, for Eq.~\ref{eq1term}:
\begin{equation} \label{eq1cvg}
\begin{split}
    &\int \frac{1}{\tilde d}\; \bs{\beta}_{\bs{w}}^{{\top}} [\bs{v} \;|\; 0] \cdot \mathbb{I}_{+}\left(\bs{w}^{\top} [\hat{\bs x}_0,\mathbf{X}]^{\top} \bs{a} \right) \mathrm{d} \mathbb{P}(\bs{w})\\
    = &\int \frac{1}{\tilde d}\; \bs{\beta}_{\bs{w}}^{{\top}} [\bs{v} \;|\; 0] \cdot \mathbb{I}_{+}\left(\bs{w}^{\top} [\hat{\bs x}_0/t,\mathbf{X}/t]^{\top} \bs{a} \right) \mathrm{d} \mathbb{P}(\bs{w})\\
     \rightarrow &\int \frac{1}{\tilde d}\; \bs{\beta}_{\bs{w}}^{{\top}} [\bs{v} \;|\; 0] \cdot \mathbb{I}_{+}\left(\bs{w}^{\top} [\bs v \;|\; 0] \right) \mathrm{d} \mathbb{P}(\bs{w}) = \frac{c'_{\bs v}}{\tilde d},  \quad \text { as } t \rightarrow \infty
\end{split}
\end{equation}
where the final result is a constant that depends on training data, direction $\bs{v}$ and node degree $\tilde d$. Moreover, the convergence of Eq.~\ref{eq2term} is given by 
\begin{equation} \label{eq2cvg}
\begin{split}
    &\int \bs{\beta}_{\bs{w}}^{{\top}} [\hat{\bs x}/\Delta t,\mathbf{X}/\Delta t]^{\top} \bs{a} \left(\mathbb{I}_{+}\left(\bs{w}^{{\top}} [\hat{\bs x},\mathbf{X}]^{\top} \bs{a} \right)-\mathbb{I}_{+}\left(\bs{w}^{{\top}} [\hat{\bs x}_0,\mathbf{X}]^{\top} \bs{a} \right)\right) \mathrm{d} \mathbb{P}(\bs{w})\\
    = &\int \bs{\beta}_{\bs{w}}^{{\top}} [\hat{\bs x}/\Delta t,\mathbf{X}/\Delta t]^{\top} \bs{a} \left(\mathbb{I}_{+}\left(\bs{w}^{{\top}} [[\bs v \;|\; \frac{1}{t+\Delta t}],\frac{\mathbf{X}}{t+\Delta t}]^{\top} \bs{a} \right)-\mathbb{I}_{+}\left(\bs{w}^{{\top}} [[\bs v \;|\; \frac{1}{t}],\frac{\mathbf{X}}{t}]^{\top} \bs{a} \right)\right) \mathrm{d} \mathbb{P}(\bs{w})\\
     \rightarrow &\int \bs{\beta}_{\bs{w}}^{{\top}} [\hat{\bs x}/\Delta t,\mathbf{X}/\Delta t]^{\top} \bs{a} \left(\mathbb{I}_{+}\left(\bs{w}^{{\top}} [\bs v \;|\; 0] \right)-\mathbb{I}_{+}\left(\bs{w}^{{\top}} [\bs v \;|\; 0] \right)\right) \mathrm{d} \mathbb{P}(\bs{w}) = 0,  \quad \text { as } t \rightarrow \infty
\end{split}
\end{equation}
The similar results in Eq.~\ref{eq1cvg} and Eq.~\ref{eq2cvg} also apply to analysis of Eq.~\ref{eq3term} and Eq.~\ref{eq4term}, respectively. By combining these results, we conclude that
\begin{equation} \label{eqn_coefficient}
    \frac{1}{\Delta t}\bs{w}^{*^\top}_{mlp}\left( \phi^{(1)}(\hat{\bs x})-\phi^{(1)}(\hat{\bs x}_0)\right) \rightarrow  \frac{c_{\bs v}}{\tilde d},  \quad \text { as } t \rightarrow \infty
\end{equation}
It follows that
\begin{equation} \label{eq_convergefinal}
    \frac{1}{\Delta t}\left(f\left(\hat{\bs{x}}\right) - f\left(\hat{\bs{x}}_0\right)\right) \rightarrow  c_{\bs v}{\tilde d}^{-1}\sum_{i\in \mathcal N_0 \cup \{0\}} \tilde{d}_i^{-1},  \quad \text { as } t \rightarrow \infty.
\end{equation}

In conclusion, both MLP and PMLP with ReLU activation will eventually converge to a linear function along directions away from the training data. In fact, this result also holds true for two-layer GNNs with weighted-sum style message passing layers by simply replacing $\bs{w}^{*}_{mlp}$ by $\bs{w}^{*}_{gnn}$ in the proof.

However, a remarkable difference between MLP and PMLP is that the linear coefficient for MLP is a constant $c_{\bs v}$ that is fixed upon a specific direction $\bs v$ and not affected by the inter-connection between testing node $\bs x$ and training data $\{(\bs x_i, y_i)\}_{i=1}^n$. In contrast, the linear coefficient for PMLP (and GNN) is also dependent on testing node's degree and the degrees of adjacent nodes. 

Moreover, by Proposition.~\ref{prop1}, MLP and PMLP share the same $\bs w_{mlp}^*$ (including $\bs{\beta}_{\bs w}$ and $\bs{\gamma}_{\bs w}$), and thus the constant $c_{\bs v}$ in Eq.~\ref{eq_convergefinal} is exactly the linear coefficient of MLP. This can also be verified by setting $\bs x$ to be an isolated node, in which case ${\tilde d}^{-1} \sum_{i\in \mathcal N_0 \cup \{0\}} \tilde{d}_i^{-1} = 1$ and PMLP is equivalent to MLP. Therefore, we can directly compare the linear coefficients for MLP and PMLP. As an immediate consequence, if all node degrees of adjacent nodes are larger than the node degree of the testing node, the linear coefficient will become smaller, vice versa.

\subsection{Proof for Theorem 5}
We next analyse the convergence rate for Eq.~\ref{eq1cvg} and Eq.~\ref{eq2cvg} to see to what extent can PMLP deviate from the converged linear coefficient as an indication of its tolerance to out-of-distribution sample. For Eq.~\ref{eq1cvg}, we have
\begin{equation}
\begin{split}
    &\left|\int \frac{1}{\tilde d}\; \bs{\beta}_{\bs{w}}^{{\top}} [\bs{v} \;|\; 0] \cdot \left(\mathbb{I}_{+}\left(\bs{w}^{\top} [\hat{\bs x}_0,\mathbf{X}]^{\top} \bs{a} \right) - \mathbb{I}_{+}\left(\bs{w}^{\top} [\bs v \;|\; 0] \right)\right)   \mathrm{d} \mathbb{P}(\bs{w})\right|\\
    \leq \quad &\frac{c'_1}{\tilde d}\cdot\int \left| \mathbb{I}_{+}\left(\bs{w}^{\top} [\hat{\bs x}_0,\mathbf{X}]^{\top} \bs{a} \right) - \mathbb{I}_{+}\left(\bs{w}^{\top} [\bs v\;|\; 0] \right) \right| \mathrm{d} \mathbb{P}(\bs{w})\\
    =\quad &\frac{c'_1}{\tilde d} \cdot\int \left| \mathbb{I}_{+}\left(\bs{w}^{\top} [[\bs x_0 \;|\; 1],\mathbf{X}]^{\top} \bs{a} \right) - \mathbb{I}_{+}\left(\bs{w}^{\top} [\bs x_0\;|\; 0] \right) \right| \mathrm{d} \mathbb{P}(\bs{w})\\
\end{split}
\end{equation}
Based on the observation that the integral of $| \mathbb{I}_{+}(\bs{w}^{\top} \bs v_1 ) - \mathbb{I}_{+}(\bs{w}^{\top} \bs v_2 ) |$ represents the volume of non-overlapping part of two half-balls that are orthogonal to $\bs v_1$ and $\bs v_2$, which grows linearly with the angle between $\bs v_1$ and $\bs v_2$, denoted by $\measuredangle\left(\bs v_1, \bs v_2\right)$. Therefore, we have 
\begin{equation} \label{eq_theta1}
\begin{split}
&\frac{c'_1}{\tilde d} \cdot\int \left| \mathbb{I}_{+}\left(\bs{w}^{\top} [[\bs x_0 \;|\; 1],\mathbf{X}]^{\top} \bs{a} \right) - \mathbb{I}_{+}\left(\bs{w}^{\top} [\bs x_0\;|\; 0] \right) \right| \mathrm{d} \mathbb{P}(\bs{w})\\
    =\quad &\frac{c_1}{\tilde d}\cdot \measuredangle\left([[\bs x_0\;|\;1],\mathbf{X}]^{\top} \bs{a} \;,\; [\bs x_0\;|\; 0]\right),
\end{split}
\end{equation}
Note that the first term in the angle can be decomposed as 
\begin{equation}
    \tilde{d}\cdot[[\bs x_0 \;|\; 1],\mathbf{X}]^{\top} \bs{a} = [\bs x_0 \;|\; 0] + [\bs 0 \;|\; 1] + \sum_{i \in \mathcal{N}(0)} \bs x_i.
\end{equation}
Suppose all node features are normalized, then we have
\begin{equation}\label{eq_rate1}
\begin{split}
    \measuredangle\left([[\bs x_0, 1],\mathbf{X}]^{\top} \bs{a} \;,\; [\bs x_0, 0]\right) & \leq \measuredangle\left([\bs x_0, 0] \;,\; [\bs x_0, 1]\right) + \measuredangle(\hat{\bs x}_0 \;,\; \hat{\bs x}_0 + \sum_{i \in \mathcal{N}(0)} \bs x_i)\\
    & = \operatorname{arctan}(\frac{1}{t}) + \operatorname{arctan}(\frac{(\tilde d-1)\sqrt{1-\alpha^2}}{(\tilde d-1) \alpha + \sqrt{t^2+1}})\\
    & = O(\frac{1+(\tilde{d}-1)\sqrt{1-\alpha^2}}{t})
\end{split}
\end{equation}

where $\alpha$ denotes the cosine similarity of the testing node and the sum of its neighbors. The last step is obtained by noting $\operatorname{arctan}(x) < x$. Using the same reasoning, for Eq.~\ref{eq2cvg}, we have
\begin{equation}
\begin{split}
    &\left| \int \bs{\beta}_{\bs{w}}^{{\top}} [\hat{\bs x}/\Delta t,\mathbf{X}/\Delta t]^{\top} \bs{a} \left(\mathbb{I}_{+}\left(\bs{w}^{{\top}} [\hat{\bs x}_0,\mathbf{X}]^{\top} \bs{a} \right)-\mathbb{I}_{+}\left(\bs{w}^{{\top}} [\hat{\bs x},\mathbf{X}]^{\top} \bs{a} \right)\right) \mathrm{d} \mathbb{P}(\bs{w}) \right|\\
    \leq \quad & \left|\bs{\beta}_{\bs{w}}^{{\top}} [\hat{\bs x}/\Delta t,\mathbf{X}/\Delta t]^{\top} \bs{a}\right|\cdot \int \left| \mathbb{I}_{+}\left(\bs{w}^{{\top}} [\hat{\bs x}_0,\mathbf{X}]^{\top} \bs{a} \right)-\mathbb{I}_{+}\left(\bs{w}^{{\top}} [\hat{\bs x},\mathbf{X}]^{\top} \bs{a} \right) \right| \mathrm{d} \mathbb{P}(\bs{w})\\ 
    = \quad &\left|\bs{\beta}_{\bs{w}}^{{\top}} [\hat{\bs x}/\Delta t,\mathbf{X}/\Delta t]^{\top} \bs{a}\right| \cdot \int \left| \mathbb{I}_{+}\left(\bs{w}^{{\top}} \Big[[\bs x_0\;|\;1],\mathbf{X}\Big]^{\top} \bs{a} \right)-\mathbb{I}_{+}\left(\bs{w}^{{\top}} \Big[[\bs x_0\;|\;\frac{t}{t+\Delta t}],\frac{t}{t+\Delta t}\mathbf{X}\Big]^{\top} \bs{a} \right) \right| \mathrm{d} \mathbb{P}(\bs{w})\\
    = \quad &\left|\bs{\beta}_{\bs{w}}^{{\top}} [\hat{\bs x}/\Delta t,\mathbf{X}/\Delta t]^{\top} \bs{a}\right|\cdot \measuredangle\left( \Big[[\bs x_0\;|\;1],\mathbf{X}\Big]^{\top} \bs{a} \;,\; \Big[[\bs x_0\;|\;\frac{t}{t+\Delta t}],\frac{t}{t+\Delta t}\mathbf{X}\Big]^{\top} \bs{a}\right),
 \end{split}
\end{equation}
Note that the second term in the angle can be re-written as
\begin{equation}
    \Big[[\bs x_0\;|\;\frac{t}{t+\Delta t}],\frac{t}{t+\Delta t}\mathbf{X}\Big]^{\top} \bs{a} =\frac{t}{t+\Delta t}  \cdot\Big[[\bs x_0\;|\;1],\mathbf{X}\Big]^{\top} \bs{a} + \frac{\Delta t}{t+\Delta t} \cdot \Big[[\bs x_0\;|\;0],\mathbf{0}\Big]^{\top} \bs{a},
\end{equation}
and hence the angle is at most $\Delta t/(t+\Delta t)$ times of that in Eq.~\ref{eq_theta1}. It follows that
\begin{equation} \label{eq_rate2}
\begin{split}
    &\left| \int \bs{\beta}_{\bs{w}}^{{\top}} [\hat{\bs x}/\Delta t,\mathbf{X}/\Delta t]^{\top} \bs{a} \left(\mathbb{I}_{+}\left(\bs{w}^{{\top}} [\hat{\bs x}_0,\mathbf{X}]^{\top} \bs{a} \right)-\mathbb{I}_{+}\left(\bs{w}^{{\top}} [\hat{\bs x},\mathbf{X}]^{\top} \bs{a} \right)\right) \mathrm{d} \mathbb{P}(\bs{w}) \right|\\
    = \quad &O(\frac{t+\Delta t}{\tilde d}) \cdot  O(\frac{1+(\tilde{d}-1)\sqrt{1-\alpha^2}}{t}) \cdot O(\frac{\Delta t}{t+\Delta t})\\
    = \quad &O(\frac{1+(\tilde{d}-1)\sqrt{1-\alpha^2}}{\tilde d t})
 \end{split}
\end{equation}
For Eq.~\ref{eq3term} and Eq.~\ref{eq4term}, similar results can be derived by bounding $\bs w$ with standard concentration techniques. By substituting the above convergence rates to Eq.~\ref{eq_phiall}, and dividing it by the linear coefficient in Eq.~\ref{eqn_coefficient}, the convergence rate for Eq.~\ref{eq_phiall} is 
\begin{equation}
\begin{split}
    & \left|\frac{\frac{1}{\Delta t}\bs{w}^{*^\top}_{mlp}\left( \phi^{(1)}(\hat{\bs x})-\phi^{(1)}(\hat{\bs x}_0)\right) -  c_{\bs v}/\tilde d}{c_{\bs v}/\tilde d}\right| = O(\frac{1+(\tilde{d} -1) \sqrt{1-\alpha^2}}{ t})\\
\end{split}
\end{equation}
It follows that the convergence rate for Eq.~\ref{eq_target} is $O(\frac{1+(\tilde{d}_{max} -1) \sqrt{1-\alpha_{min}^2}}{ t})$, where $\tilde{d}_{max}$ denotes the maximum node degree in the testing node' neighbors (including itself) and 
\begin{equation}
    \alpha_{min} = \operatorname{min}\{\alpha_i\}_{i\in \mathcal N_0 \cup \{0\}}
\end{equation}
denotes the minimum cosine similarity for the testing node' neighbors (including itself). This completes the proof.

\clearpage
\section{Over-Smoothing, Deeper GNNs, and Heterophily} \label{app_oversmooth}

\begin{figure}[t]
	\centering
 	\subfigure{
    	\includegraphics[width=0.99\textwidth]{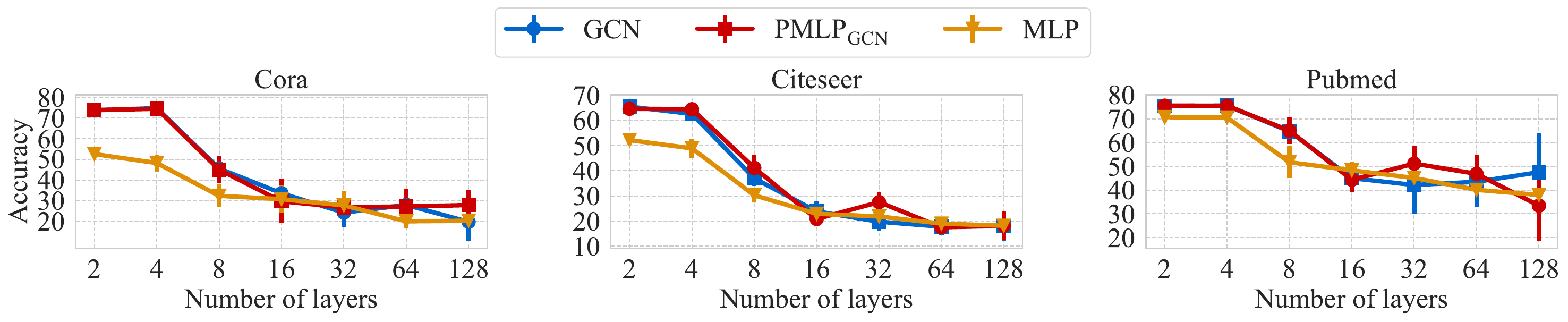}
	}
	\subfigure{
    	\includegraphics[width=0.99\textwidth]{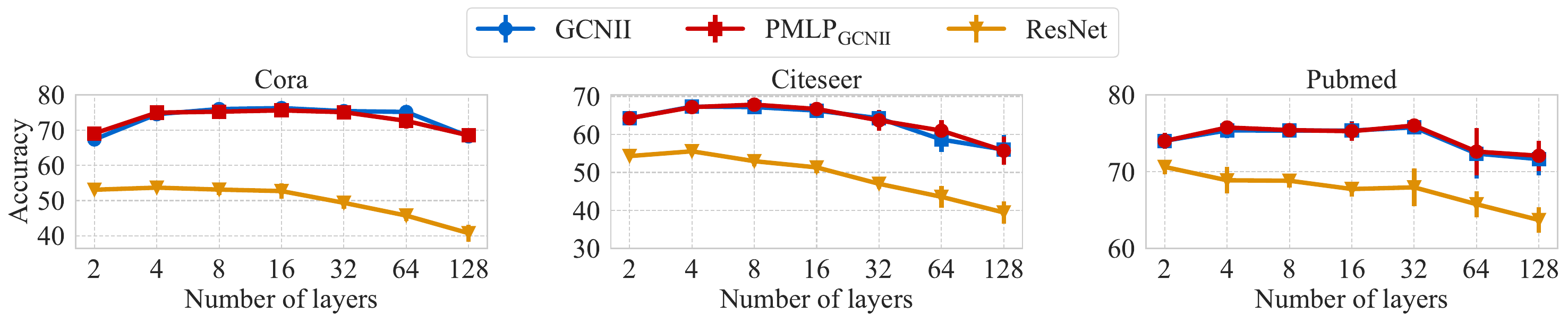}
	}
 	\subfigure{
    	\includegraphics[width=0.99\textwidth]{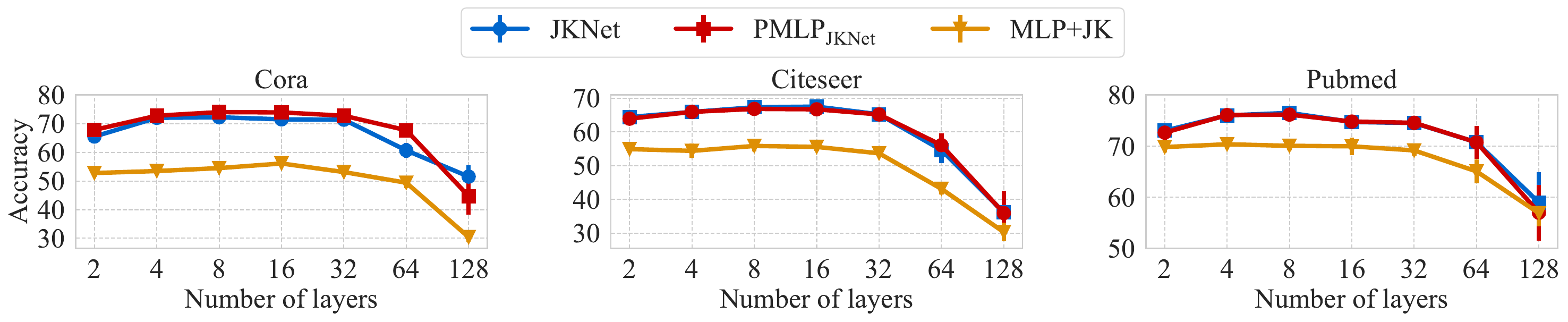}
	}
	\subfigure{
    	\includegraphics[width=0.99\textwidth]{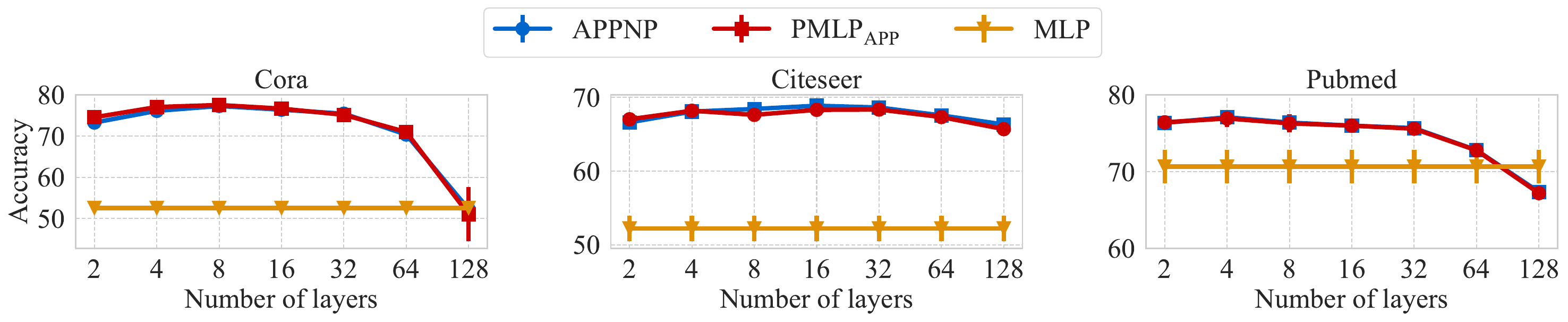}
	}
 	\subfigure{
    	\includegraphics[width=0.99\textwidth]{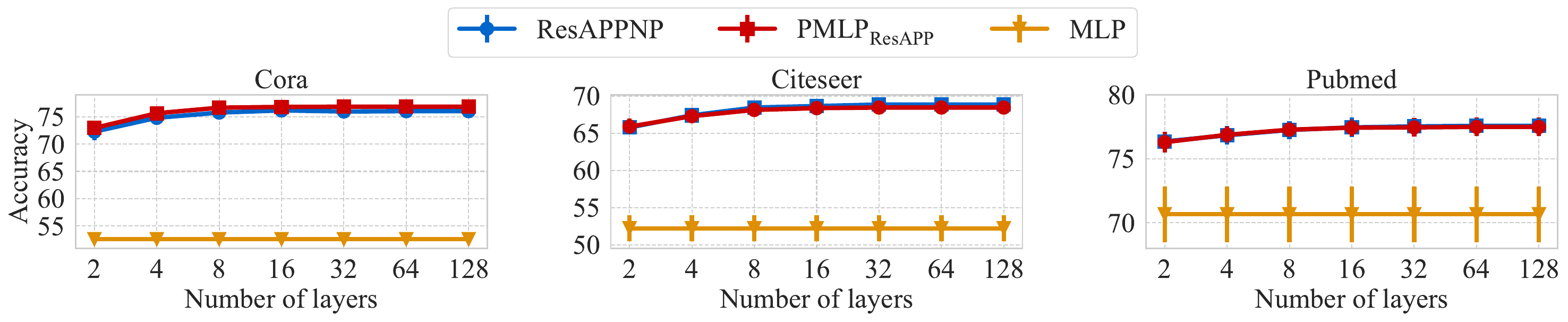}
	}
	\subfigure{
    	\includegraphics[width=0.99\textwidth]{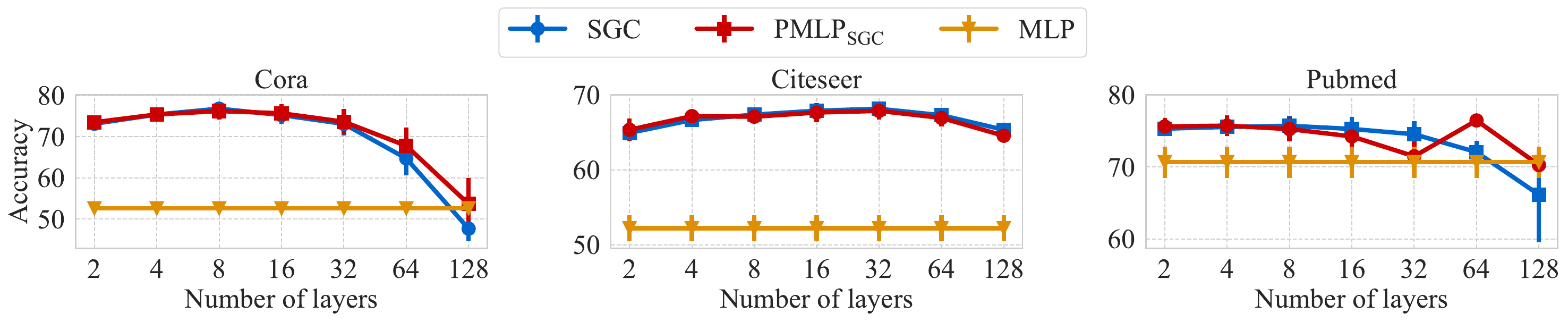}
	}
	\subfigure{
    	\includegraphics[width=0.99\textwidth]{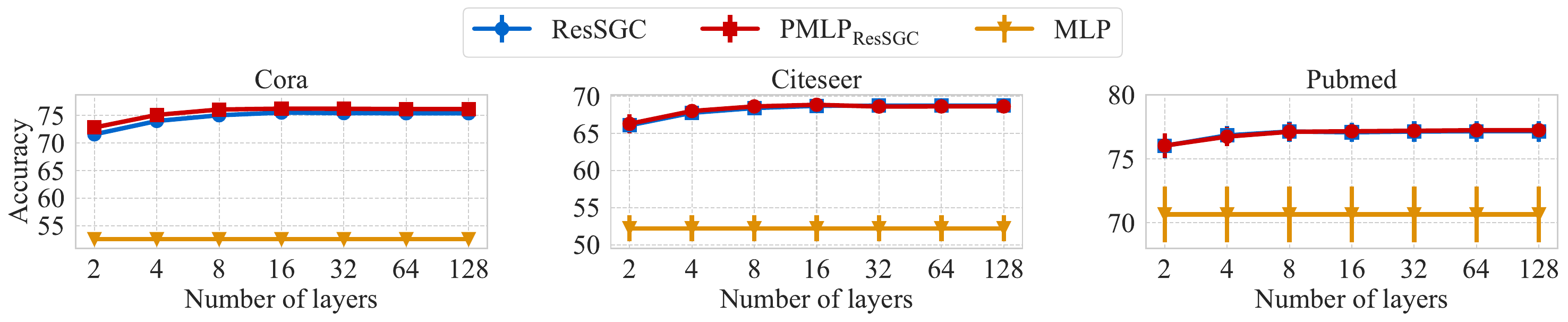}
	}
	\caption{Performance variation of different GNN architectures with (GCNII, JKNet, ResAPPNP/SGC) or without (GCN, APPNP/SGC) residual links w.r.t. increasing layer number (from $2$ to $128$).} \label{fig_residual}
\end{figure}

\paragraph{Over-Smoothing and Deeper GNNs.} To gain more insights into the over-smoothing problem and the impact of model depth, we further investigate GNN architectures with residual connections (including GCN-style ones where residual connections are employed across FF layers, e.g., JKNet~\citep{xu2018representation} and GCNII~\citep{chen2020simple}, and SGC/APPNP-style ones where the implementation of MP layers involve residual connections, e.g., APPNP with non-zero $\alpha$). Table.~\ref{tab_over1} and Table.~\ref{tab_over2} respectively report the results for SGC/APPNP-style and GCN-style GNNs on Cora dataset. Similar trends are also observed on other datasets, some of which are plotted in Fig.~\ref{fig_residual}. 

As we can see, the performances of PMLPs without residual connections (i.e., PMLP$_{GCN}$, PMLP$_{SGC}$ and PMLP$_{APP}$) exhibit very similar downward trends with their GNN counterparts w.r.t. increasing layer number (from $2$ to $128$). Such phenomenon in GNNs is commonly thought to be caused by the oversmoothing issue wherein node features become hard to distinguish after multiple steps of message passing, and since PMLP is immune from such problem in the training stage but still perform poorly in testing, we may conclude that oversmoothing is more of a problem related to failure modes of GNNs' generalization ability, rather than impairing their representational power, which is somewhat in alignment with \citep{cong2021provable} where the authors theoretically show very deep GNNs that are vulnerable to oversmoothing can still achieve high training accuracy but will perform poorly in terms of generalization. We experimental results further suggest the reason why additional residual connections (either in MP layers, e.g., ResAPPNP/SGC, or across FF layers, e.g., GCNII and JKNet) are empirical effective for solving oversmoothing is that they can improve GNN's generalization ability according to results of PMLP$_{GCNII}$, PMLP$_{JKNet}$, PMLP$_{ResSGC}$ and PMLP$_{ResAPP}$ where model depth seems to have less impact on their generalization performance.

\paragraph{Graph Heterophily.} We further conduct experiments on six datasets (i.e., Chameleon, Squirrel, Film, Cornell, Texas, Wisconsin~\citep{pei2020geom}) with high heterophily levels, and the results are shown in Table.~\ref{hetero_result}. As we can see, PMLP achieves better performance than MLP when its GNN counterpart can outperform MLP, but is otherwise inferior than MLP. This indicates that the inherent generalization ability of a certain GNN architecture is also related to whether the message passing scheme is suitable for the characteristics of data, which also partially explains why training more dedicated GNN architectures such as H2GCN~\citep{zhu2020beyond} is helpful for improving model's generalization performance on heterophilic graphs.

\clearpage

\begin{table}[t!]
\centering
\caption{For SGC and APPNP styles, layer number denotes the number of MP layers. The number of FF layers and hidden size are fixed as $2$ and $64$. `Res' denotes using residual connection in the form of $\boldsymbol{X}^{(k)}=(1-\alpha) \operatorname{MP}(\boldsymbol{X}^{(k-1)})+\alpha \boldsymbol{X}^{(0)}$.} \label{tab_over1}
\resizebox{1.0\textwidth}{!}{
\setlength{\tabcolsep}{1.5mm}{
\begin{tabular}{@{}ll|ccccccc@{}}
\toprule & \# Layers & 2 & 4 & 8 & 16 & 32 & 64 & 128 \\ 
\hline\specialrule{0em}{2pt}{2pt}

& MLP & $52.56 \pm 0.41$ & $52.56 \pm 0.41$ & $52.56 \pm 0.41$ & $52.56 \pm 0.41$ & $52.56 \pm 0.41$ & $52.56 \pm 0.41$ & $52.56 \pm 0.41$\\\cmidrule{1-9}

\multirow{2}{*}{\rotatebox{90}{$\alpha=0$}}
& SGC & $73.12 \pm 1.20$ & $75.36 \pm 1.40$ & $76.78 \pm 1.63$ & $75.22 \pm 2.08$ & $73.08 \pm 2.75$ & $64.70 \pm 4.14$ & $47.64 \pm 3.09$\\ \cmidrule(l){2-9} 

& PMLP$_{SGC}$ & $73.48 \pm 1.30$ & $75.38 \pm 1.73$ & $76.20 \pm 2.12$ & $75.66 \pm 2.18$ & $73.60 \pm 3.07$ & $67.76 \pm 4.38$ & $53.64 \pm 6.35$\\

\cmidrule{1-9} 
\multirow{5}{*}{\rotatebox{90}{\quad$\alpha=0.1$}}

& SGC+Res & $71.56 \pm 1.11$ & $73.98 \pm 1.16$ & $75.02 \pm 1.28$ & $75.50 \pm 1.05$ & $75.40 \pm 1.05$ & $\mathop{75.38 \pm 1.05}\limits_{\mbox{(Converged)}}$ & $\mathop{75.38 \pm 1.05}\limits_{\mbox{(Converged)}}$ \\ \cmidrule(l){2-9}

& PMLP$_{SGC+Res}$ & $72.80 \pm 0.67$ & $75.08 \pm 1.14$ & $76.06 \pm 1.13$ & $76.22 \pm 1.18$ & $76.20 \pm 1.23$ & $76.14 \pm 1.14$ & $\mathop{76.14 \pm 1.14}\limits_{\mbox{(Converged)}}$ \\
\hline\hline\specialrule{0em}{2pt}{2pt}

& MLP & $52.56 \pm 0.41$ & $52.56 \pm 0.41$ & $52.56 \pm 0.41$ & $52.56 \pm 0.41$ & $52.56 \pm 0.41$ & $52.56 \pm 0.41$ & $52.56 \pm 0.41$\\\cmidrule{1-9}

\multirow{2}{*}{\rotatebox{90}{$\alpha=0$}}

& APPNP & $73.30 \pm 1.39$ & $76.14 \pm 1.07$ & $77.32 \pm 0.55$ & $76.42 \pm 0.65$ & $75.46 \pm 0.92$ & $70.38 \pm 1.03$ & $52.24 \pm 2.46$\\ \cmidrule(l){2-9} 

& PMLP$_{APP}$ & $74.58 \pm 0.58$ & $77.08 \pm 0.77$ & $77.56 \pm 0.60$ & $76.68 \pm 0.77$ & $75.18 \pm 0.83$ & $71.00 \pm 0.85$ & $51.00 \pm 6.58$\\
\cmidrule{1-9}
\multirow{5}{*}{\rotatebox{90}{\quad$\alpha=0.1$}}

& APPNP+Res & $72.28 \pm 1.56$ & $74.82 \pm 1.31$ & $75.78 \pm 1.18$ & $76.18 \pm 1.11$ & $75.98 \pm 1.01$ & $\mathop{76.04 \pm 0.92}\limits_{\mbox{(Converged)}}$ & $\mathop{76.04 \pm 0.92}\limits_{\mbox{(Converged)}}$ \\ \cmidrule(l){2-9}

& PMLP$_{APP+Res}$ & $72.96 \pm 0.75$ & $75.66 \pm 1.05$ & $76.68 \pm 0.82$ & $76.82 \pm 0.81$ & $76.86 \pm 0.87$ & $76.86 \pm 0.87$ & $\mathop{76.86 \pm 0.87}\limits_{\mbox{(Converged)}}$\\
\bottomrule
\end{tabular}}} 
\end{table}

\begin{table}[t!]
\centering
\caption{Layer number denotes the number of MP+FF layers, and the number of FF layers is fixed as $2$. The hidden size is fixed as $64$. For GCNII, $\mathbf{H}^{(\ell+1)}=\sigma(((1-\alpha_{\ell}) \operatorname{MP}(\mathbf{H}^{(\ell)})+\alpha_{\ell} \mathbf{H}^{(0)})((1-\beta_{\ell}) \mathbf{I}_n+\beta_{\ell} \mathbf{W}^{(\ell)}))$, we set $\alpha_{\ell}=0.1$, $\beta_{\ell}=0.5/\ell$. ResNet here denotes MLP with residual connections (or equivalently, GCNII without MP operations). For JKNet, we use concatenation for layer aggregation. MLP+JK denotes MLP with jumping knowledge (or equivalently, JKNet without MP operations)}. \label{tab_over2}
\resizebox{1.0\textwidth}{!}{
\setlength{\tabcolsep}{1.5mm}{
\begin{tabular}{@{}ll|ccccccc@{}}
\toprule
& \# Layers & 2 & 4 & 8 & 16 & 32 & 64 & 128 \\ 
\hline\specialrule{0em}{2pt}{2pt}

\multirow{4}{*}{\rotatebox{90}{w/o res.}} 

& MLP & $52.56 \pm 0.41$ & $48.18 \pm 4.17$ & $32.26 \pm 5.66$ & $30.64 \pm 6.75$ & $27.60 \pm 6.76$ & $19.86 \pm 3.25$ & $20.14 \pm 2.01$\\\cmidrule(l){2-9}

& GCN & $73.84 \pm 1.49$ & $74.86 \pm 3.03$ & $45.52 \pm 6.10$ & $33.50 \pm 5.23$ & $24.08 \pm 6.90$ & $27.82 \pm 2.76$ & $19.60 \pm 9.52$\\ \cmidrule(l){2-9} 

& PMLP$_{GCN}$ & $73.96 \pm 0.78$ & $74.56 \pm 2.94$ & $44.98 \pm 6.51$ & $29.58 \pm 10.74$ & $26.66 \pm 5.31$ & $27.10 \pm 8.63$ & $27.76 \pm 7.14$\\
\cmidrule[1pt]{1-9}

\multirow{4}{*}{\rotatebox{90}{with res.}} 

& ResNet & $53.08 \pm 1.15$ & $53.68 \pm 0.52$ & $53.12 \pm 1.57$ & $52.70 \pm 2.16$ & $49.36 \pm 1.70$ & $45.74 \pm 1.54$ & $40.74 \pm 2.39$\\\cmidrule(l){2-9} 

& GCNII & $67.36 \pm 1.57$ & $74.54 \pm 1.02$ & $76.00 \pm 0.62$ & $76.28 \pm 0.68$ & $75.42 \pm 1.65$ & $75.18 \pm 0.87$ & $68.24 \pm 1.89$ \\ \cmidrule(l){2-9}

& PMLP$_{GCNII}$ & $69.08 \pm 0.80$ & $74.98 \pm 0.51$ & $75.26 \pm 1.29$ & $75.62 \pm 1.48$ & $75.12 \pm 1.38$ & $72.64 \pm 2.24$ & $68.52 \pm 1.47$ \\
\cmidrule[1pt]{1-9} 

\multirow{4}{*}{\rotatebox{90}{with res.}} 

& MLP+JK & $52.76 \pm 1.38$ & $53.46 \pm 1.29$ & $54.48 \pm 1.19$ & $56.10 \pm 0.73$ & $53.10 \pm 1.30$ & $49.38 \pm 1.96$ & $30.40 \pm 1.63$\\\cmidrule(l){2-9} 

& JKNet & $65.50 \pm 0.78$ & $72.02 \pm 1.15$ & $72.22 \pm 1.08$ & $71.50 \pm 0.83$ & $71.38 \pm 1.61$ & $60.66 \pm 2.63$ & $51.52 \pm 4.00$ \\ \cmidrule(l){2-9}

& PMLP$_{JKNet}$ & $67.82 \pm 0.78$ & $72.76 \pm 1.59$ & $74.02 \pm 0.53$ & $73.92 \pm 1.04$ & $72.76 \pm 1.28$ & $67.70 \pm 0.83$ & $44.70 \pm 6.47$ \\
\bottomrule
\end{tabular}}} 
\end{table}

\begin{table}[t!]
\centering
\caption{Mean and STD of testing accuracy on datasets with high heterophily level.}
\resizebox{1.0\textwidth}{!}{
    \setlength{\tabcolsep}{2.5mm}{
        \begin{tabular}{@{}ll|cccccc@{}}
            \toprule & Dataset & Chameleon & Squirrel & Film & Cornell & Texas & Wisconsin \\ 
            \hline
            \hline\specialrule{0em}{2pt}{2pt}
            \multirow{3}{*}{\rotatebox{90}{GNNs}} & GCN & $56.27 \pm 1.60$ & $41.21 \pm 0.48$ & $29.88 \pm 0.56$ & $67.03 \pm 4.44$ & $63.24 \pm 8.02$ & $63.92 \pm 4.51$ \\ 
            \cmidrule(l){2-8} 
            & SGC & $58.73 \pm 2.09$ & $43.73 \pm 1.95$ & $30.92 \pm 0.68$ & $58.92 \pm 10.54$ & $62.70 \pm 7.00$ & $66.67 \pm 4.60$ \\ 
            \cmidrule(l){2-8} 
            & APPNP & $57.98 \pm 3.70$ & $41.59 \pm 1.50$ & $32.20 \pm 1.37$ & $65.95 \pm 3.08$ & $65.95 \pm 2.42$ & $66.27 \pm 2.56$ \\ \hline
            \hline\specialrule{0em}{2pt}{2pt}
            \multirow{4}{*}{\rotatebox{90}{MLPs$\qquad\qquad$}} & MLP & $49.61 \pm 1.36$ & $29.20 \pm 2.32$ & $36.43 \pm 2.21$ & $78.92 \pm 6.73$ & $76.76 \pm 3.08$ & $83.53 \pm 3.28$ \\ 
            \cmidrule(l){2-8} 
            & \textbf{PMLP}$_{GCN}$ & $57.02 \pm 3.74$ & $36.39 \pm 3.28$ & $29.75 \pm 1.16$ & $65.41 \pm 8.20$ & $62.70 \pm 11.04$ & $66.67 \pm 2.77$ \\
            & $\Delta_{GNN}$ & $+1.33\%$ & $-11.70\%$ & $-0.44\%$ & $-2.42\%$ & $-0.85\%$ & $+4.30\%$ \\
            & $\Delta_{MLP}$ & $+14.94\%$ & $+24.62\%$ & $-18.34\%$ & $-17.12\%$ & $-18.32\%$ & $-20.18\%$ \\
            \cmidrule(l){2-8} 
            & \textbf{PMLP}$_{SGC}$ & $55.83 \pm 2.51$ & $39.44 \pm 1.93$ & $28.66 \pm 0.61$ & $55.14 \pm 9.09$ & $62.70 \pm 4.44$ & $62.35 \pm 5.26$ \\
            & $\Delta_{GNN}$ & $-4.94\%$ & $-9.81\%$ & $-7.31\%$ & $-6.42\%$ & $0.00\%$ & $-6.48\%$ \\
            & $\Delta_{MLP}$ & $+12.54\%$ & $+35.07\%$ & $-21.33\%$ & $-30.13\%$ & $-18.32\%$ & $-25.36\%$ \\
            \cmidrule(l){2-8} 
            & \textbf{PMLP}$_{APP}$ & $57.28 \pm 2.25$ & $39.73 \pm 1.85$ & $31.54 \pm 0.74$ & $68.11 \pm 6.16$ & $65.95 \pm 8.24$ & $67.45 \pm 4.51$ \\
            & $\Delta_{GNN}$ & $-1.21\%$ & $-4.47\%$ & $-2.05\%$ & $+3.28\%$ & $0.00\%$ & $+1.78\%$ \\
            & $\Delta_{MLP}$ & $+15.46\%$ & $+36.06\%$ & $-13.42\%$ & $-13.70\%$ & $-14.08\%$ & $-19.25\%$ \\
            \bottomrule
        \end{tabular}
    }
}
\label{hetero_result}
\end{table}

\clearpage
\section{Implementation Details} \label{app_detail}
We present implementation details for our experiments for reproducibility. We implement our model as well as the baselines with Python 3.7, Pytorch 1.9.0 and Pytorch Geometric 1.7.2. All parameters are initialized with Xavier initialization procedure. We train the model by Adam optimizer. Most of the experiments are running with a NVIDIA 2080Ti with 11GB memory, except that for large-scale datasets we use a NVIDIA 3090 with 24GB memory.

\begin{table}[t!]
\centering
\caption{Statistics of datasets.}\label{tab_data}
\resizebox{0.99\textwidth}{!}{
\setlength{\tabcolsep}{4mm}{
\begin{tabular}{l|llll}
\toprule
Dataset & \# Classes & \# Nodes & \# Edges & \# Node features  \\ \midrule
Cora~\citep{mccallum2000automating} & 7 & 2,708 & 5,278 & 1,433  \\
Citeseer~\citep{sen2008collective} & 6 & 3,327 & 4,552 & 3,703  \\
Pubmed~\citep{namata2012query} & 3 & 19,717 & 44,324 & 500  \\
A-Photo~\citep{amazon-dataset} & 8 & 7,650 & 119,081 & 745  \\
A-Computer & 10 & 13,752 & 245,861 & 767  \\
Coauthor-CS~\citep{coauthor-dataset} & 15 & 18,333 & 81,894 & 6,805  \\
Coauthor-Physics & 5 & 34,493 & 247,962 & 8,415  \\
OGBN-Arxiv~\citep{hu2020open} & 40 & 169,343 & 1,157,799 & 128  \\
OGBN-Products & 47 & 2,449,029 & 61,859,076 & 100  \\
Flickr~\citep{zeng2019graphsaint} & 7 & 89,250 & 449,878 & 500  \\ \bottomrule
\end{tabular}}}
\end{table}

\begin{table}[t!]
\centering
\caption{Summary of model architectures.}
\resizebox{0.99\textwidth}{!}{
\setlength{\tabcolsep}{2mm}{
\begin{tabular}{@{}ll|cccccccccc@{}}
\toprule
                      &    & Cora & Citeseer & Pubmed & Photo & Computer & CS & Physics & Arxiv & Products & Flickr  \\ 
\hline
\hline\specialrule{0em}{2pt}{2pt}
\multirow{3}{*}{\rotatebox{90}{GCN}} &Hidden Size  & 64 & 64 & 64 & 32 & 256 & 128 & 128 & 64 & 64 & 64 \\ \cmidrule(l){2-12} 
                      & \# MP Layer      & 2  & 2 & 2 & 2 & 2 & 2& 2 &2 &2 &2\\ \cmidrule(l){2-12}
                      & \# FF Layer     & 2  & 2 & 2 & 2 & 2 & 2& 2 &2 &2 &2 \\ 
\hline\specialrule{0em}{2pt}{2pt}
\multirow{3}{*}{\rotatebox{90}{PMLP$_{gcn}$}} &Hidden Size  & 64 & 64 & 64 & 32 & 256 & 128 & 128 & 64 & 64 & 64 \\ \cmidrule(l){2-12} 
                      & \# MP Layer      & 2  & 2 & 2 & 2 & 2 & 2& 2 &2 &2 &2\\ \cmidrule(l){2-12}
                      & \# FF Layer     & 2  & 2 & 2 & 2 & 2 & 2& 2 &2 &2 &2 \\ \hline
\hline\specialrule{0em}{2pt}{2pt}
\multirow{3}{*}{\rotatebox{90}{SGC}} &Hidden Size & 64 & 64 & 64 & 32 & 256 & 128 & 128 & 64 & 64 & 64 \\ \cmidrule(l){2-12} 
                      & \# MP Layer      & 2  & 2 & 2 & 2 & 2 & 2& 2 &2 &2 &2\\ \cmidrule(l){2-12}
                      & \# FF Layer     & 2  & 2 & 2 & 2 & 2 & 2& 2 &2 &2 &2 \\ 
\hline\specialrule{0em}{2pt}{2pt}
\multirow{3}{*}{\rotatebox{90}{PMLP$_{sgc}$}} &Hidden Size  & 64 & 64 & 64 & 32 & 256 & 128 & 128 & 64 & 64 & 64 \\ \cmidrule(l){2-12} 
                      & \# MP Layer      & 2  & 2 & 2 & 2 & 2 & 2& 2 &2 &2 &2\\ \cmidrule(l){2-12}
                      & \# FF Layer     & 2  & 2 & 2 & 2 & 2 & 2& 2 &2 &2 &2 \\ \hline
\hline\specialrule{0em}{2pt}{2pt}
\multirow{3}{*}{\rotatebox{90}{APPNP}} &Hidden Size & 64 & 64 & 64 & 32 & 256 & 128 & 128 & 64 & 64 & 64 \\ \cmidrule(l){2-12} 
                    & \# MP Layer      & 2  & 2 & 2 & 2 & 2 & 2& 2 &2 &2 &1\\ \cmidrule(l){2-12}
                      & \# FF Layer     & 2  & 2 & 2 & 2 & 2 & 2& 2 &2 &2 &2 \\ 
\hline\specialrule{0em}{2pt}{2pt}
\multirow{3}{*}{\rotatebox{90}{PMLP$_{app}$}} &Hidden Size  & 64 & 64 & 64 & 32 & 256 & 128 & 128 & 64 & 64 & 64 \\ \cmidrule(l){2-12} 
                     & \# MP Layer      & 2  & 2 & 2 & 2 & 2 & 2& 2 &2 &2 &1\\ \cmidrule(l){2-12}
                      & \# FF Layer     & 2  & 2 & 2 & 2 & 2 & 2& 2 &2 &2 &2 \\ \hline
\hline\specialrule{0em}{2pt}{2pt}
\multirow{3}{*}{\rotatebox{90}{MLP}} &Hidden Size  & 64 & 64 & 64 & 32 & 256 & 128 & 128 & 64 & 64 & 64 \\ \cmidrule(l){2-12} 
                      & \# MP Layer      & /  &  /& / & / &/ &/ &/&/&/&/\\ \cmidrule(l){2-12}
                      & \# FF Layer     & 2  & 2 & 2 & 2 & 2 & 2& 2 &2 &2 &2 \\  \bottomrule
\end{tabular}}} \label{hyper_set}
\end{table}

\subsection{Dataset Description} 
We use sixteen widely adopted node classification benchmarks involving different types of networks: three citations networks (Cora, Citeseer and Pubmed), two product co-occurrency networks (Amazon-Computer and Amazon-Photo), two coauthor-ship networks (Coauthor-CS, Coauthor-Computer), and three large-scale networks (OGBN-Arxiv, OGBN-Products and Flickr).
For Cora, Citeseer and Pubmed, we use the provided split in~\citep{kipf2017semi}. For Amazon-Computer, Amazon-Photo, Coauthor-CS and Coauthor-Computer, we randomly sample 20 nodes from each class as labeled nodes, 30 nodes for validation and all other nodes for test following~\citep{shchur2018pitfalls}. 
For two large-scale datasets, we follow the original splitting~\cite{hu2020open} for evaluation. For Flickr, we use random split where the training and validation proportions are 10\%.
The statistics of these datasets are summarized in Table.~\ref{tab_data}.

\subsection{Hyperparameter Search}
We use the same MLP architecture (i.e., number of FF layers and size of hidden states) as backbone for models in the same dataset, same GNN architecture (i.e., number of MP layers) for PMLP and its GNN counterpart, and finetune hyperparameters for each model including dropout rate (from $\{0.1, 0.2, 0.3, 0.4, 0.5, 0.6, 0.7, 0.8, 0.9\}$), weight decay factor (from $\{0, 0.0001, 0.001, 0.01, 0.1\}$), and learning rate (from $\{0.0001, 0.001, 0.01, 0.1\}$) using grid search. 

For model architecture hyperparameters (i.e., layer number, size of hidden states), we fix them as reported in Table.~\ref{hyper_set}, instead of fine-tuning them in favor of GNN or PMLP for each dataset which might introduce bias into their comparison. By default, we set (FF and MP) layer number as 2, and hidden size as 64, but manually adjust in case the performance of GNN is far from the optimal. Furthermore, we have discussed different settings for these architecture hyperparameters and find the performance of PMLP is consistently close to its GNN counterpart.

For other hyperparameters (i.e., learning rate, dropout rate, weight decay factor), we finetune them separately for each model on each dataset based on the performance on validation set. For PMLP, we use the MLP architecture for validation rather than using GNN since we find there is only slight difference in performance. In that sense, all PMLPs share the same training and validation process as the vanilla MLP, making them exactly the same model before inference.

\clearpage
\section{Additional Experimental Results} \label{app_add}
We supplement more experimental results in this section including extensions of the results in the main text and visualizations of the internal representations of nodes learned by 2-layer MLP, GNN, and PMLP on Cora and Citeseer datasets. 

\begin{figure}[t]
	\centering
	\subfigure[$\rm GCN$]{
    	\includegraphics[width=0.32\textwidth]{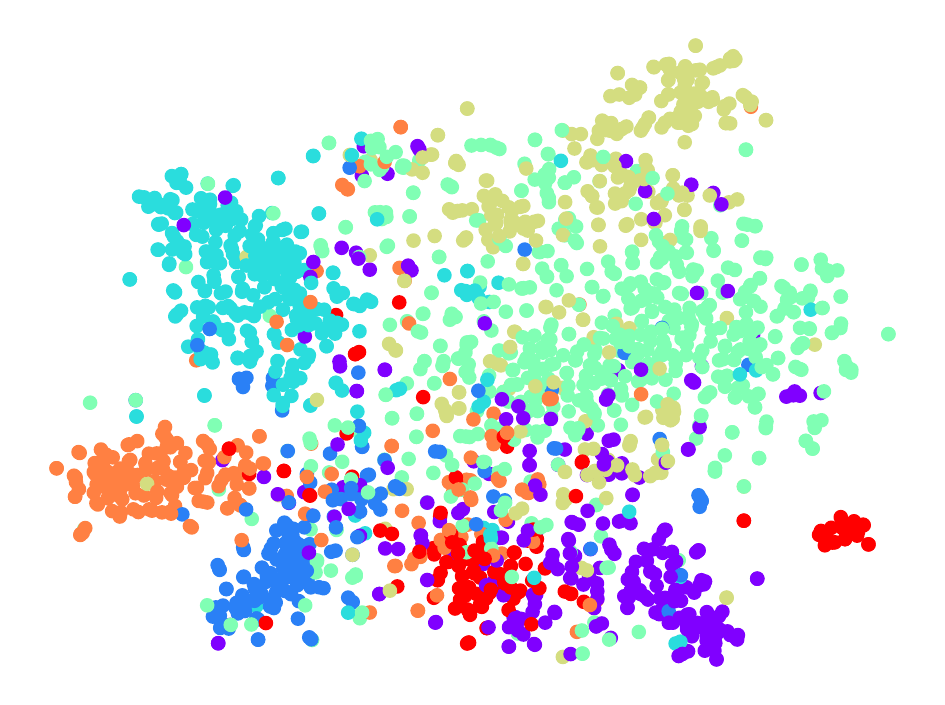}
	}%
	\subfigure[$\rm{PMLP}_{GCN}$]{
    	\includegraphics[width=0.32\textwidth]{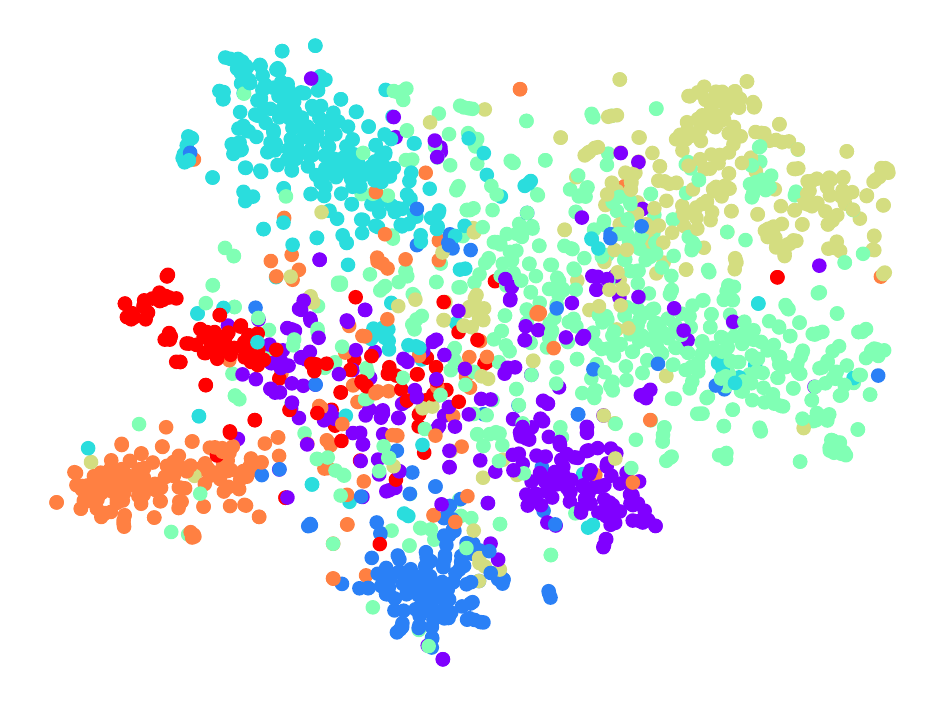}
	}%
	\subfigure[$\rm MLP$]{
    	\includegraphics[width=0.32\textwidth]{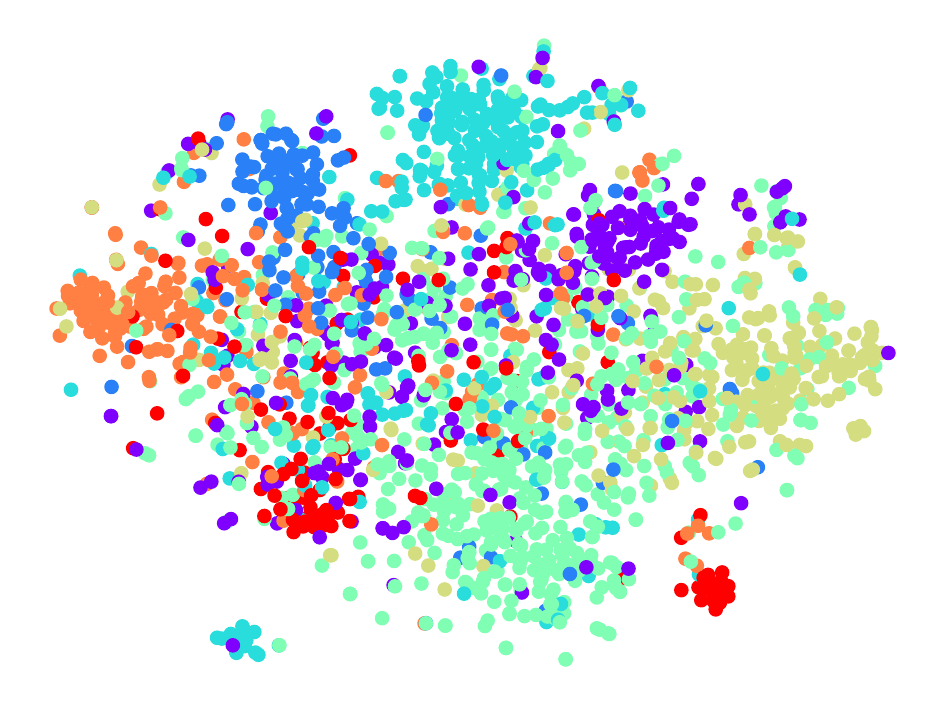}
	}%
	\caption{Visualization of node embeddings (2-D projection by t-SNE) in the internal layer for two-layer MLP, GCN and PMLP on Cora.} \label{fig_embed1}
\end{figure}

\begin{figure}[t]
	\centering
	\subfigure[$\rm SGC$]{
    	\includegraphics[width=0.32\textwidth]{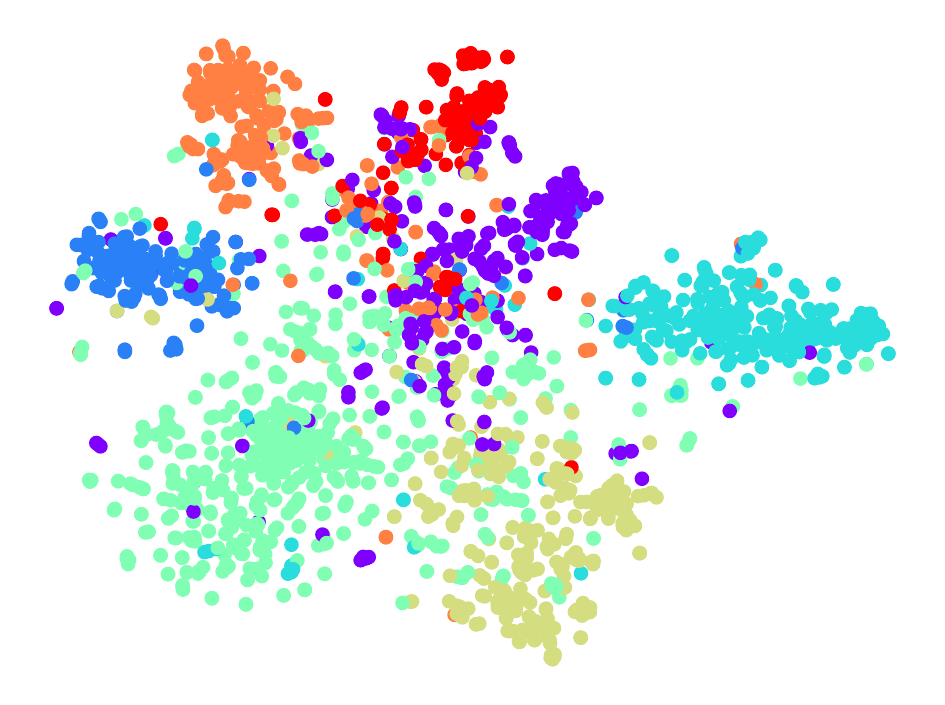}
	}%
	\subfigure[$\rm{PMLP}_{SGC}$]{
    	\includegraphics[width=0.32\textwidth]{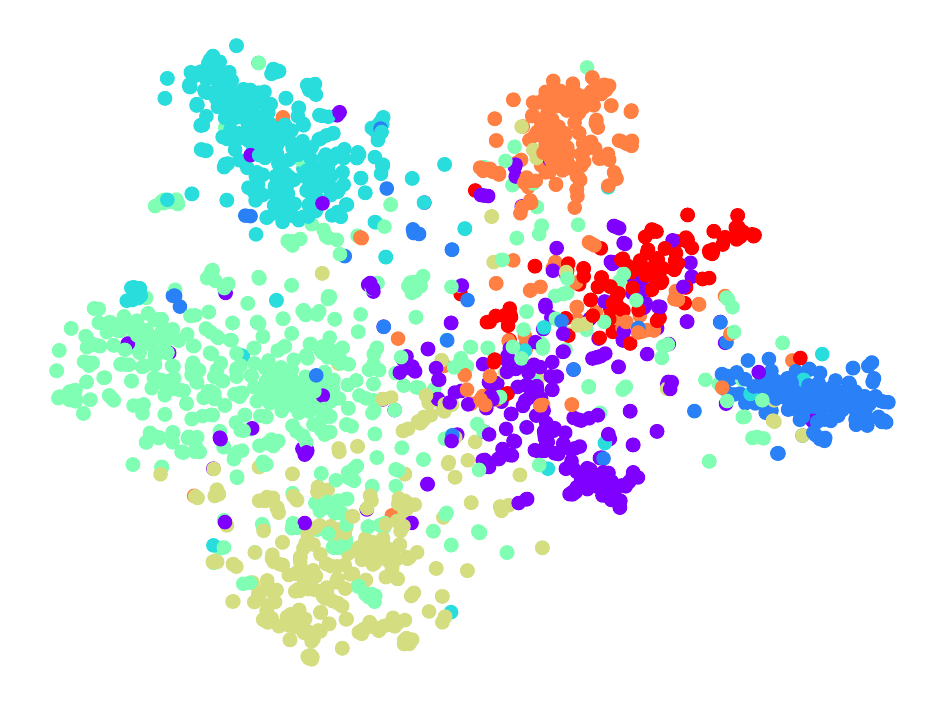}
	}%
	\subfigure[$\rm MLP$]{
    	\includegraphics[width=0.32\textwidth]{fig/emb-cora-mlp.sh.npy.pdf}
	}%
	\caption{Visualization of node embeddings (2-D projection by t-SNE) in the internal layer for two-layer MLP, SGC and PMLP on Cora.} \label{fig_embed2}
\end{figure}

\begin{figure}[t]
	\centering
	\subfigure[$\rm GCN$]{
    	\includegraphics[width=0.32\textwidth]{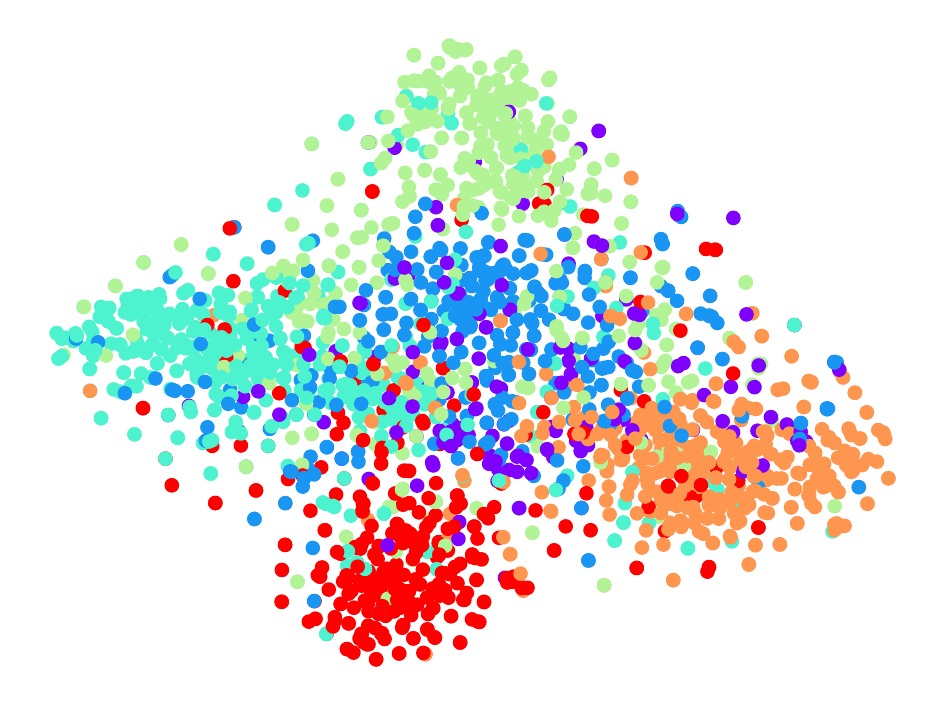}
	}%
	\subfigure[$\rm{PMLP}_{GCN}$]{
    	\includegraphics[width=0.32\textwidth]{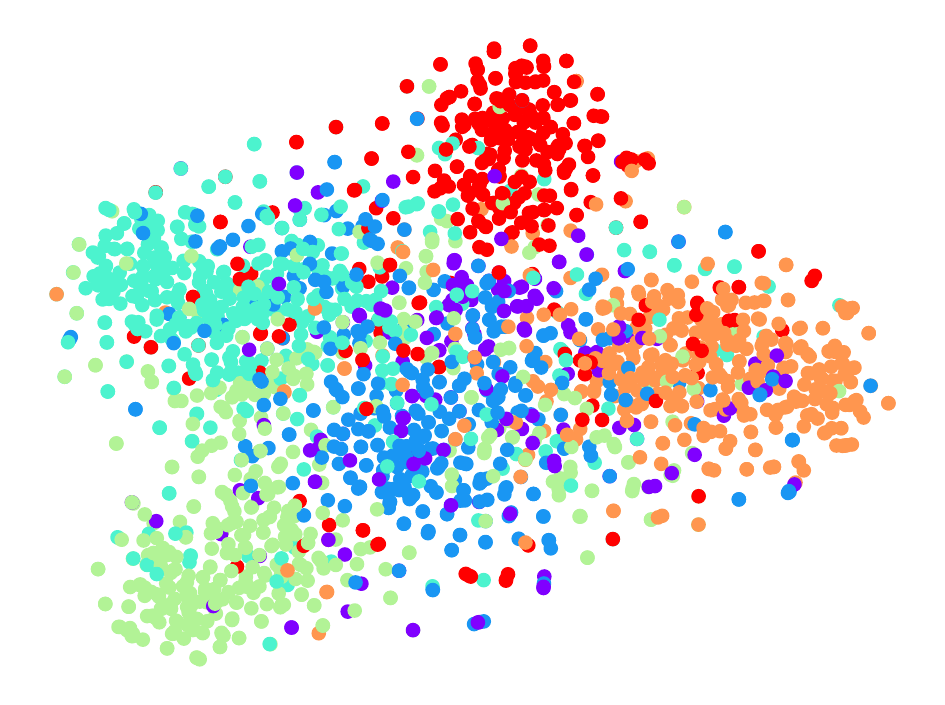}
	}%
	\subfigure[$\rm MLP$]{
    	\includegraphics[width=0.32\textwidth]{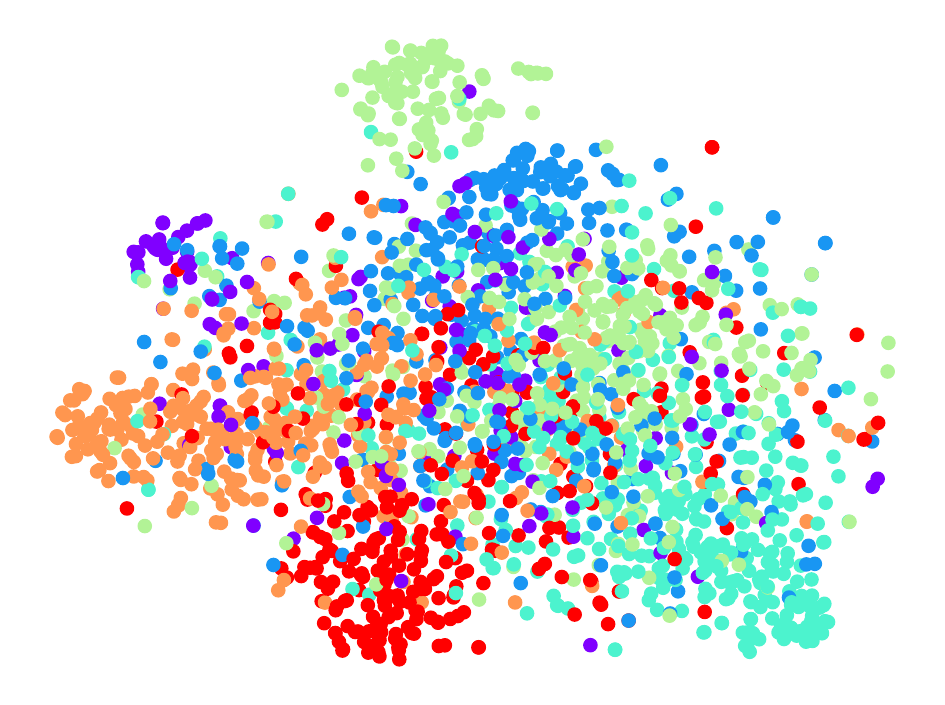}
	}%
	\caption{Visualization of node embeddings (2-D projection by t-SNE) in the internal layer for two-layer MLP, GCN and PMLP on Citeseer.} \label{fig_embed3}
\end{figure}

\begin{figure}[t]
	\centering
	\subfigure[$\rm SGC$]{
    	\includegraphics[width=0.32\textwidth]{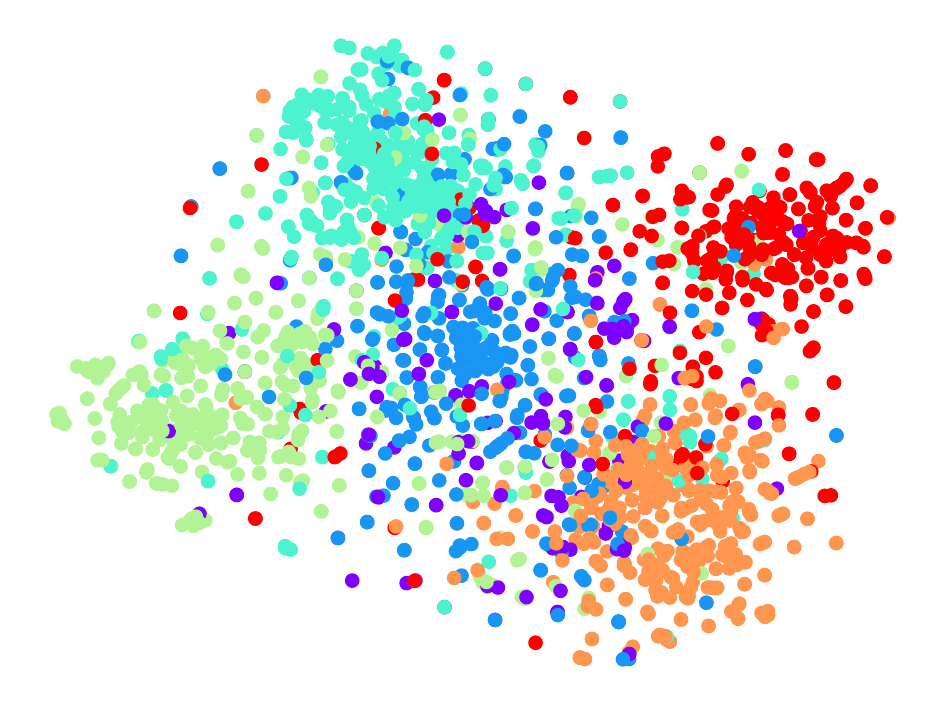}
	}%
	\subfigure[$\rm{PMLP}_{SGC}$]{
    	\includegraphics[width=0.32\textwidth]{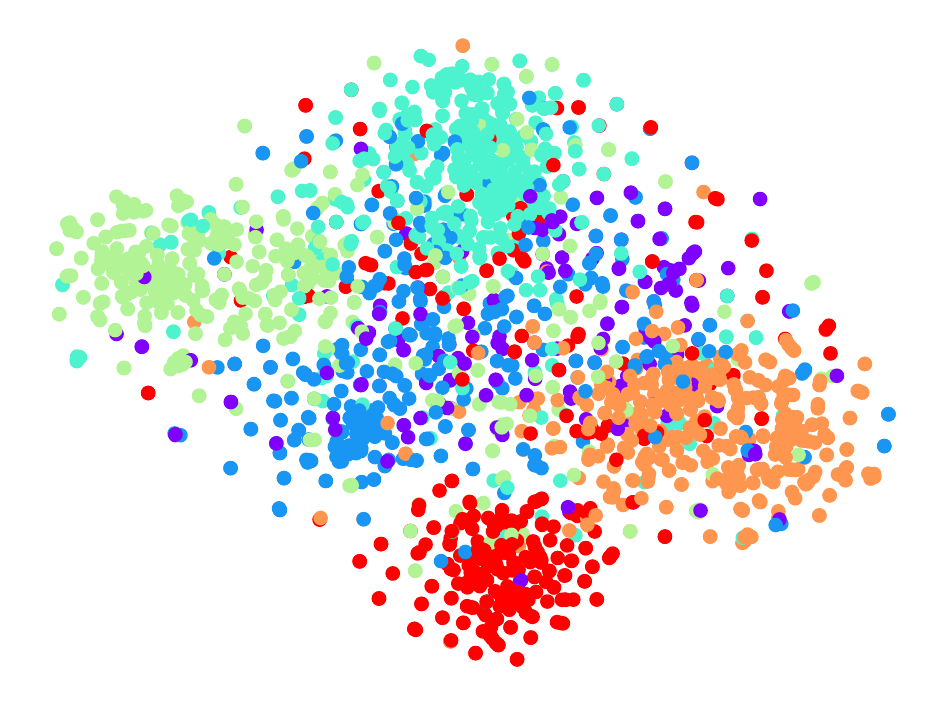}
	}%
	\subfigure[$\rm MLP$]{
    	\includegraphics[width=0.32\textwidth]{fig/emb-citeseer-mlp.sh.npy.pdf}
	}%
	\caption{Visualization of node embeddings (2-D projection by t-SNE) in the internal layer for two-layer MLP, SGC and PMLP on Citeseer.} \label{fig_embed4}
\end{figure}

As we can see from Fig.~\ref{fig_embed1}-\ref{fig_embed4}, both PMLPs and GNNs show better capability for separating nodes of different classes than MLP in the internal layer, despite the fact that PMLPs share the same set of weights with MLP. Such results might indicate that the superior classification performance of GNNs mainly stem from the effects of message passing in inference, rather than GNNs' ability of learning better node representations.

\begin{figure}[h!]
	\centering
    \vspace{30pt}
	\subfigure[GCN v.s. PMLP$_{GCN}$ v.s. MLP]{
    	\includegraphics[width=1.0\textwidth]{fig/split_coarsen_noise_gcn.pdf}
	}
	\subfigure[SGC v.s. PMLP$_{SGC}$ v.s. MLP]{
    	\includegraphics[width=1.0\textwidth]{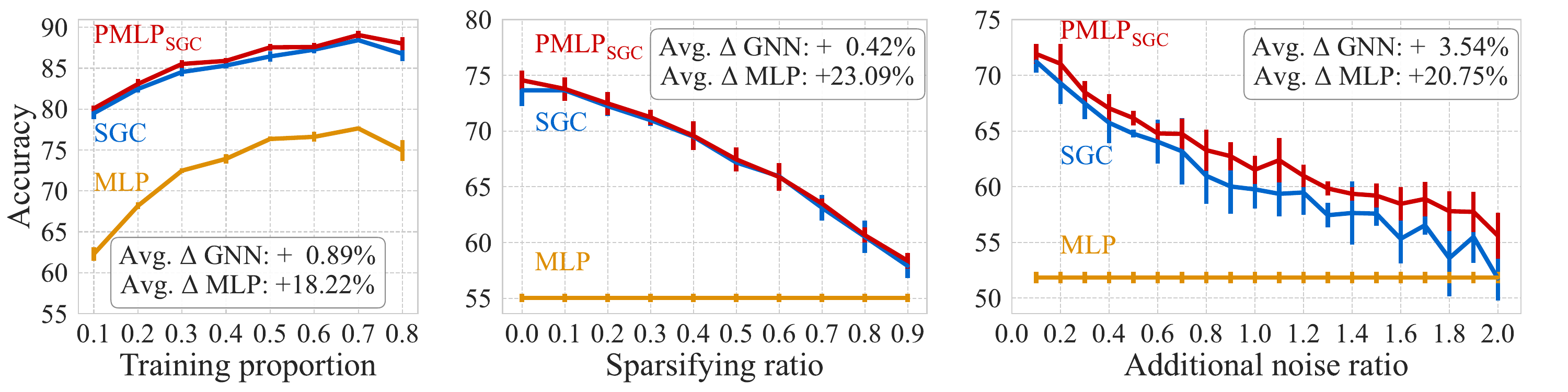}
	}
	\subfigure[APPNP v.s. PMLP$_{APP}$ v.s. MLP]{
    	\includegraphics[width=1.0\textwidth]{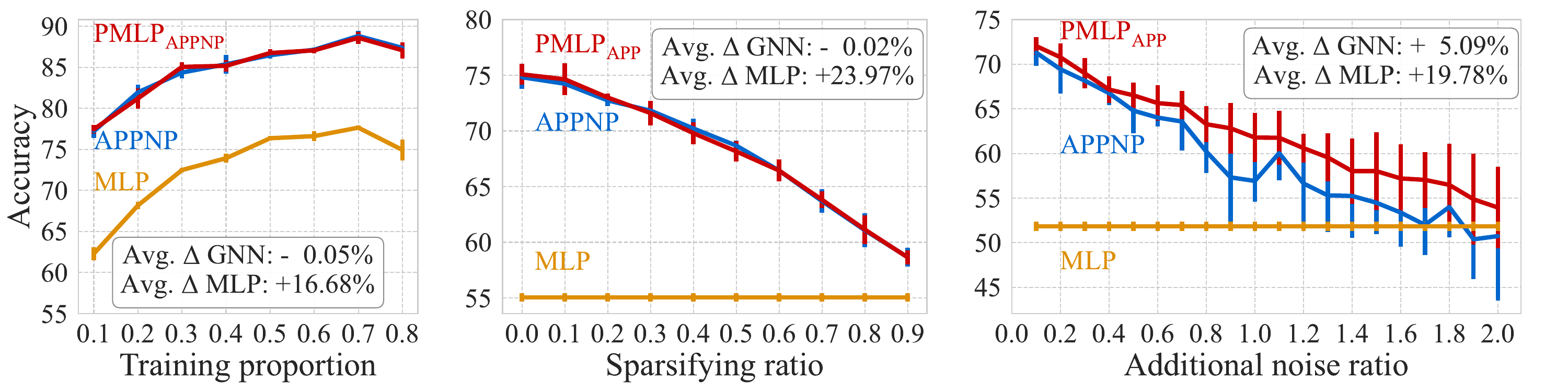}
	}
	\caption{Impact of graph structural information by changing data split, sparsifying the graph, adding random structural noise on Cora.} 
\end{figure}

\clearpage
\begin{figure}[t]
	\centering
	\subfigure{
    	\includegraphics[width=0.99\textwidth]{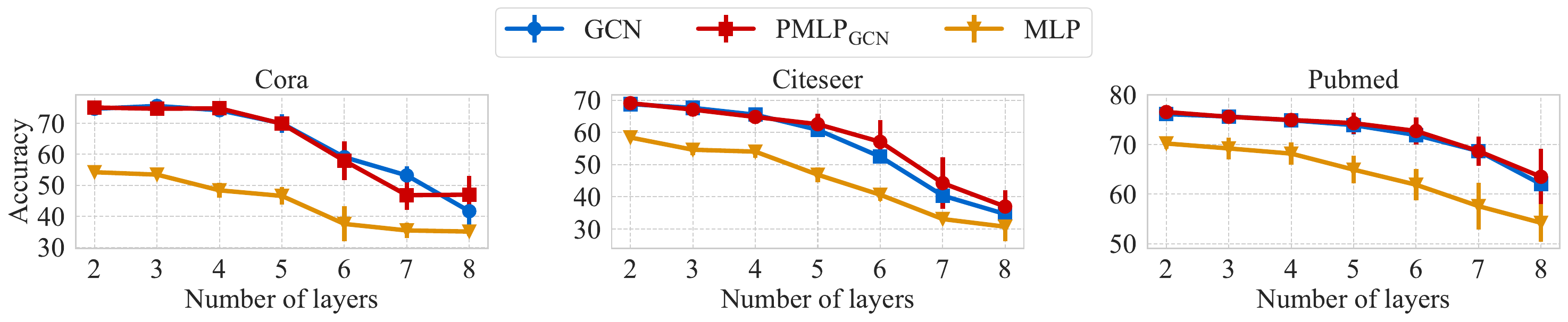}
	}
	\subfigure{
    	\includegraphics[width=0.99\textwidth]{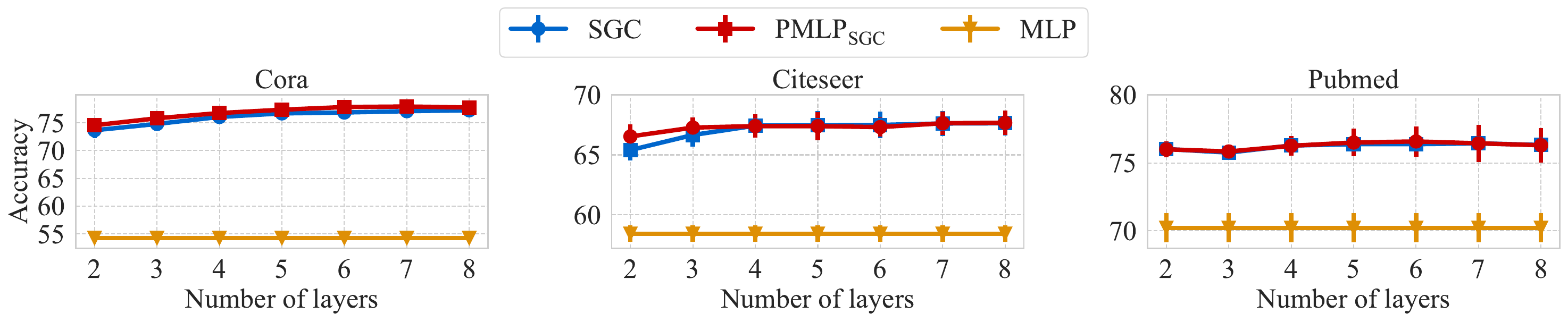}
	}
	\subfigure{
    	\includegraphics[width=0.99\textwidth]{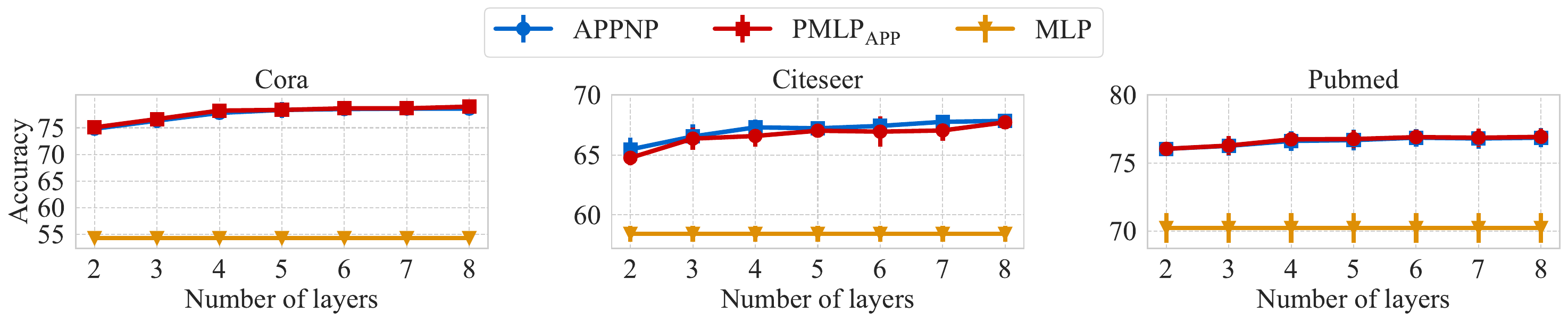}
	}
	\caption{Performance variation with increasing layer number (from $2$ to $8$). Layer number here denotes number of FF and MP layers for GCN, and number of MP layers for SGC and APPNP.} 
\end{figure}

\begin{figure}[t]
	\centering
	\subfigure{
    	\includegraphics[width=0.99\textwidth]{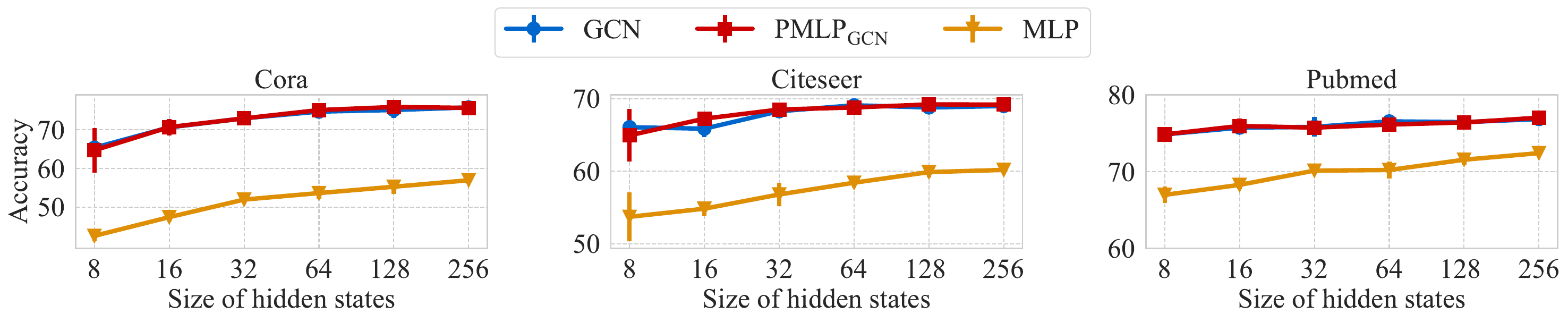}
	}
	\subfigure{
    	\includegraphics[width=0.99\textwidth]{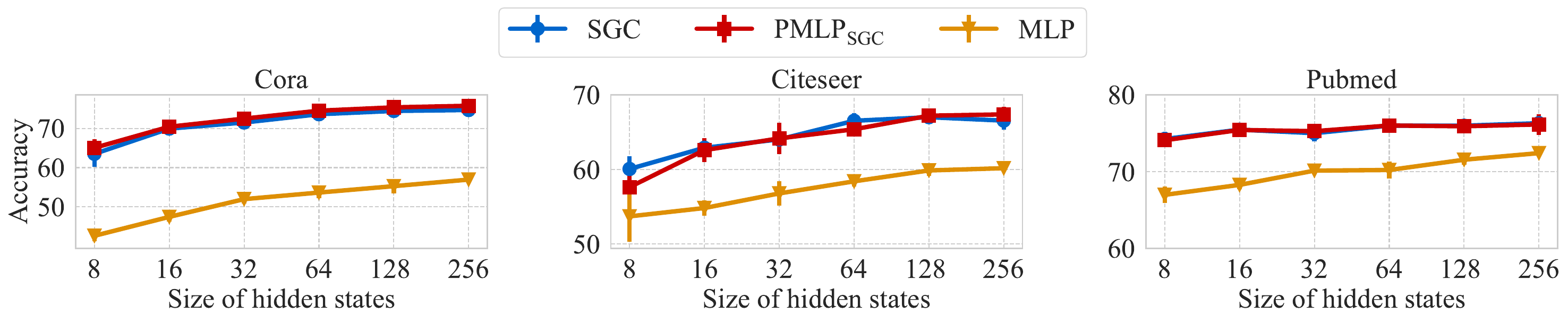}
	}
	\subfigure{
    	\includegraphics[width=0.99\textwidth]{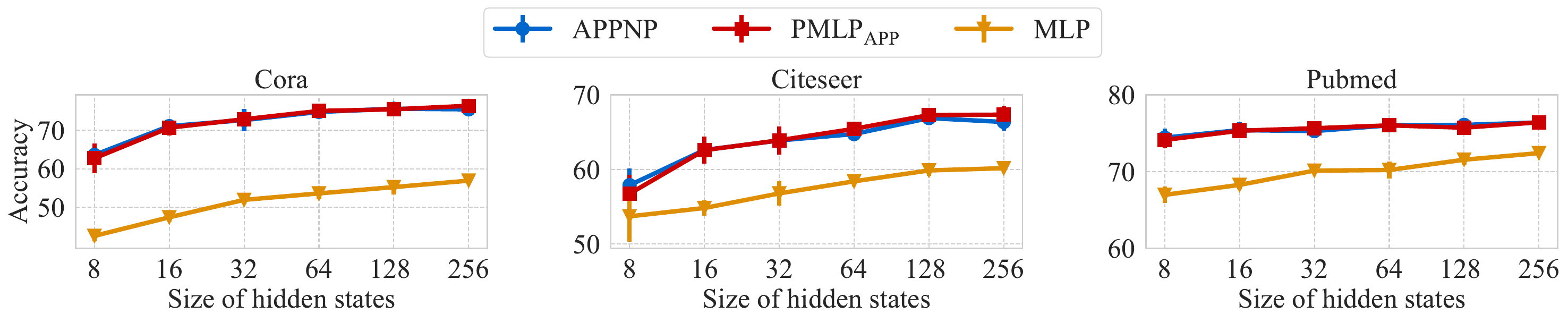}
	}
	\caption{Performance variation with increasing size of hidden states.} 
\end{figure}

\clearpage
\begin{figure}[t]
	\centering
	\subfigure[GCN v.s. PMLP$_{GCN}$ v.s. MLP]{
    	\includegraphics[width=0.99\textwidth]{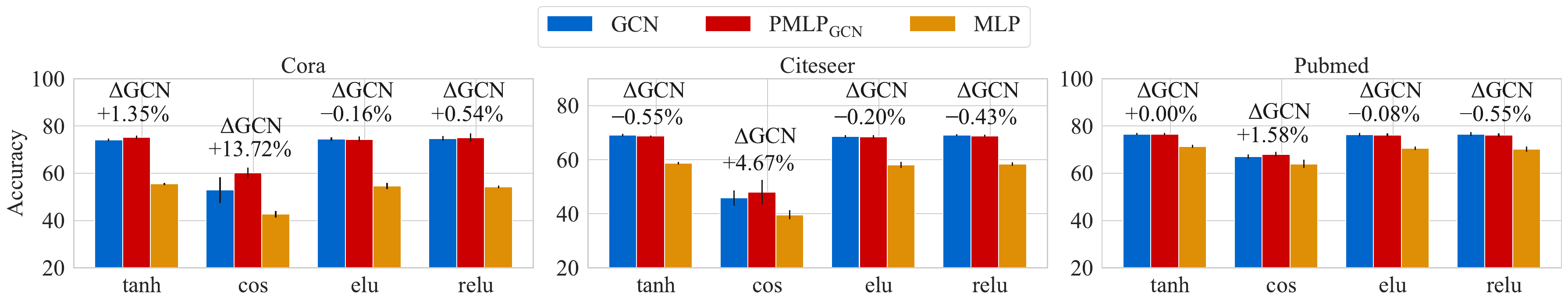}
	}
	\subfigure[SGC v.s. PMLP$_{SGC}$ v.s. MLP]{
    	\includegraphics[width=0.99\textwidth]{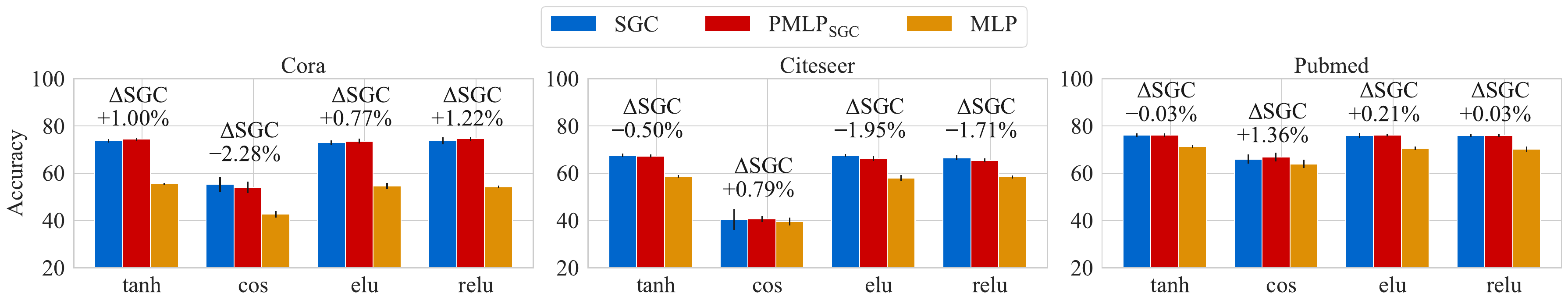}
	}
	\subfigure[APPNP v.s. PMLP$_{APP}$ v.s. MLP]{
    	\includegraphics[width=0.99\textwidth]{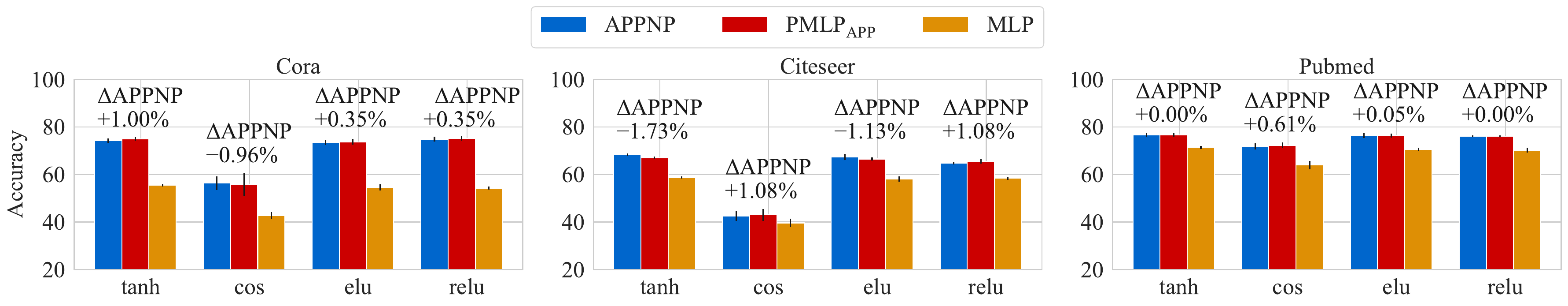}
	}
	\caption{Performance variation with different activation functions in FF layer.} 
\end{figure} 

\begin{figure}[t]
	\centering
	\subfigure[GCN v.s. PMLP$_{GCN}$ v.s. MLP]{
    	\includegraphics[width=0.99\textwidth]{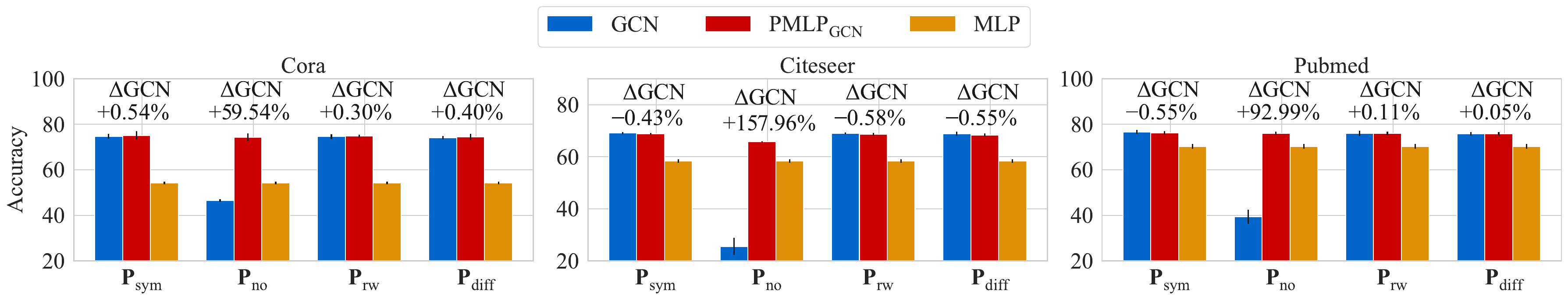}
	}
	\subfigure[SGC v.s. PMLP$_{SGC}$ v.s. MLP]{
    	\includegraphics[width=0.99\textwidth]{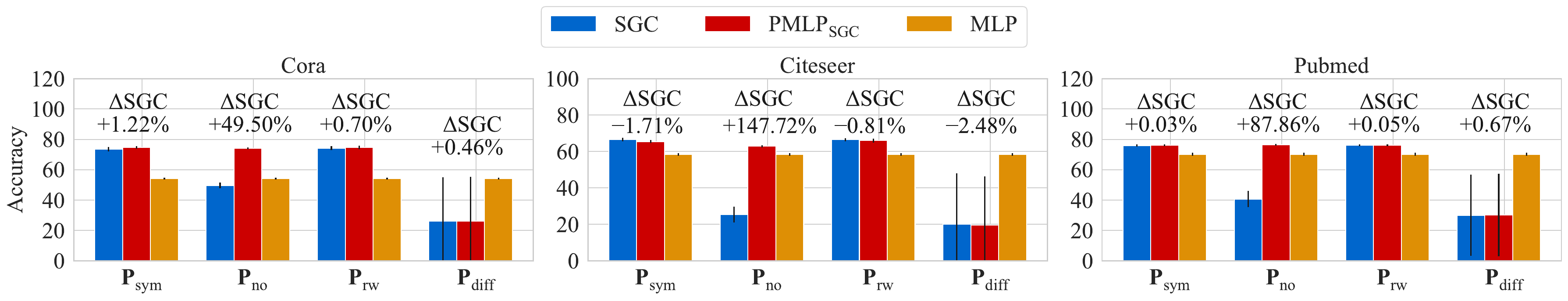}
	}
	\subfigure[APPNP v.s. PMLP$_{APP}$ v.s. MLP]{
    	\includegraphics[width=0.99\textwidth]{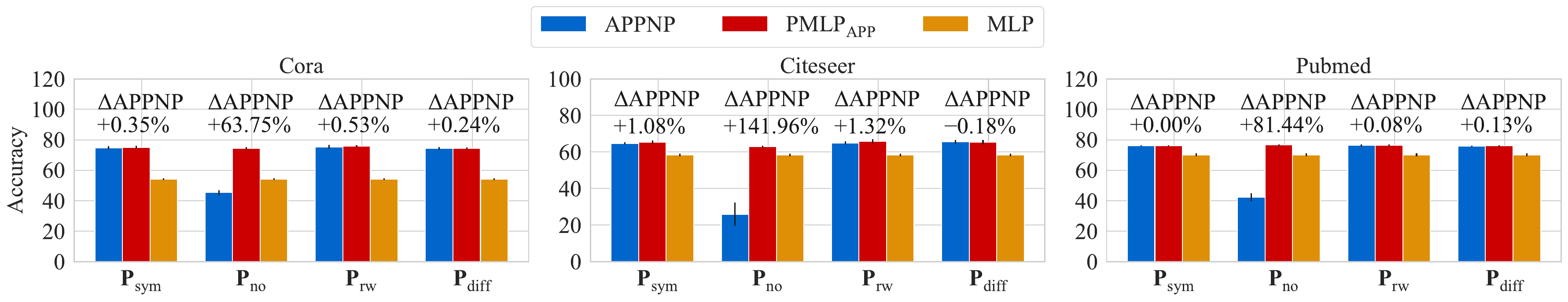}
	}
	\caption{Performance variation with different transition matrices in MP layer.} 
\end{figure}

\clearpage

\end{document}